\documentclass[11pt]{article}
\usepackage[margin=1in]{geometry}
\usepackage{amsthm, amsopn, bm, xcolor, amsmath}
\usepackage{multirow}
\usepackage{booktabs}
\usepackage{pifont}
\usepackage{footnote}

\newcommand{\R}{\mathbb{R}}
\newcommand{\normal}{{\sf N}}
\newcommand{\ep}{\varepsilon}

\DeclareMathOperator*{\argmin}{arg\,min}
\newcommand{\cD}{\mathcal{D}}
\def\stsd{\mathrm{Dis}}

\newcommand{\E}{\mathbb{E}}
\definecolor{dl}{RGB}{255,0,100}

\definecolor{ed}{RGB}{225,100,0}

\newcommand{\cI}{\mathcal{I}}
\newcommand\scalemath[2]{\scalebox{#1}{\mbox{\ensuremath{\displaystyle #2}}}}

\newcommand*\colourcheck[1]{%
  \expandafter\newcommand\csname #1check\endcsname{\textcolor{#1}{\ding{52}}}%
}
\colourcheck{green}

\usepackage{microtype}
\usepackage{graphicx}
\usepackage{subfigure}
\usepackage{booktabs} 

\usepackage{amssymb}
\usepackage{mathtools}
\usepackage[utf8]{inputenc}

\theoremstyle{plain}
\newtheorem{theorem}{Theorem}[section]
\newtheorem{proposition}[theorem]{Proposition}
\newtheorem{lemma}[theorem]{Lemma}
\newtheorem{corollary}[theorem]{Corollary}
\theoremstyle{definition}
\newtheorem{definition}[theorem]{Definition}
\newtheorem{condition}[theorem]{Condition}
\newtheorem{claim}[theorem]{Claim}
\theoremstyle{remark}
\newtheorem{remark}[theorem]{Remark}
\newtheorem{example}[theorem]{Example}

\renewcommand{\epsilon}{\varepsilon}

\title{Demystifying Disagreement-on-the-Line in High Dimensions} 
\author{
Donghwan Lee\footnote{Equal Contribution.} \footnote{Graduate Group in Applied Mathematics and Computational Science, University of Pennsylvania.}
\quad
Behrad Moniri\footnotemark[1]\,\,\footnote{Department of Electrical and Systems Engineering, University of Pennsylvania.} 
\quad
Xinmeng Huang\footnotemark[2]
\quad 
Edgar Dobriban\footnote{Department of Statistics and Data Science, University of Pennsylvania.\newline\hspace*{10pt}
\texttt{\{dh7401, xinmengh\}@sas.upenn.edu, \;\{bemoniri, hassani\}@seas.upenn.edu,\;
dobriban@wharton.upenn.edu}.}
\quad 
Hamed Hassani\footnotemark[3]
}
\date{\today}
\def\hmath$#1${\texorpdfstring{{\rmfamily\textit{#1}}}{#1}}

\usepackage{setspace}
\usepackage[pdfencoding=auto, psdextra]{hyperref}

\begin{document}
\maketitle

\begin{abstract}
Evaluating the performance of machine learning models under distribution shift is challenging, especially
when we only have 
unlabeled data from the shifted (target) domain,
along with labeled data from the original (source) domain.
Recent work suggests that the notion of disagreement, 
the degree to which two models trained with different randomness differ on the same input, is a key to tackle this problem. 
Experimentally, disagreement and prediction error have been shown to be strongly connected,
which has been used to estimate model performance.
Experiments have led to the discovery of the \emph{disagreement-on-the-line} phenomenon, 
whereby the classification error under the target domain is often a linear function of the classification error under the source domain; 
and whenever this property holds, disagreement under the source and target domain follow the same linear relation.
In this work, we develop a theoretical foundation for analyzing disagreement in high-dimensional random features regression; 
and study under what conditions the disagreement-on-the-line phenomenon occurs in our setting. 
Experiments on CIFAR-10-C, Tiny ImageNet-C, and Camelyon17 are consistent with our theory and support the universality of the theoretical findings.
\end{abstract}


\section{Introduction}
Modern machine learning methods such as
deep neural networks are effective at prediction tasks when the input test data is similar to the data used during training. 
However, they can be extremely sensitive to changes in the input data distribution (e.g., \cite{biggio2013evasion,szegedy2013intriguing,hendrycks2019augmix}, etc.). 
This is a significant concern in safety-critical applications where errors are costly (e.g., \cite{oakden2020hidden}, etc.). 
In such scenarios, it is important to  estimate how well the predictive model performs on out-of-distribution (OOD) data.

Collecting labeled data from new distributions can be costly, but unlabeled data is often readily available.
As such, recent research efforts have focused on developing methods that can estimate a predictive model's OOD performance using only unlabeled data (e.g., \cite{garg2022leveraging,deng2021labels,chen2021detecting,guillory2021predicting}, etc.).

In particular, works dating back at least to \cite{recht2019imagenet} 
suggest that the out-of-distribution (OOD) and in-distribution (ID) errors of predictive models of different complexities are highly correlated. This was rigorously proved in \cite{tripuraneni2021covariate} for random features model under covariate shift. However, determining the correlation requires labeled OOD data. 
To sidestep this requirement, \cite{baekagreement} proposed an alternative approach that looks at the \emph{disagreement}
 on an unlabeled set of data points
between pairs of neural networks with the same architecture trained with different sources of randomness. 
They observed a linear trend between ID and OOD disagreement, as for ID and OOD error. Surprisingly, the linear trend \emph{had the same empirical slope and intercept} as the linear trend between ID and OOD accuracy. 
This phenomenon, termed \emph{disagreement-on-the-line}, allows estimating the linear relationship between OOD and ID error using only unlabeled data, and finally the estimation of OOD error. 

\begin{figure}
\centering
\includegraphics{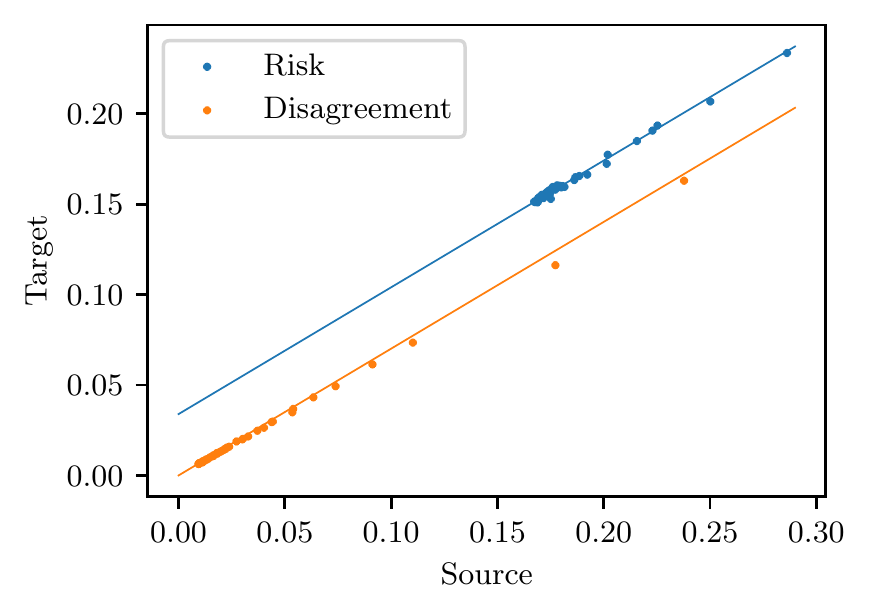}
\caption{Target vs. source risk and shared-sample disagreement of random features model trained on CIFAR-10. 
Solid lines are derived from Theorem \ref{thm:linear relation}.
Target domain is CIFAR-10-C-Fog \cite{hendrycks2018benchmarking}.  See Section \ref{sec:simulation} for details.
}
\label{fig:cifar10fog}
\end{figure}

At the moment, the theoretical basis for disagreement-on-the-line remains unclear.
It is unknown how generally it occurs, and what factors (such as the type of models or data used) may influence it.
To better understand---or even demystify---these empirical findings, in this paper, we develop a theoretical foundation for studying disagreement. We focus on the following key questions:

\begin{center}
\textit{Is disagreement-on-the-line a universal phenomenon?  
Under what conditions is it guaranteed to happen, and what happens if those conditions fail?}    
\end{center}

To work towards answering these questions, we study disagreement in 
a widely used theoretical framework for high dimensional learning,
\emph{random features models}.
We consider a setting where input data is from a Gaussian distribution, but possibly with a different covariance structure at training and test time, and study disagreement under the high-dimensional/proportional limit setting.
We define various types of disagreement depending on what randomness the two models share.
We \emph{rigorously prove that depending on the type of shared randomness and the regime of parameterization, the disagreement-on-the-line may or may not happen in random feature models} trained using ridgeless least squares.
Moreover, in contrast to prior observations, 
the line for disagreement and the line for risk may have \emph{different intercepts}, even if they share the same slope.
Additionally, we prove that adding ridge regularization  breaks the exact linear relation, but an approximate linear relation still exists. 
Thus, we find that even in a simple theoretical setting, disagreement-on-the-line is a nuanced phenomenon that can depend on the type of randomness shared, regularization, and the level of overparametrization.

Experiments we performed on CIFAR-10-C and other datasets are consistent with our theory, 
even though the assumptions of Gaussianity of inputs and linearity of the data generation are not met (Figure \ref{fig:cifar10fog}, \ref{fig:experiment}). 
This suggests that our theory is relevant beyond our theoretical setting.

\subsection{Main Contributions}
We provide an overview of the paper and our results.

\begin{itemize}
    \item We propose a framework for the theoretical study of disagreement.
    We introduce a comprehensive and unifying set of notions of disagreement (Definition \ref{def:disagreement}). 
    Then, we find a limiting formula for 
    disagreement in the high-dimensional limit where the sample size, input dimension, and feature dimension grow proportionally (Theorem \ref{thm:asymp}).

    \item Based on this characterization, we study how disagreement under source and target domains are related. We identify under what conditions and for which type of disagreement does the \emph{disagreement-on-the-line} phenomenon hold (Section \ref{sec:disagreement-on-the-line}).
    Theorem \ref{thm:approxlinear1} and Corollary \ref{cor:approxlinear2} show an approximate linear relation when the conditions are not met.

    \item When the disagreement-on-the-line holds in our model, our results imply 
    that the
    \emph{target vs. source line for risk} 
    and the 
    \emph{target vs. source line for disagreement} have the same slope.
    This is consistent with the findings of \cite{baekagreement},
    that whenever OOD vs. ID accuracy is on a line, OOD vs. ID agreement is also on the same line.
    However, unlike their finding, in our problem, the intercepts of the lines can be different (Remark \ref{rmk:linear}).

    \item In Section \ref{sec:simulation}, we conduct experiments on several datasets including CIFAR-10-C, Tiny ImageNet-C, and Camelyon17. 
    The experimental results are generally consistent with our theoretical findings, 
    even as the theoretical conditions we use (e.g., Gaussian input, linear generative model, etc.) may not hold. This suggests a possible universality of the theoretical predictions.

\end{itemize}

\subsection{Related Work}

\paragraph{Random features model.}
Random features models were introduced by \cite{RahimiRecht} as an approach for scaling kernel methods to massive datasets. 
Recently, they have been used as a standard model for the theoretical study of deep neural networks.
They are one of the simplest models that capture empirical observations such as the double descent phenomenon in well-specified models \cite{mei2022generalization, adlam2019random, lin2021causes}.
In particular, in this model, the number of parameters and the ambient dimension are disentangled, hence the effect of overparameterization can be studied on its own.

Random features models can also be analyzed in settings beyond the standard i.i.d. data model. \cite{tripuraneni2021covariate}
studied the random features model under covariate shift and derived precise asymptotic limits of the risk in the proportional limit.

\paragraph{Linear relation under distribution shift.}
Several intriguing phenomena have been observed in empirical studies of distribution shift. \cite{recht2019imagenet,hendrycks2021many,koh2021wilds,taori2020measuring,miller2021accuracy} observed linear trends between OOD and ID test error. \cite{tripuraneni2021covariate} proved this phenomenon in random features models under covariate shift.

Recently, the notion of disagreement has been gaining a lot of attention (e.g., \cite{hacohen2020let,chen2021detecting,jiang2021assessing,nakkiran2020distributional,baekagreement,atanov2022task,pliushch2022deep}, etc.). 
In particular, \cite{baekagreement} empirically showed that OOD agreement between the predictions of pairs of neural networks also has a strong linear correlation with their ID agreement. 
They further observed that the slope and intercept of the OOD vs ID agreement line closely match that of the accuracy. 
This can be used to predict OOD performance of predictive models only using unlabeled data.

\paragraph{High-dimensional asymptotics.}
Work on high-dimensional asymptotics dates back at least to the 1960s
\cite{raudys1967determining,deev1970representation,raudys1972amount}
and has more recently been studied in a wide range of areas, 
such as 
high-dimensional statistics (e.g., \cite{raudys2004results,serdobolskii2007multiparametric,paul2014random,yao2015large,dobriban2018high}, etc.),
wireless communications  (e.g., \cite{tulino2004random,couillet2011random},  etc.),
and machine learning  (e.g., \cite{gyorgyi1990statistical,opper1995statistical,opper1996statistical,couillet2022random,engel2001statistical},  etc.).

\paragraph{Technical tools.} 
The results derived in this paper rely on the Gaussian equivalence conjecture studied and used extensively for random features model (e.g., \cite{goldt2022gaussian,hu2022universality, montanari2022universality,mei2022generalization, hassani2022curse, tripuraneni2021covariate, loureiro2021learning, d2021interplay}, etc.). Our analytical results build upon the series of recent work \cite{mel2021anisotropic,adlam2020neural,tripuraneni2021covariate}
using random matrix theory and operator-valued free probability \cite{far2008slow,mingo2017free}.

\section{Preliminaries}\label{sec:prelim}
\subsection{Problem Setting}

We study a supervised learning setting where the training data $(x_i, y_i) \in \R^{d} \times \R$, $i \in [n]$, of dimension $d$ and sample size $n$, is generated according to
\begin{align}\label{eqn:datagen}
x_i \stackrel{\text{i.i.d.}}{\sim} \normal(0, \Sigma_{\rm s}), \text{ and } y_i = \frac{1}{\sqrt{d}} \beta^\top x_i + \ep_i,
\end{align}
where $ \ep_i \stackrel{\text{i.i.d.}}{\sim} \normal(0, \sigma_\ep^2).$ Additionally, the true coefficient $\beta \in \mathbb{R}^{d}$ is assumed to be randomly drawn from $\normal(0, I_{d})$.
The linear relationship between $(x_i,y_i)$ is not known. We fit a model to the data, which can then be used to predict labels for unlabeled examples at test time.

We consider two-layer neural networks with fixed, randomly generated weights in the first layer---a random features model---as the learner.
We let the width of the internal layer be $N \in \mathbb{N}$.
For 
a weight matrix $W \in \R^{N \times d}$ with i.i.d. random  entries sampled from $\normal(0, 1)$, an activation function $\sigma: \R \to \R$ applied elementwise, and the weights $a \in \R^{N}$ of a linear layer, 
the random features model is defined by
\begin{equation*}
    f_{W,a}(x) = \frac{1}{\sqrt{N}}a^\top \sigma\left(W x / \sqrt{d}\right).
\end{equation*}
The trainable parameters $a \in \R^{N}$ are fit via ridge regression to the training data $X = (x_1, \dots, x_n) \in \R^{d \times n}$ and $Y = (y_1, \dots, y_n)^\top \in \R^n$.
Specifically, for a regularization parameter $\gamma > 0$, we solve
\begin{align*}
    \hat{a} 
    &= \argmin_{a \in \R^{N}} \left\Vert Y - \sigma\left(WX / \sqrt{d} \right)^\top a / \sqrt{N} \right\Vert_2^2 + \gamma \Vert a \Vert_2^2,
\end{align*}
and use $\hat{y}(x) = \hat{a}^\top \sigma(W x / \sqrt{d}) / \sqrt{N}$ as the model prediction for a data point $x \in \R^d$.
Defining $F = \sigma(WX / \sqrt{d})$ and $f = \sigma(W x / \sqrt{d})$, we can write
\begin{align}\label{eqn:yhat}
    \hat{y}(x) = Y^\top \left( \frac{1}{N} F^\top F + \gamma I_n \right)^{-1} \left( \frac{1}{N} F^\top f \right).
\end{align}
To emphasize the dependence on $W, X, Y$, we also use the notation $\hat{y}_{W, X, Y}$.

It has been recognized in e.g., \cite{adlam2020neural, ghorbani2021linearized, mei2022generalization} that only linear data generative models can be learned in the proportional-limit high-dimensional regime by random features models, and the non-linear part behaves like an additive noise.
Thus, we consider linear generative models as in \eqref{eqn:datagen}. Results for non-linear models can be obtained via linearization, as is standard in the above work.

We also highlight that our theoretical findings are validated by simulations on standard datasets (such as CIFAR-10-C) whose data generation model is non-linear.

\subsection{Distribution Shift}
At training time \eqref{eqn:datagen}, the inputs $x_i$ are sampled from the \emph{source domain}, $\cD_{\rm s} = \normal(0, \Sigma_{\rm s})$.
At test time, we assume the input distribution shifts to the \emph{target domain}, $\cD_{\rm t} = \normal(0, \Sigma_{\rm t})$.
We do not restrict the change in $\mathbb{P}(y | x)$ since disagreement is independent of the label $y$.
Previous work \cite{lei2021near, tripuraneni2021covariate, wu2022power} found that the learning problem under covariate shift is fully characterized by input covariance matrices.
For this reason, we do not consider shifts in the mean of the input distribution.

\subsection{Definition of Disagreement}
\cite{hacohen2020let,chen2021detecting,jiang2021assessing,nakkiran2020distributional,baekagreement} define notions of \emph{disagreement} (or \emph{agreement}) to quantify the difference (or similarity) between the predictions of two randomly trained predictive models in \emph{classification} tasks.

Prior work on disagreement considers three sources of randomness that lead to different predictive models: (i) random initialization, (ii) sampling of the training set, and (iii) sampling/ordering of mini-batches.

Motivated by these results, we propose analogous notions of disagreement in random features regression.
We consider (i), (ii) and their combination, as (iii) is not present in our problem. 
The \textit{independent disagreement} measures how much the prediction of two models with independent random weights and trained on two independent set of training samples disagree, on average.
The \textit{shared-sample disagreement} measures the average disagreement of two models with independent random weights, but trained on a shared training set.
The \textit{shared-weight disagreement} measures the average disagreement of two models with shared random weights, but trained on two independent training samples.

While the prior work typically used 0-1 loss to define agreement/disagreement in classification, we use the squared loss to measure disagreement of real-valued outputs.

\begin{definition}[Disagreement]\label{def:disagreement}
Consider two random features models trained on the data $(X_1, Y_1), (X_2, Y_2) \in \R^{d \times n} \times \R^n$ with random weight matrices $W_1, W_2 \in \R^{N \times d}$, respectively.
We measure the disagreement of two models by their mean squared difference
\begin{align*}
        \stsd_i^j (&n, d, N, \gamma) = \E \left[\left(\hat{y}_{W_1, X_1, Y_1}(x) - \hat{y}_{W_2, X_2, Y_2}(x)\right)^2\right],
    \end{align*}
where the expectation is over $\beta, W_1, W_2, X_1, Y_1 X_2, Y_2$, and $j \in \{\rm s, \rm t\}$ is the domain that $x \sim \cD_{j}$ is from, and the index $i \in \{\textnormal{I}, \textnormal{SS}, \textnormal{SW} \}$ corresponds to one of the following cases.
\begin{itemize}
    \item \textbf{I}ndependent disagreement ($i = \textnormal{I}$): the training data $(X_1, Y_1), (X_2, Y_2)$ are independently generated from \eqref{eqn:datagen}, with the same $\beta$.
    The weights $W_1, W_2 \in \mathbb{R}^{N \times d}$ are independent matrices with i.i.d. $\normal(0,1)$ entries.
    \item \textbf{S}hared-\textbf{S}ample disagreement ($i = \textnormal{SS}$): the training samples are shared, i.e., $(X_1, Y_1) = (X_2, Y_2) = (X, Y)$, where $(X, Y)$ is generated from \eqref{eqn:datagen}.
    The weights $W_1, W_2 \in \mathbb{R}^{N \times d}$ are independent matrices with i.i.d. $\normal(0,1)$ entries.

    \item \textbf{S}hared-\textbf{W}eight disagreement ($i = \textnormal{SW}$): the training data $(X_1, Y_1), (X_2, Y_2)$ are independently generated from \eqref{eqn:datagen}, with the same $\beta$.
    Two models share the weights, i.e., $W_1 = W_2 = W$.
    The weights are shared, i.e., $W_1 = W_2 = W$, where $W \in \mathbb{R}^{N \times d}$ is a matrix with i.i.d. $\normal(0, 1)$ entries.
\end{itemize}

\end{definition}

\subsection{Conditions}
We characterize the asymptotics of disagreement in the proportional limit asymptotic regime defined as follows.

\begin{condition}[Asymptotic setting]
\label{cond:limit}
    We assume that $n, d, N \to \infty$ with $d / n \to \phi >0$ and $d / N \to \psi>0$.
\end{condition}

To characterize the limit of disagreement, 
we need conditions on the spectral properties of $\Sigma_{\rm s}$ and $\Sigma_{\rm t}$ as their dimension $d$ grows.
When multiple growing matrices are involved, it is not sufficient to make assumptions on the individual spectra of the matrices, but rather, they have to be considered \emph{jointly} \cite{wu2020optimal, tripuraneni2021covariate, mel2021anisotropic}.
We assume that the \emph{joint spectral distribution} of $\Sigma_{\rm s}$ and $\Sigma_{\rm t}$ converges to a limiting distribution $\mu$ on $\R_+^2$ as $d \to \infty$.

\begin{condition}\label{cond:jointspectral}
Let $\lambda_{1}^{\rm s}, \dots, \lambda_{d}^{\rm s} \ge 0$ be the eigenvalues of $\Sigma_{\rm s}$ and $v_1, \dots, v_{d}$ be the corresponding eigenvectors.
Define $\lambda_i^{\rm t} = v_i^\top \Sigma_{\rm t} v_i$ for $i \in [d]$.
We assume the joint empirical spectral distribution of $(\lambda_i^{\rm s}, \lambda_i^{\rm t}), i \in [d]$ converges in distribution to a limiting distribution $\mu$ on $\R_+^2$.
That is,
\begin{align*}
    \frac{1}{d} \sum_{i = 1}^{d} \delta_{(\lambda_i^{\rm s}, \lambda_i^{\rm t})} \to \mu,
\end{align*}
where $\delta$ is the Dirac delta measure.
We additionally assume that $\mu$ has a compact support.
We denote random variables drawn from $\mu$ by $(\lambda^{\rm s}, \lambda^{\rm t})$, and write $m_{\rm s} = \E_\mu[\lambda^{\rm s}]$ and $m_{\rm t} = \E_\mu[\lambda^{\rm t}]$.
\end{condition}

For the existence of certain derivatives and expectations, 
we assume the following mild condition on the activation function $\sigma: \R \to \R$.
\begin{condition}\label{cond:sigmabound}
    The activation function $\sigma: \R \to \R$ is differentiable almost everywhere.
    There are constants $c_0$ and $c_1$ such that $|\sigma(x)|, |\sigma'(x)| \leq c_0 e^{c_1x}$, whenever $\sigma'(x)$ exists.
    For $j \in \{\rm s, \rm t\}$ and a standard Gaussian random variable $Z \sim \normal(0, 1)$, define
\begin{align}\label{eqn:defconstants}
    &\rho_j =  \frac{\mathbb{E}[Z \sigma(\sqrt{m_j} Z)]^2}{m_j}, \,\omega_j = \frac{\mathbb{V}[\sigma(\sqrt{m_j}Z)]}{\rho_j} - m_j.
\end{align}
\end{condition}

These constants characterize the non-linearity of the activation $\sigma$ and will appear in the asymptotics of disagreement. 
Note that when $\sigma$ is ReLU activation $\sigma(x) = \max(x, 0)$, we have $\rho_{j} = 1/4$, $\omega_j = m_j (1 - 2/\pi)$ for $j \in \{\rm s, \rm t\}$.

\section{Asymptotics of Disagreement}
 In this section, we present our results on characterizing the limits of disagreement defined in Definition \ref{def:disagreement} for random features models.
 We introduce results for general ridge regression and also study the \emph{ridgeless limit} $\gamma \to 0$.

\subsection{Ridge Setting}
For $i \in \{\textnormal{I}, \textnormal{SS}, \textnormal{SW} \}$ and $j \in \{\rm s, \rm t\}$, define
the \emph{asymptotic disagreement}
\begin{align*}
    \stsd_i^j (\phi, \psi, \gamma)= \lim_{n, d, N \to \infty} \stsd_i^j(n,d,N, \gamma),
\end{align*}
where the limit is in the regime considered in Condition~\ref{cond:limit}.

Asymptotics in random features models (e.g., training/test error, bias, variance, etc.) typically do not have a closed form, 
and can only be  implicitly described through \emph{self-consistent equations} \cite{adlam2019random, mei2022generalization, hastie2022surprises}.
To facilitate analysis of these implicit quantities, previous work (e.g., \cite{dobriban2021distributed,dobriban2020wonder,tripuraneni2021covariate, mel2021anisotropic}, etc.)
proposed using expressions containing \emph{only one implicit scalar}.
We show that similar to the asymptotic risk derived in \cite{tripuraneni2021covariate}, the asymptotic disagreements can be expressed using a scalar $\kappa$ which is the unique non-negative solution of the self-consistent equation
\begin{align}\label{eqn:defkappa}
    \kappa = 
    \frac{\psi + \phi - 
    \sqrt{(\psi - \phi)^2 + 4 \kappa \psi \phi \gamma / \rho_{\rm s}}}{2\psi (\omega_{\rm s} + \mathcal{I}_{1, 1}^{\rm s}(\kappa))},
\end{align}
where $\cI_{a, b}^{j}$ is the \emph{integral functional} of $\mu$ defined by
\begin{align}
\label{eqn:deffunctionals}
    &\mathcal{I}_{a, b}^{j}(\kappa) =\phi \E_{\mu} \left[ \frac{ (\lambda^{\rm s})^{a - 1} \lambda^{j}}{(\phi + \kappa \lambda^{\rm s})^b} \right], \quad j \in \{\rm s, \rm t\}.
\end{align}
We omit $\kappa$ and simply write $\cI_{a, b}^j$ whenever the argument is clear from the context. 
Recall from Condition \ref{cond:jointspectral} that $\mu$ describes the joint spectral properties of source and target covariance matrices, so $\cI_{a, b}^j$ can be viewed as a summary of the joint spectral properties.

The following theorem---our first main result---shows that $\stsd_\textnormal{I}^j (\phi, \psi, \gamma)$, $\stsd_\textnormal{SS}^j (\phi, \psi, \gamma)$, $\stsd_\textnormal{SW}^j (\phi, \psi, \gamma)$ are well defined, and characterizes them.

\begin{theorem}[Disagreement in general ridge regression]\label{thm:asymp}
For  $j \in \{\rm s, \rm t\}$, the asymptotic independent disagreement is
\begin{align*}
    &\stsd_\textnormal{I}^j (\phi, \psi, \gamma)
    =  \frac{2\rho_{j} \psi \kappa}{\phi\gamma + \rho_{\rm s} \gamma(\tau \psi +  \bar\tau \phi)(\omega_{\rm s} + \phi \mathcal{I}^{\rm s}_{1, 2})} \Big[ \gamma \tau  (\omega_{j} + \phi \mathcal{I}_{1, 2}^{j}) \mathcal{I}_{2, 2}^{\rm s}  \\
    &\hspace{4cm}+ (\sigma_\ep^2 + \mathcal{I}_{1, 1}^{\rm s})(\omega_{\rm s} + \phi \mathcal{I}_{1, 2}^{\rm s})(\omega_{j} + \mathcal{I}_{1, 1}^{j})
     +  \frac{\phi}{\psi} \gamma \bar\tau (\sigma_\ep^2 + \phi \mathcal{I}_{1, 2}^{\rm s}) \mathcal{I}_{2, 2}^{j}\Big], 
\end{align*}
and the asymptotic shared-sample disagreement is
\begin{align*}
    \stsd_\textnormal{SS}^j &(\phi, \psi, \gamma)
    = \stsd_\textnormal{I}^j (\phi, \psi, \gamma) -  \frac{2\rho_j\kappa^2(\sigma_\ep^2 + \phi \cI^{\rm s}_{1, 2}) \cI^j_{2, 2}}{\rho_{\rm s}(1 - \kappa^2 \cI_{2, 2}^{\rm s})},
\end{align*}
and the asymptotic shared-weight disagreement is
\begin{align*}
    \stsd_\textnormal{SW}^j &(\phi, \psi, \gamma)
    = \stsd_\textnormal{I}^j (\phi, \psi, \gamma) -  \frac{2\rho_j \psi \kappa^2(\omega_j + \phi \cI^{j}_{1, 2}) \cI^{\rm s}_{2, 2}}{\rho_{\rm s}(\phi - \psi \kappa^2 \cI_{2, 2}^{\rm s})},
\end{align*}
where $\tau$ and $\bar{\tau}$ are the limiting normalized trace of $(F^\top F / N + \gamma I_n)^{-1}$ and $(FF^\top /N + \gamma I_N)^{-1}$, respectively.
 They can be expressed as functions of $\kappa$ as follows:
\begin{align}\label{eqn:deftau}
    \tau &= \frac{\sqrt{(\psi - \phi)^2 + 4\kappa \psi \phi \gamma / \rho_{\rm s}} + \psi - \phi}{2 \psi \gamma}, 
    \quad\bar\tau = \frac{1}{\gamma} + \frac{\psi}{\phi}\left( \tau - \frac{1}{\gamma} \right).
\end{align}
\end{theorem}

The expressions in Theorem \ref{thm:asymp} are written in terms of the non-linearity constants $\rho_{\rm s}, \rho_{\rm t}, \omega_{\rm s}, \omega_{\rm t}$, the dimension parameters $\psi, \phi$, the regularization $\gamma$, the noise level $\sigma_\ep^2$, the summary statistics $\cI_{a, b}^{\rm s}, \cI_{a, b}^{\rm t}$ of $\mu$, and $\tau, \bar \tau, \kappa$.
Since $\tau, \bar\tau$ are algebraic functions of $\kappa$, the expressions are functions of \emph{one implicit variable} $\kappa$.

\begin{figure}[h!]
\centering
\includegraphics{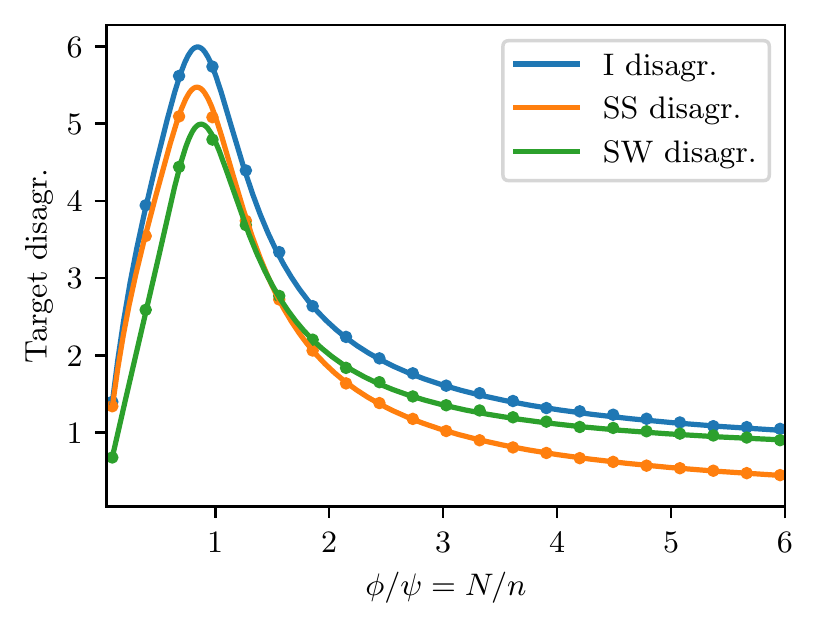}

\caption{Independent, shared-sample, and shared-weight disagreement under target domain in random features regression with ReLU activation function, $\phi = \lim d / n = 0.5$, versus $\phi/\psi = \lim N/n$.  We set $\gamma = 0.01$, $\sigma_\epsilon^2 = 0.25$, and $\mu = 0.5 \delta_{(1.5,5)} + 0.5 \delta_{(1,1)}$. Simulations are done with $d = 512$, $n = 1024$, and averaged over 300 trials. The continuous lines are theoretical predictions from Theorem \ref{thm:asymp}, and the dots are simulation results.}
\label{fig:asymptotics_vs_simulation} 
\end{figure}

This theorem can be used to make numerical predictions for disagreement.
To do so, we first solve the self-consistent equation \eqref{eqn:defkappa} using a fixed-point iteration and find $\kappa$. Then, we plug $\kappa$ into the terms appearing in the theorem. Figure \ref{fig:asymptotics_vs_simulation} shows an example, supporting that the theoretical predictions of Theorem \ref{thm:asymp} match very well with simulations even for moderately large $d, n, N$.

\paragraph{Theoretical Innovations.}
To prove this theorem, we first rely on Gaussian equivalence (Section \ref{sec:centering}, \ref{sec:gaussianequivalents}) to express disagreement as a combination of traces of rational functions of i.i.d. Gaussian matrices.  Then, we construct linear pencils (Section \ref{sec:linearpencil}) and use the theory of operator-valued free probability (Section \ref{sec:operator}, \ref{sec:gaussianR}) to derive the limit of these trace objects. This general strategy has been used previously in \cite{adlam2019random, adlam2020understanding, tripuraneni2021covariate, mel2021anisotropic}. However, in the expressions of disagreement, new traces appear that did not exist in prior work. We construct new suitable linear pencils to derive the limit of these traces.  While this leads to a coupled system of self-consistent equations of many variables, it turns out that they can be factored into a single scalar variable $\kappa$ defined through the self-consistent equation \eqref{eqn:defkappa}, and every term appearing in the limiting disagreements, can be written as algebraic functions of $\kappa$. These results might also be of independent interest. The fact that limiting disagreements only rely on the same implicit variable as the variable appearing in the limiting risk, enables us to derive the results in Section \ref{sec:disagreement-on-the-line}.

\vspace{-0.1cm}

\subsection{Ridgeless Limit}
In the ridgeless limit $\gamma \to 0$, the self-consistent equation \eqref{eqn:defkappa} for $\kappa$ becomes
\vspace{-0.1cm}
\begin{align}\label{eqn:defkapparidgeless}
    \kappa = \frac{\min(1, \phi / \psi)}{\omega_{\rm s} + \mathcal{I}_{1, 1}^{\rm s}(\kappa)}.
\end{align}
Further, the asymptotic limits in Theorem \ref{thm:asymp} can be simplified as follows.

\begin{corollary}[Ridgeless limit]\label{cor:ridgeless}
    For  $j \in \{\rm s, \rm t\}$ and in the ridgeless limit $\gamma \to 0$, the asymptotic independent disagreement is
    \begin{align*}
        \lim_{\gamma \to 0} \stsd_\textnormal{I}^j (\phi, \psi, \gamma) &= \frac{2 \rho_j \psi \kappa}{\rho_{\rm s}|\phi - \psi|} (\sigma_\ep^2 + \mathcal{I}_{1, 1}^{\rm s})(\omega_{j} + \mathcal{I}_{1, 1}^{j})
         +  \begin{cases} \frac{2\rho_j\kappa(\sigma_\ep^2 + \phi \cI^{\rm s}_{1, 2}) \cI^j_{2, 2}}{\rho_{\rm s}(\omega_{\rm s} + \phi \cI^{\rm s}_{1, 2})} & \phi > \psi, \\ \frac{2\rho_j \kappa(\omega_j + \phi \cI_{1, 2}^j) \cI_{2, 2}^{\rm s}}{\rho_{\rm s}(\omega_{\rm s} + \phi \cI_{1, 2}^{\rm s})}  & \phi < \psi, \end{cases}
    \end{align*}
    and the asymptotic shared-sample disagreement is
    \begin{align*}
        \lim_{\gamma \to 0} \stsd_\textnormal{SS}^j (&\phi, \psi, \gamma)\\ &= \frac{2 \rho_j \psi \kappa}{\rho_{\rm s}|\phi - \psi|} (\sigma_\ep^2 + \mathcal{I}_{1, 1}^{\rm s})(\omega_{j} + \mathcal{I}_{1, 1}^{j})
        +  \begin{cases} 0 & \phi > \psi, \\ \frac{2\rho_j \kappa}{\rho_{\rm s}} \left(\frac{(\omega_j + \phi \cI_{1, 2}^j) \cI_{2, 2}^{\rm s}}{\omega_{\rm s} + \phi \cI_{1, 2}^{\rm s}} - \frac{\kappa(\sigma_\ep^2 + \phi \cI^{\rm s}_{1, 2}) \cI^j_{2, 2}}{1 - \kappa^2 \cI_{2, 2}^{\rm s}} \right)& \phi < \psi, \end{cases}
    \end{align*}
    and the asymptotic shared-weight disagreement is
    \begin{align*}
        \lim_{\gamma \to 0} \stsd_\textnormal{SW}^j (&\phi, \psi, \gamma)\\ &= \frac{2 \rho_j \psi \kappa}{\rho_{\rm s}|\phi - \psi|} (\sigma_\ep^2 + \mathcal{I}_{1, 1}^{\rm s})(\omega_{j} + \mathcal{I}_{1, 1}^{j})+  \begin{cases} \frac{2\rho_j \kappa}{\rho_{\rm s}} \left( \frac{(\sigma_\ep^2 + \phi \cI_{1, 2}^{\rm s}) \cI_{2, 2}^j}{\omega_{\rm s} + \phi \cI_{1, 2}^{\rm s}} - \frac{\psi \kappa (\omega_j + \phi \cI_{1, 2}^j) \cI_{2, 2}^{\rm s}}{\phi - \psi \kappa^2 \cI_{2, 2}^{\rm s}} \right) & \phi > \psi, \\ 0 & \phi < \psi, \end{cases}
    \end{align*}
    where $\kappa$ is defined in \eqref{eqn:defkapparidgeless}.
\end{corollary}
In the ridgeless limit, I and SS disagreement have a \emph{single term} that depends on $\psi$, which motivates the analysis in Section \ref{sec:disagreement-on-the-line} that examines the \emph{disagreement-on-the-line} phenomenon.
In contrast, SW disagreement has two linearly independent terms that are functions of $\psi$, leading to a distinct behavior compared to I and SS disagreement.

The asymptotics in Corollary \ref{cor:ridgeless} reveal other interesting phenomenon regarding disagreements of random features model in the ridgeless limit.
For example, it follows from Corollary \ref{cor:ridgeless} that SS disagreement
tends to zero in the \emph{infinite overparameterization} limit
where the width $N$ of the internal layer is much larger than the data dimension $d$, so that   $\psi = \lim d/N \to 0$.
However, the same is not true for I and SW disagreement.
This indicates that, in the infinite overparameterization limit, the randomness caused by the random weights disappears, and the model is solely determined by the training sample.

\section{When Does Disagreement-on-the-Line Hold?}
\label{sec:disagreement-on-the-line}
In this section, based on the characterizations of disagreements derived in the previous section, we study for which types of disagreement and under what conditions, the \emph{linear relationship} between disagreement under source and target domain of models of varying complexity holds.

\subsection{I and SS disagreement}
\paragraph{Ridgeless.} In the overparameterized regime $\phi > \psi$, the self-consistent equation \eqref{eqn:defkapparidgeless} is independent of $\psi = \lim d/N$, and so is $\kappa$. 
This implies the following linear trend of I and SS disagreement, in the ridgeless limit.

\begin{theorem}[Exact linear relation]
\label{thm:linear relation}
Define
\begin{align}\label{eqn:slopeintercept}
&a = \frac{\rho_{\rm t}(\omega_{\rm t} + \cI_{1, 1}^{\rm t})}{\rho_{\rm s}(\omega_{\rm s} + \cI_{1, 1}^{\rm s})}, \quad b_{\textnormal{SS}} = 0,\quad  b_{\textnormal{I}} = \frac{2\kappa^2(\sigma_\ep^2 + \phi \cI_{1, 2}^{\rm s})(\rho_{\rm t} \cI_{2, 2}^{\rm t} - a \rho_{\rm s} \cI_{2, 2}^{\rm s})}{\rho_{\rm s}(1 - \kappa^2 \cI_{2, 2}^{\rm s})},
\end{align}
for $\kappa$ satisfying \eqref{eqn:defkapparidgeless}.
We fix $\phi$ and regard the disagreement $\stsd_i^j (\phi, \psi, \gamma)$, $i \in \{\textnormal{I}, \textnormal{SS} \}$, $j \in \{\rm s, \rm t\}$, as a function of $\psi$.
In the overparameterized regime $\phi > \psi$ and for $i \in \{\textnormal{I}, \textnormal{SS}\}$,
\begin{align}\label{eqn:disontheline}
&\lim_{\gamma \to 0} \stsd_i^{\rm t} (\phi, \psi, \gamma) =a  \lim_{\gamma \to 0} \stsd_i^{\rm s} (\phi, \psi, \gamma) + b_i,
\end{align}
where the slope $a$ and the intercept $b_\textnormal{I}$ are independent of $\psi$.
\end{theorem}

Recall from \eqref{eqn:defconstants} and \eqref{eqn:deffunctionals} that $\rho_{\rm s}, \rho_{\rm t}, \omega_{\rm s}, \omega_{\rm s}$ are constants describing non-linearity of the activation $\sigma$, and $\cI_{a, b}^{\rm s}, \cI_{a, b}^{\rm t}$ are statistics summarizing spectra of $\Sigma_{\rm s}, \Sigma_{\rm t}$.
Therefore, the slope $a$ is determined by the property of $\sigma, \Sigma_{\rm s}, \Sigma_{\rm t}$.
By plugging in sample covariance, we can build an estimate of the slope in finite-sample settings.
Also as a sanity check, if we set $\Sigma_{\rm s} = \Sigma_{\rm t}$, then we recover $a = 1$ and $b_{\textnormal{I}} = 0$ as there will be no difference between source and target domain.

\begin{figure*}
\centering
\includegraphics[scale=0.85]{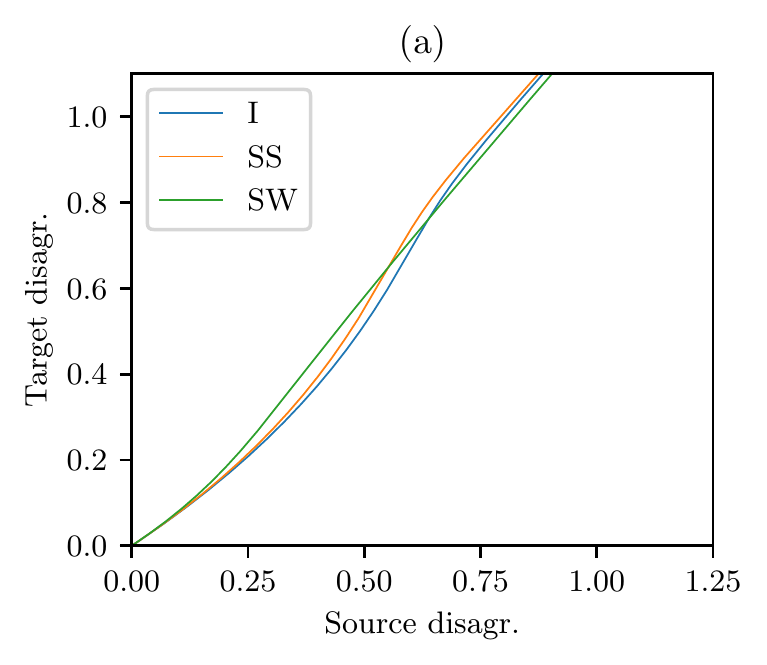}
\hspace{-10pt}
\includegraphics[scale=0.85]{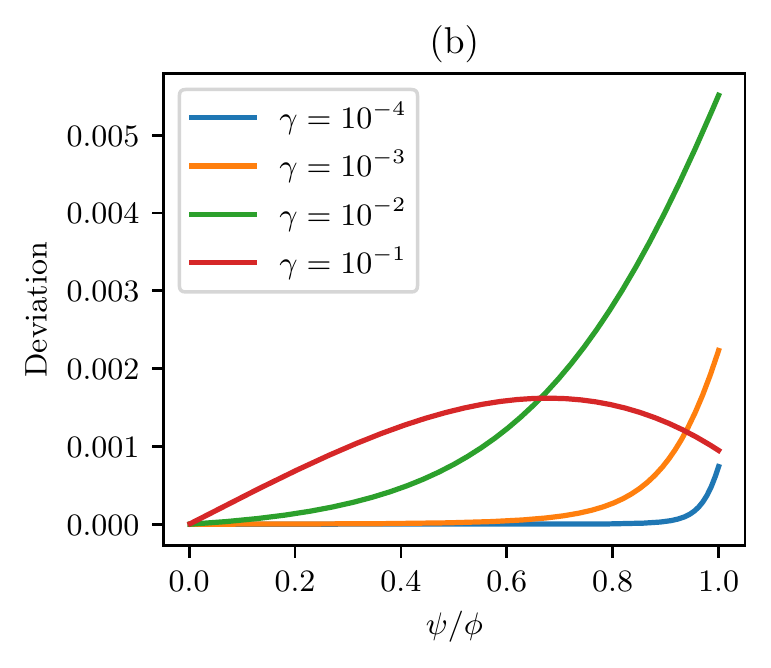}
\hspace{-11pt}\\
\includegraphics[scale=0.85]{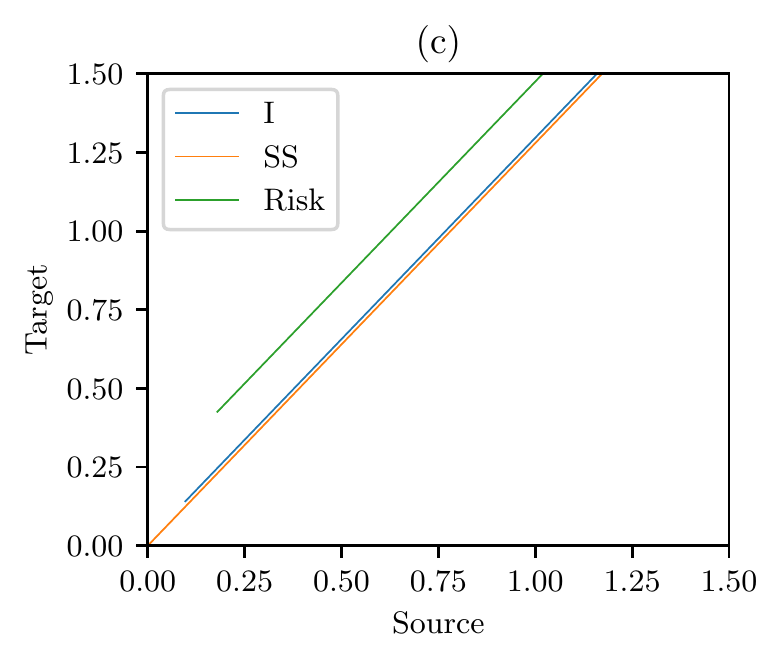}
\hspace{-15pt}
\includegraphics[scale=0.85]{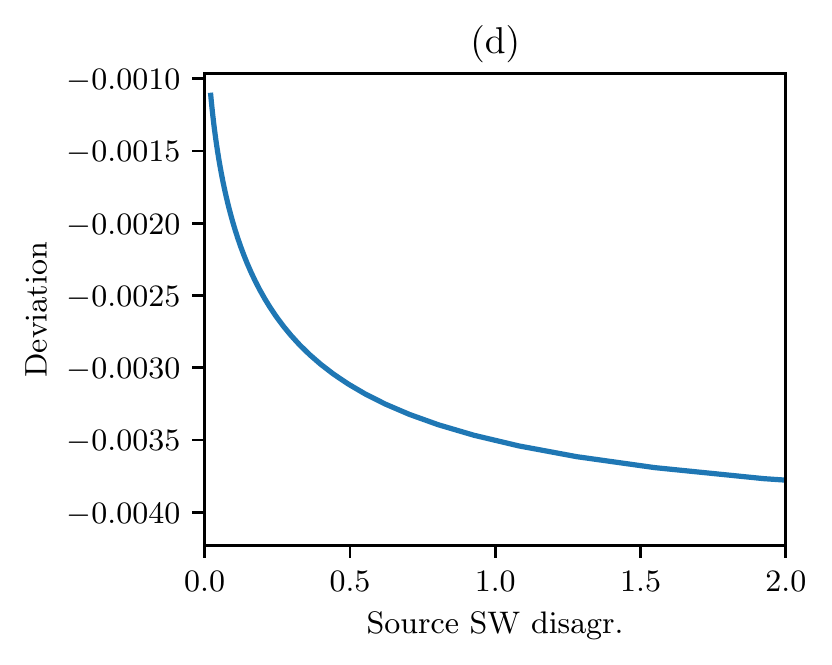}
\vspace{-15pt}
\caption{\textbf{(a)} Target vs. source I, SS, SW disagreement in the ridgeless and underparameterized regime ($\phi < \psi$).
There is no linear trend in this regime. \textbf{(b)} Deviation from the line,  $\stsd_{\textnormal{SS}}^{\rm t}(\phi, \psi, \gamma) - a \stsd_{\textnormal{SS}}^{\rm s}(\phi, \psi, \gamma)$, as a function of $\psi$ for non-zero $\gamma$. 
The deviation becomes larger as $\gamma$ increases. See Section \ref{sec:deviation} for figures for I disagreement and risk.
\textbf{(c)} Target vs. source lines for I, SS disagreement and risk, in the overparameterized regime $\psi / \phi \in (0, 1)$. 
The lines have \emph{identical} slopes but different \emph{intercepts}.
\textbf{(d)} Deviation from the line, $ \lim_{\gamma \to 0} \stsd_{\textnormal{SW}}^{\rm t}(\phi, \psi, \gamma) - a \stsd_{\textnormal{SW}}^{\rm s}(\phi, \psi, \gamma)$, vs. $\lim_{\gamma \to 0} \stsd_{\textnormal{SW}}^{\rm s}(\phi, \psi, \gamma)$, in the overparameterized regime ($\phi > \psi$).
This shows disagreement-on-the-line does not happen for SW disagreement.
We use $\phi = 0.5$, $\sigma_\ep^2 = 10^{-4}$, and ReLU activation $\sigma$. 
We set $\mu = 0.4 \delta_{(0.1, 1)} + 0.6 \delta_{(1, 0.1)}$ in (a), (b), (d) and $\mu = 0.5 \delta_{(4, 1)} + 0.5 \delta_{(1, 4)}$ in (c).}
\label{fig:linear}
\end{figure*}

\begin{remark}\label{rmk:linear}
The slope $a = \rho_{\rm t}(\omega_{\rm t} + \cI_{1, 1}^{\rm t}) / \rho_{\rm s}(\omega_{\rm s} + \cI_{1, 1}^{\rm s})$ is same as the slope from Proposition \ref{prop:linearrisk}.
This is consistent with the empirical observations from \cite{baekagreement} that the linear trend between ID disagreement and OOD disagreement has the \textit{same slope} as the linear trend between ID risk and OOD risk. However, unlike in \cite{baekagreement}, in our case, the intercepts can be different. This can be seen in Figure \ref{fig:cifar10fog} and Figure \ref{fig:linear} \textbf{(c)}, and also from \eqref{eqn:brisk}.
\end{remark}

\paragraph{Ridge.} When $\gamma > 0$, the exact linear relation between source disagreement and target disagreement no longer holds in our model.
However, it turns out that there is still an \emph{approximate linear relation}, as we show next.
\begin{theorem}[Approximate linear relation of disagreement]\label{thm:approxlinear1}
Let $a, b_{\textnormal{SS}}, b_{\textnormal{I}}$ be defined as in \eqref{eqn:slopeintercept}.
Given $\phi > \psi$, deviation from the line, for I and SS disagreement, is bounded by 
\begin{align*}
&|\stsd_\textnormal{I}^{\rm t}(\phi, \psi, \gamma) - a\stsd_\textnormal{I}^{\rm s}(\phi, \psi, \gamma)| \leq
\frac{C(\gamma + \sqrt{\psi \gamma} + \psi \gamma + \gamma \sqrt{\psi \gamma})}{(1 - \psi / \phi + \sqrt{\psi \gamma})^2},
\end{align*}
and
\begin{align*}
|\stsd_\textnormal{SS}^{\rm t}(\phi, \psi, \gamma) - a\stsd_\textnormal{SS}^{\rm s}(\phi, \psi, \gamma)| \leq
\frac{C(\sqrt{\psi \gamma} + \psi \gamma + \gamma \sqrt{\psi \gamma})}{(1 - \psi / \phi + \sqrt{\psi \gamma})^2},
\end{align*}
where $C > 0$ depends on $\phi, \mu, \sigma_\ep^2$, and $\sigma$.
\end{theorem}
We see the upper bounds vanish as $\gamma \to 0$, consistent with Theorem \ref{thm:linear relation}.
Also, the upper bound for SS disagreement vanishes as $\psi \to 0$, which is confirmed in Figure \ref{fig:linear} \textbf{(b)}.

We now present an analog of Theorem \ref{thm:approxlinear1} for prediction error of random features model.
This is a generalization of Proposition \ref{prop:linearrisk}, which shows an exact linear relation between risks in ridgeless and overparameterized regime.

\begin{corollary}[Approximate linear relation of risk]\label{cor:approxlinear2}
Denote prediction risk in the source and target domains by $E_{\rm s}, E_{\rm t}$, respectively (see Section \ref{sec:recap} for definitions).
Let $a, b_{\textnormal{risk}}$ be defined as in \eqref{eqn:slopeintercept} and \eqref{eqn:brisk}.
Given $\phi > \psi$, deviation from the line, for risk, is bounded by 
\begin{align*}
&|E_{\rm t} -a E_{\rm s} - b_{\textnormal{risk}}| \leq \frac{C(\gamma + \sqrt{\psi \gamma} + \psi \gamma + \gamma \sqrt{\psi \gamma} + \psi \gamma^2)}{(1 - \psi / \phi + \sqrt{\psi \gamma})^2},
\end{align*}
where $C > 0$ depends on $\phi, \mu, \sigma_\ep^2$, and $\sigma$.
\end{corollary}
Theorem \ref{thm:approxlinear1} and Corollary \ref{cor:approxlinear2} together show that the phenomenon we discussed in Remark \ref{rmk:linear} occurs, at least approximately, even when applying ridge regularization.

In the underparameterized case $\psi > \phi$, the self-consistent equation \eqref{eqn:defkapparidgeless} is dependent on $\psi$, and so is $\kappa$.
Hence, there is no analog of the linear relation we find in Theorem \ref{thm:linear relation} in this regime.
Figure \ref{fig:linear} \textbf{(a)} displays this phenomenon.

\begin{table}
\centering
\caption{Existence of disagreement-on-the-line in the overparameterized regime for different regularization and types of disagreement. The symbols {\color{green}{\ding{51}}}, {\color{yellow}{\ding{115}}}, {\color{red}{\ding{55}}} correspond to exact, approximate, no linear relation, respectively.}
\vspace{5pt}
\begin{tabular}{ccc} 
\toprule

               & $\stsd_{\textnormal{I}}$ and $\stsd_{\textnormal{SS}}$ & $\stsd_{\textnormal{SW}}$                    \\ 
\hline\vspace{-0.3cm}\\

$\gamma \to 0$ & \begingroup \color{green}{\ding{51}} \endgroup \hspace{4pt} (Theorem \ref{thm:linear relation})  & \multirow{2}{*}{\begingroup \color{red}{\ding{55}} \endgroup \hspace{4pt} (Section \ref{sec:swdisagreement})}  \\
$\gamma > 0$   & \begingroup \color{yellow}{\ding{115}} \endgroup \hspace{4pt} (Theorem \ref{thm:approxlinear1})  &                       \\
\bottomrule
\end{tabular}
\end{table}

\subsection{SW disagreement}\label{sec:swdisagreement}
In Corollary \ref{cor:ridgeless}, unlike I and SS disagreement, SW disagreement contains two linearly independent functions of $\psi$.
Hence, the disagreement-on-the-line phenomenon \eqref{eqn:disontheline} cannot occur for any choice of slope and intercept independent of $\psi$.
Figure \ref{fig:linear} \textbf{(a)} and \textbf{(d)} confirm the non-linear relation between target vs. source SW disagreement in underparameterized and overparameterized regimes, respectively.

\section{Experiments}\label{sec:simulation}
\subsection{Experiments Setup}
We run random features regression with ReLU activation on the following datasets. The codes can be found in \url{https://github.com/dh7401/RF-disagreement}.

\paragraph{CIFAR-10-C.} \cite{hendrycks2018benchmarking} introduced a corrupted version of CIFAR-10 \cite{krizhevsky2009learning}.
We choose two classes and assign the label $y \in \{0, 1\}$ to each.
We use CIFAR-10 as the source domain and CIFAR-10-C as the target domain.

\paragraph{Tiny ImageNet-C.} Tiny ImageNet \cite{wu2017tiny}, a smaller version of ImageNet \cite{deng2009imagenet}, consists of natural images of size $64 \times 64$ in 200 classes. 
Tiny ImageNet-C \cite{hendrycks2019robustness} is a corrupted version of Tiny ImageNet. We down-sample images to $32 \times 32$ and create two super-classes each consisting of 10 of the original classes.
We consider Tiny ImageNet as the source domain and Tiny ImageNet-C as the target domain.

\paragraph{Camelyon17.} Camelyon17 \cite{bandi2018detection} consists of tissue slide images collected from five different hospitals, and the task is to identify tumor cells in the images.
\cite{koh2021wilds} proposed a patch-based variant of the task, where the input $x$ is $96 \times 96$ image and the label $y \in \{0, 1\}$ indicates whether the central $32 \times 32$ contains any tumor tissue.
We crop the central $32 \times 32$ region and use it as the input in our problem.
We use hospital 0 as the source domain and hospital 2 as the target domain.
\vspace{0.5cm}

We use training sample size $n = 1000$, random features dimension $N \in \{3000, 4000, \dots, 49000\}$, input dimension $d = 3072$, regularization $\gamma = 0$.
We test the trained model on the rest of the sample and
plot target vs. source SS disagreement and risk. Plots for I and SW disagreements can be found in Section \ref{sec:ISWdisagreement}.

We estimate the covariance $\Sigma_{\rm s}$ and $\Sigma_{\rm t}$ using the test sample and derive the theoretical slope of target vs. source line predicted by Theorem \ref{thm:linear relation} (see Section \ref{sec:slopeestimation}).
Since the limiting spectral distribution of sample covariance is generally different from that of population covariance, we remark that this may lead to a biased estimate of the slope.
As the intercept $b_{\textnormal{risk}}$ involves the unknown noise level $\sigma_\ep^2$, it is difficult to make a theoretical prediction on its value.
For this reason, we fit the intercept instead using its theoretical value.

\subsection{Results}

\begin{figure*}[h!]
\includegraphics[scale=0.85]{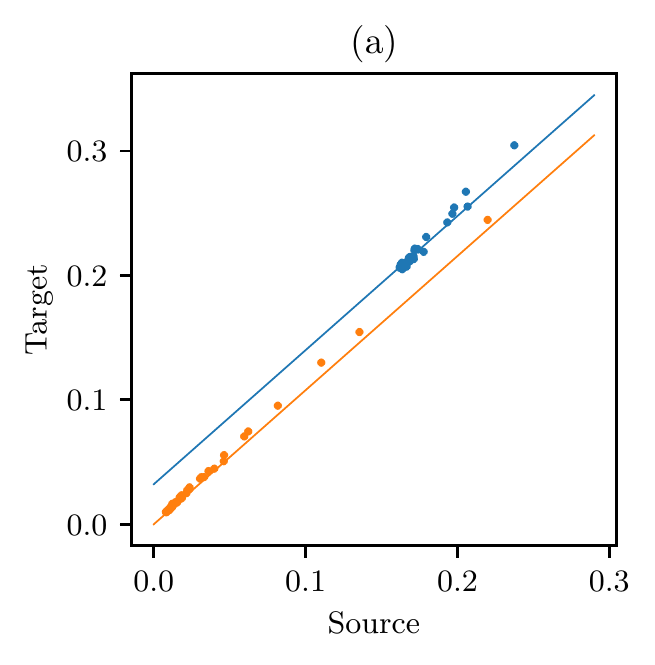}
\includegraphics[scale=0.85]{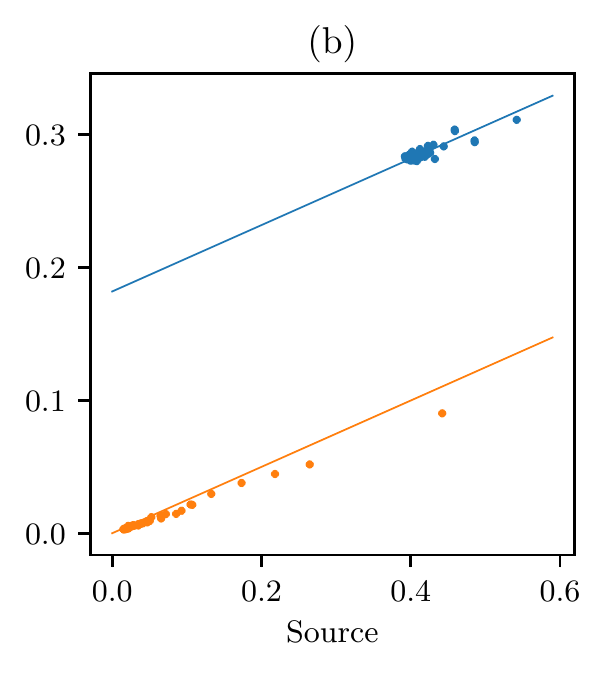}
\includegraphics[scale=0.85]{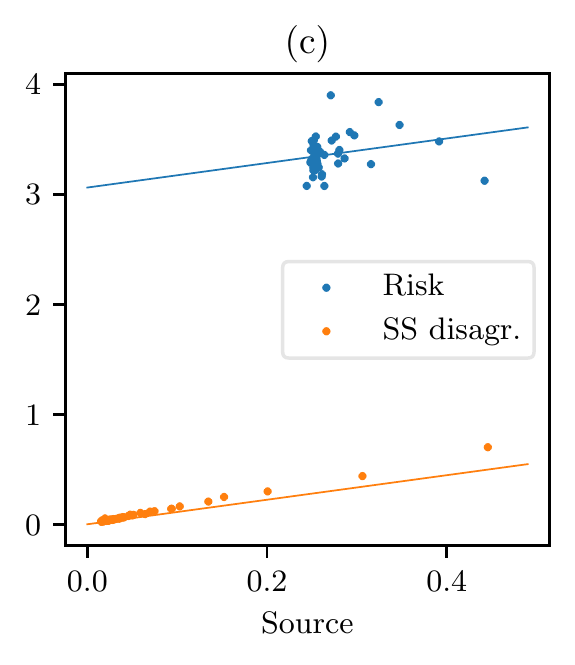}
\caption{\textbf{(a)} CIFAR-10-C-Snow (severity 3) \textbf{(b)} Tiny ImageNet-C-Fog (severity 3) \textbf{(c)} Camelyon17; For more results, see Section \ref{sec:varyingcorruption}.}
\label{fig:experiment}
\end{figure*}

While Theorem \ref{thm:linear relation} is proved only for Gaussian input and linear generative model, we observe the \emph{disagreement-on-the-line} phenomenon on all three datasets (Figure \ref{fig:experiment}), in which these assumptions are violated.

 In this regard, a flurry of recent research (see e.g., \cite{hastie2022surprises,hu2022universality,loureiro2021learning,goldt2022gaussian,wang2022universality,dudeja2022spectral,montanari2022universality}) has proved that findings assuming Gaussian inputs often hold in a much wider range of models.  While none of the existing work exactly fits the setting considered in this paper, this gives yet another indication that our theory should remain true more generally. 
The rigorous characterization of this universality is left for future work. 

Also, we find that target vs. source risk does not exhibit a clear linear trend, especially in Tiny ImageNet and Camelyon17.
This is because Proposition \ref{prop:linearrisk} does not hold in the case of \emph{concept shift}, i.e., the shift in $\mathbb{P}(y | x)$.
However, since disagreement is oblivious to the change of $\mathbb{P}(y | x)$, the \emph{disagreement-on-the-line} is a general phenomenon happening regardless of the type of distribution shift.

\section{Conclusion}
In this paper, we propose a framework to study various types of disagreement in the random features model. We precisely characterize disagreement in high dimensions and study how disagreement under the source and target domains relate to each other.
Our results show that the occurrence of disagreement-on-the-line in the random features model can vary depending on the type of disagreement, regularization, and regime of parameterization.
We show that, contrary to the prior observation, the line for disagreement and the line for risk can differ in their intercepts.
Our findings indicate potential for further examination of the disagreement-on-the-line principle.
We run experiments on several real-world datasets and show that the results hold in settings more general than the theoretical setting that we consider.

\section*{Acknowledgements}


The work of Behrad Moniri is supported by The Institute for Learning-enabled Optimization at Scale (TILOS), under award number NSF-CCF-2112665. Donghwan Lee was supported in part by ARO W911NF-20-1-0080, DCIST, Air Force Office of Scientific Research Young Investigator Program (AFOSR-YIP) \#FA9550-20-1-0111 award; Xinmeng Huang was supported in part by the NSF DMS 2046874 (CAREER), NSF CAREER award CIF-1943064.

{\small
\bibliographystyle{alpha}
\bibliography{ArXiv/Arxiv_Refs}
}

\newpage
\appendix

\section{Technical Tools}

\subsection{Operator-valued Free Probability}\label{sec:operator}
Operator-valued free probability (e.g., \cite{speicher1998combinatorial,mingo2017free, helton2007operator})
has appeared in various studies of random features models including \cite{adlam2019random, adlam2020neural, adlam2020understanding, mel2021anisotropic, ba2022high}.
Here, we briefly outline the most relevant concepts, 
which are used in our computation.

Recall that a set $\mathcal{A}$ is an \emph{algebra} (over the field $\mathbb{C}$ of complex numbers)
if it is a vector space over $\mathbb{C}$
and is endowed with a
\emph{bilinear}
multiplication operation denoted by  ``$\cdot$".
Thus, for all $a,b,c\in \mathcal{A}$
we have 
the distributivity relations 
$a\cdot (b+c) = a\cdot b+a\cdot c$
and
$(b+c) \cdot a = b\cdot a+c\cdot a$;
and the 
relation indicating that multiplication in the algebra is compatible with the usual multiplication over $\mathbb{C}$,
namely that for and 
$x,y \in \mathbb{C}$,
$(x\cdot y) \cdot (a\cdot b) = (x\cdot a) \cdot (y\cdot b)$.
All algebras we consider will be associative, so that the multiplication operation over the algebra is associative.
Further, an algebra is called \emph{unital}  if
it contains a multiplicative identity element; this is denoted as ``1".
Often, we drop the  ``$\cdot$" symbol to denote multiplication (both over the algebra and by scalars), and no confusion may arise.

\begin{definition}[Non-commutative probability space]
Let $\mathcal{C}$ be a unital algebra and $\varphi: \mathcal{C} \to \mathbb{C}$ be a linear map such that $\varphi(1) = 1$.
We call the pair $(\mathcal{C}, \varphi)$ a \emph{non-commutative probability space}.
\end{definition}
\begin{example}[Deterministic matrices]\label{ex:normtrace}
For a matrix $A \in \mathbb{C}^{m \times m}$, we denote its normalized trace by $\overline{\operatorname{tr}}(A) = \frac{1}{m} \sum_{i = 1}^m A_{ii}$.
The pair $(\mathbb{C}^{m \times m}, \overline{\operatorname{tr}})$ is a non-commutative probability space.
\end{example}

\begin{example}[Random matrices]\label{ex:expectednormtrace}
Let $(\Omega, \mathcal{F}, \mathbb{P})$ be a (classical) probability space and $L^{-\infty}(\Omega)$ be the set of scalar random variables with all moments finite.
The pair $(L^{-\infty}(\Omega)^{m \times m}, \mathbb{E} \overline{\operatorname{tr}})$ is a non-commutative probability space.
\end{example}

\begin{definition}[Operator-valued probability space]
    Let $\mathcal{A}$ be a unital algebra and consider a unital sub-algebra $\mathcal{B} \subseteq \mathcal{A}$.
    A linear map $E: \mathcal{A} \to \mathcal{B}$ is a \emph{conditional expectation} if $E(b) = b$ for all $b \in \mathcal{B}$ and $E(b_1 a b_2) = b_1 E(a) b_2$ for all $a \in \mathcal{A}$ and $b_1, b_2 \in \mathcal{B}$.
    The triple $(\mathcal{A}, E, \mathcal{B})$ is called an \emph{operator-valued probability space}.
\end{definition}
The name ``conditional expectation'' can be understood from the following example.
\begin{example}[Classical conditional expectation]
Let $(\Omega, \mathcal{F}, \mathbb{P})$ be a probability space and $\mathcal{G}$ be a sub-$\sigma$-algebra of $\mathcal{F}$.
Then,
considering $E = \mathbb{E}[\cdot | \mathcal{G}]$,
any unital algebra $\mathcal{A} \subset  L^1(\Omega, \mathcal{F}, \mathbb{P})$ and its unital sub-algebra $\mathcal{B} \subset L^1(\Omega, \mathcal{G}, \mathbb{P})$, 
such that all required integrals in the definition 
of
$E(b_1 a b_2) = b_1 E(a) b_2$ 
exist
for all $a \in \mathcal{A}$ and $b_1, b_2 \in \mathcal{B}$,
form an operator-valued probability space $(\mathcal{A}, E, \mathcal{B})$.
\end{example}

\begin{example}[block random matrices]\label{ex:randomblock}
Let $(\mathcal{C}, \varphi) = (L^{-\infty}(\Omega)^{m \times m}, \mathbb{E} \overline{\operatorname{tr}})$ be the non-commutative probability space of random matrices defined in Example \ref{ex:expectednormtrace}.
Define $\mathcal{A} = \mathbb{C}^{M \times M} \otimes \mathcal{C}$ and $\mathcal{B} = \mathbb{C}^{M \times M}$.
In words, $\mathcal{A}$ is the space of $M \times M$ block matrices with entries in $\mathcal{C}$, and $\mathcal{B}$ is the space of $M \times M$ scalar matrices.
Note that $\mathcal{B}$ can be viewed as a unital sub-algebra of $\mathcal{A}$ by the canonical inclusion $\iota: \mathcal{A} \hookrightarrow \mathcal{B}$ defined by
\begin{align}\label{eqn:inclusion}
\iota(B) = B \otimes 1_{\mathcal{C}},
\end{align}
where $1_{\mathcal{C}}$ is the unity of $\mathcal{C}$ (in this example $1_{\mathcal{C}} = I_{m}$).
We also define the block-wise normalized  expected trace $E = \operatorname{id} \otimes \mathbb{E} \overline{\operatorname{tr}}: \mathcal{A} \to \mathcal{B}$ by
\begin{align}
    E(A) = (\mathbb{E}\overline{\operatorname{tr}} A_{ij} )_{1 \leq i, j \leq M}, \quad A = (A_{ij})_{1 \leq i, j \leq M} \in \mathcal{A}.
\end{align}
\end{example}

\begin{remark}\label{rmk:rectangular}
While we have only discussed squared blocks with identical sizes in Example \ref{ex:randomblock}, it is possible to extend the definition to block matrices with rectangular blocks \cite{far2006spectra, far2008slow, benaych2009rectangular, speicher2012free}.
The idea of \cite{benaych2009rectangular} is to embed each rectangular matrix into a block of a common larger square matrix.
For example, if we have rectangular blocks whose dimensions are one of $q_1, \dots, q_K \in \mathbb{N}$, we consider the space of $(q_1 + \cdots + q_K) \times (q_1 + \cdots + q_K)$ square matrices with a block structure
\begin{align*}
\left[
\begin{array}{c|c|c}
q_1 \times q_1 & \cdots & q_1 \times q_K\\ \hline
 \vdots & \ddots & \vdots \\ \hline
q_K \times q_1 & \cdots & q_K \times q_K
\end{array}
\right].
\end{align*}
Then, we identify a rectangular matrix $C \in \mathbb{C}^{q_i \times q_j}$ with a square matrix $\widetilde C \in \mathbb{C}^{(q_1 + \cdots + q_K) \times (q_1 + \cdots + q_K)}$, having the aforementioned block structure, whose $(i, j)$-block is $C$ and all other blocks are zero.
This identification preserves scalar multiplication, 
 addition, multiplication, transpose, and trace,
 in the sense that, for rectangular matrices $C, D$ and a scalar $c \in \mathbb{C}$,
\begin{align*}
    &c \widetilde{C} = \widetilde{c C}, \quad\widetilde{C} + \widetilde{D} = \widetilde{C + D} \quad \text{if $C$ and $D$ have same shape}, \quad (\widetilde C)^\top = \widetilde{C^\top},\\
    &\widetilde C \widetilde D = \begin{cases} \widetilde{CD} &\text{if $C$ and $D$ are conformable,} \\ 0 &\text{otherwise},\end{cases} \quad \quad \operatorname{tr}(\widetilde{C}) = \begin{cases} \operatorname{tr}(C) &\text{if $C$ is a square matrix,} \\ 0 &\text{otherwise.}\end{cases}
\end{align*}
Through this identification, the space of rectangular matrices (with finitely many different dimension types) can be also understood as an algebra over $\mathbb{C}$.
Further, by replacing $\mathcal{C}$ in Example \ref{ex:randomblock} with the space of rectangular random matrices, we can define the space of block random matrices with rectangular blocks.
The space of block random matrices with rectangular blocks, equipped with the block-wised expected trace, will be the operator-valued probability space we consider in our proof. 
\end{remark}

\begin{definition}[Operator-valued Cauchy transform]
    Let $(\mathcal{A}, E, \mathcal{B})$ be an operator-valued probability space.
    For $a \in \mathcal{A}$, define its operator-valued Cauchy transform $\mathcal{G}_a: \mathcal{B} \setminus \{a\} \to \mathcal{B}$ by
        \begin{align*}
        \mathcal{G}_a(b) = E[(b - a)^{-1}].
        \end{align*}
\end{definition}

\begin{definition}[Operator-valued freeness]\label{def:operatorfreeness}
Let $(\mathcal{A}, E, \mathcal{B})$ be an operator-valued probability space and $(\mathcal{A}_i)_{i \in I}$ be a family of sub-algebras of $\mathcal{A}$ which contain $\mathcal{B}$.
The sub-algebras $\mathcal{A}_i$ are \emph{freely independent} over $\mathcal{B}$, if $E[a_1 \cdots a_n] = 0$ whenever $E[a_1] = \cdots = E[a_n] = 0$ and $a_i \in \mathcal{A}_{j(i)}$ for all $i \in [n]$ with $j(1) \neq \cdots \neq j(n)$.
Variables $a_1, \dots, a_n \in \mathcal{A}$ are freely independent over $\mathcal{B}$ if the sub-algebras generated by $a_i$ and $\mathcal{B}$ are freely independent over $\mathcal{B}$.
\end{definition}

Another important transform, introduced in \cite{voiculescu1986addition, voiculescu2006symmetries}, is the $R$-transform.
It enables the characterization of the spectrum of a sum of asymptotically freely independent random matrices.
It was generalized to operator-valued probability spaces in \cite{shlyakhtenko1996random, mingo2017free}.
The definition of operator-valued $R$-transform can be found in Definition 10, Chapter 9 of \cite{mingo2017free}.
Our work does not directly require the definition of $R$-transforms, and instead uses the following property.

\begin{proposition}[Subordination property, (9.21) of \cite{mingo2017free}]
Let $(\mathcal{A}, E, \mathcal{B})$ be an operator-valued probability space.
If $x, y \in \mathcal{A}$ are freely independent over $\mathcal{B}$, then
\begin{equation}\label{eqn:subordination}
    \mathcal{G}_{x + y}(b) = \mathcal{G}_x[b - \mathcal{R}_y(\mathcal{G}_{x + y}(b))]
\end{equation}
for all $b \in \mathcal{B}$, where $\mathcal{R}_y$ is the operator-valued $R$-transform of $y$.
\end{proposition}

\subsection{Limiting $R$-transform of Gaussian Block Matrices}\label{sec:gaussianR}
\cite{shlyakhtenko1996random, shlyakhtenko1998gaussian} proposed using operator-valued free probability to study spectra of Gaussian block matrices.
Their insight was that operator-valued free independence among Gaussian block matrices is guaranteed for general covariance structure, whereas scalar-valued freeness among them only holds in special cases.
Later \cite{far2006spectra, far2008slow, anderson2006clt} revisited this idea.
We present a theorem of \cite{far2008slow}, which characterizes limiting $R$-transform of Gaussian block matrices with rectangular blocks.
\begin{theorem}[Theorem 5 of \cite{far2008slow}]
For $m = m_1 + \cdots + m_M$, let $A = (A_{ij})_{1 \leq i, j \leq M} \in \mathbb{R}^{m \times m}$ be an $M \times M$ block random matrix whose block $A_{ij}$ is a $m_i \times m_j$ random matrix with i.i.d. $\normal(0, c_{ij}^2 / m)$ entries.
Define the covariance function $\sigma(i, j; k, l)$ to be $c_{ij} c_{kl}$ if $A_{ij} / c_{ij} = A_{kl}^\top / c_{kl}$
and 0 otherwise.
We assume the proportional limit where $m_1, \dots, m_M \to \infty$ with $m_i / m \to \alpha_i \in (0,\infty),\, i = 1,\ldots, M$.
Then, the limiting $R$-transform of $A$ can be expressed as
\begin{align}\label{eqn:gaussianrtransform}
     [\mathcal{R}_{A}(D)]_{ij} = \sum_{1 \leq k, l \leq M} \sigma(i, k; l, j) \alpha_k D_{kl},
\end{align}
for any $D \in \mathbb{R}^{M \times M}$.
\end{theorem}
We remark the above statement should be understood in the space of block random matrices with rectangular blocks we discussed in Remark \ref{rmk:rectangular}.
Also, the original statement used a different terminology ``covariance mapping'', but it is identical to the $R$-transform of $A$ (see discussion in \cite{mingo2017free} p.242 and \cite{far2006spectra} p.24)

\subsection{Centering Random Features}\label{sec:centering}
We first argue that the random features $F, f$ can be centered without changing the asymptotics of disagreement.
This centering argument became a standard technique after it was introduced in \cite{mei2022generalization} (Section 10.4).
More generally, centering arguments are standard in random matrix theory (see e.g., \cite{bai2010spectral}).
For a standard Gaussian random variable $Z \sim \normal(0, 1)$, define centered random features by
\begin{align*}
\bar F = F - \E \sigma(\sqrt{m_{\rm s }} Z), \quad \bar f= f - \E\sigma(\sqrt{m_j}Z),
\end{align*}
where $j \in \{\rm s, \rm t\}$ is the domain that input $x$ comes from.
Subtracting a scalar from a matrix/vector should be understood entry-wise.
The following lemma states that model prediction obtained from these centered random features is close to the original prediction $\hat{y}(x)$ with high probability.
\begin{lemma}
Define centered model prediction by
\begin{align*}
\bar{\hat{y}}(x) = Y^\top \left( \frac{1}{N} \bar F^\top \bar F + \gamma I_n \right)^{-1} \left( \frac{1}{N} \bar F^\top \bar f \right).
\end{align*}
There exist constants $c_1, c_2, c_3, c_4 > 0$ such that
\begin{align*}
|\bar{\hat{y}}(x) - \hat{y}(x)| \leq c_1 d^{-c_2}
\end{align*}
with probability at least $1 - c_3 d^{-c_4}$.
\end{lemma}
This lemma is a consequence of Lemma I.7 and Lemma I.8 of \cite{tripuraneni2021covariate}.
Since we consider the limit $n, d, N \to \infty$, disagreement $\stsd_{i} (\phi, \psi, \gamma), i \in \{\textnormal{I}, \textnormal{SS}, \textnormal{SW}\}$ are invariant to the centering.
We also remark that the non-linearity constants defined in \eqref{eqn:defconstants} are also unchanged after this centering.
For these reasons, perhaps with a slight abuse of notation, we assume $F$ and $f$ are centered from now on.

\subsection{Gaussian Equivalence}\label{sec:gaussianequivalents}
For domain $j \in \{\rm s, \rm t\}$ that input $x$ is drawn from,
we consider the following \emph{noisy linear} random features
\begin{align}\label{eqn:ge}
    \tilde F &= \sqrt{\frac{\rho_{\rm s}}{d}} WX + \sqrt{\rho_{\rm s} \omega_{\rm s}} \Theta, \quad \tilde f = \sqrt{\frac{\rho_j}{d}} Wx + \sqrt{\rho_j \omega_j} \theta,
\end{align}
where $\Theta \in \R^{N \times n}$ and $\theta \in \R^{N}$ have i.i.d. standard Gaussian entries independent from all other Gaussian matrices.
The coefficients above are chosen so that the first and second moment of $\tilde F$ and $\tilde f$ match those of $F$ and $f$, respectively.
We call $\tilde F, \tilde f$
the \emph{Gaussian equivalent} of $F, f$ as we claim the following.

\begin{claim}[Gaussian equivalence]\label{claim:ge}
The asymptotic limit (Condition \ref{cond:limit}) of the disagreement (Definition \ref{def:disagreement}) of the random features model \eqref{eqn:yhat} is invariant to the substitution $F, f \to \tilde{F}, \tilde{f}$.
\end{claim}

This idea was introduced in context of random kernel matrices \cite{el2010spectrum, cheng2013spectrum, fan2019spectral} and has been repeatedly used in recent studies of random feature models.
\cite{mei2022generalization} proved the Gaussian equivalence for random weights uniformly distributed on a sphere.
\cite{montanari2019generalization} conjectured that the same holds for classification.
\cite{adlam2020neural, adlam2020understanding, tripuraneni2021covariate} derived several asymptotic properties of random features models building on the Gaussian equivalence conjecture.
\cite{goldt2022gaussian} provided theoretical and numerical evidence suggesting that the Gaussian equivalence holds for a wide class of models including random features models. \cite{mel2021anisotropic, d2021interplay, loureiro2021learning} conjectured the Gaussian equivalence for anisotropic inputs. 
\cite{hassani2022curse} showed the Gaussian equivalence holds for the adversarial risk of adversarially trained random features models.
\cite{hu2022universality} showed the conjecture for isotropic Gaussian inputs, under mild technical conditions. \cite{montanari2022universality} generalized this by removing the isotropic condition and relaxing the Gaussian input assumption.

More generally, the phenomenon that eigenvalue statistics in the bulk spectrum of a random matrix do not depend on the specific law of the matrix entries is referred to as ``bulk universality'' \cite{wigner1955characteristic, gaudin1961loi, mehta2004random, dyson1962brownian} and has been a central subject in the random matrix theory literature \cite{erdHos2010bulk, erdHos2012bulk, el2010spectrum, tao2011random}.

It is known that local spectral laws of correlated random hermitian matrices can be fully determined by their first and second moments, through the matrix Dyson equation \cite{erdos2019matrix}.
Also, \cite{banna2015limiting, banna2020clt} showed that spectral distributions of correlated symmetric random matrices and sample covariance matrices can be characterized by Gaussian matrices with identical correlation structure.
However, these results do not directly imply Claim \ref{claim:ge} since we do not study the spectral properties of $F, f$ on their own.

\subsection{Linear Pencils}\label{sec:linearpencil}
After applying the Gaussian equivalence \eqref{eqn:ge}, each of the quantities that we study becomes an expected trace of a rational function of random matrices.
To analyze this, we use the \emph{linear pencil} method \cite{MR2183281, MR2233126, anderson2013convergence, helton2018applications}, in which we build a large block matrix whose blocks are linear functions of variables and one of the blocks of its inverse is the desired rational function.
Then, operator-valued free probability can be used to extract block-wise spectral properties of the inverse.
For example, if we want to compute $\E \operatorname{tr} [(\frac{X^\top X}{d} + \gamma I_n)^{-1}]$ for $X \in \mathbb{R}^{d \times n}$, we consider
\begin{align*}
    \begin{bmatrix} I_n & -\frac{X^\top}{\sqrt{\gamma d}}\\
    \frac{X}{\sqrt{\gamma d}} & I_d\end{bmatrix},
\end{align*}
inverse has as its (1, 1)-block $\gamma (\frac{X^\top X}{d} + \gamma I_n)^{-1}$.
Block matrices for more complicated rational functions can be constructed using the following proposition.
\begin{proposition}[Algorithm 4.3 of \cite{helton2018applications}]
Let $x_1, \dots, x_g$ be elements of an algebra $\mathcal{A}$ over a field $\mathbb{K}$.
For an $m \times m$ matrix $Q$ and vectors $u, v \in \mathbb{K}^m$, a triple $(u, Q, v)$ is called a linear pencil of a rational function $r \in \mathbb{K}(x_1, \dots, x_g)$ if each entry of $Q$ is a $\mathbb{K}$-affine function of $x_1, \dots, x_g$ and $r = -u^\top Q^{-1} v$.
The following holds.
\begin{enumerate}
    \item (Addition) If $(u_1, Q_1, v_1)$ and $(u_2, Q_2, v_2)$ are linear pencils of $r_1$ and $r_2$, respectively, then
    \begin{align*}
        \left( \begin{bmatrix} u_1 \\ u_2 \end{bmatrix}, \begin{bmatrix} Q_1 & 0_{m \times m} \\ 0_{m \times m} 
 & Q_2 \end{bmatrix}, \begin{bmatrix} v_1 \\ v_2 \end{bmatrix} \right)
    \end{align*}
    is a linear pencil of $r_1 + r_2$.

    \item (Multiplication) If $(u_1, Q_1, v_1)$ and $(u_2, Q_2, v_2)$ are linear pencils of $r_1$ and $r_2$, respectively, then
    \begin{align*}
    \left( \begin{bmatrix} 0_m \\ u_1 \end{bmatrix}, \begin{bmatrix} x_g v_1 u_2^\top & Q_1 \\ Q_2 & 0_{m \times m} \end{bmatrix}, \begin{bmatrix} 0_m \\ v_2  \end{bmatrix}\right)
    \end{align*}
    is a linear pencil of $r_1 x_g r_2$.

    \item (Inverse) If $(u, Q, v)$ is a linear pencil of $r$, then
    \begin{align*}
    \left( \begin{bmatrix} 1 \\ 0_m \end{bmatrix}, \begin{bmatrix} 0 & u^\top \\ v & - Q^{-1} \end{bmatrix}, \begin{bmatrix} 1 \\ 0_m \end{bmatrix} \right)
    \end{align*}
    is a linear pencil of $r^{-1}$.
\end{enumerate}
\end{proposition}

In this language, 
the example before the algorithm can be interpreted
in the space we consider in Remark \ref{rmk:rectangular}
as 
$r = -\gamma (\frac{X^\top X}{d} + \gamma I_n)^{-1}$
being a rational function of $X$ and $X^\top$, and 
\begin{align}\label{eqn:linearpencilexample}
    \left( \begin{bmatrix} 1 \\ 0 \end{bmatrix}, 
      \begin{bmatrix} I_n & -\frac{X^\top}{\sqrt{\gamma d}}\\
    \frac{X}{\sqrt{\gamma d}} & I_d\end{bmatrix}, 
    \begin{bmatrix} 1 \\ 0 \end{bmatrix} \right)
\end{align}
being a linear pencil of $r$.

In principle, repeated application of the above rules to basic building blocks such as \eqref{eqn:linearpencilexample} can produce a linear pencil for any rational function of given random matrices.
For example, consider $X_1, X_2 \in \mathbb{R}^{d \times n}, \Sigma \in \mathbb{R}^{d \times d}$ and their transpose as elements of the algebra over $\mathbb{R}$ we discussed in Remark \ref{rmk:rectangular}.
Then,
\begin{align*}
\left(\begin{bmatrix} 1 \\ 0 \\ 0 \\ 0 \end{bmatrix} , \begin{bmatrix}
 I_n & -\frac{X_1^\top}{\sqrt{\gamma d}} & -\frac{\Sigma}{\gamma^2} & \cdot \\
 \frac{X_1}{\sqrt{\gamma d}} & I_d & \cdot & \cdot \\
 \cdot & \cdot & I_n & -\frac{X_2^\top}{\sqrt{\gamma d}}\\
 \cdot & \cdot & \frac{X_2}{\sqrt{\gamma d}} & I_d
\end{bmatrix}, \begin{bmatrix} 0 \\ 0 \\ 1 \\ 0 \end{bmatrix} \right)
\end{align*}
is a linear pencil of $r' = -(\frac{X_1^\top X_1}{d} + \gamma I_n)^{-1} \Sigma (\frac{X_2^\top X_2}{d} + \gamma I_n)^{-1}$.
Here, we denote zero blocks by dots.
This can be seen by applying the multiplication rule to two copies of \eqref{eqn:linearpencilexample} and $x_g = \Sigma$, and then switching the first and the second pairs of columns.

However, constructing a suitably small linear pencil is a non-trivial problem of independent interest (see discussions on reductions of linear pencils in e.g., \cite{volvcivc2018matrix, helton2018applications} and references therein).
This is one of the challenges we need to overcome in our proofs.

\section{Proofs}
\subsection{Proof of Theorem \ref{thm:asymp}}
Starting from this section, we omit the high-dimensional limit signs $\lim_{n, d, N \to \infty}$ (Condition \ref{cond:limit}) for a simpler presentation.
However, every expectation appearing in the derivation should be understood as its high-dimensional limit.

For $j \in \{\rm s, \rm t\}$, independent disagreement satisfies
\begin{align*}
    \stsd_{\textnormal{I}}^j (\phi, \psi, &\gamma) = \E [(\hat{y}_{W_1, X_1, Y_1}(x) -\hat{y}_{W_2, X_2, Y_2}(x))^2]\\
    &= \E[(\hat{y}_{W_1, X_1, Y_1}(x) - \E_{W_1, X_1, Y_1}[\hat{y}_{W_1, X_1, Y_1}(x)] + \E_{W_2, X_2, Y_2}[\hat{y}_{W_2, X_2, Y_2}(x)] - \hat{y}_{W_2, X_2, Y_2}(x))^2]\\
    &= \E_{\beta, x \sim \cD_j}[(\hat{y}_{W_1, X_1, Y_1}(x) - \E_{W_1, X_1, Y_1}[\hat{y}_{W_1, X_1, Y_1}(x)])^2]\\
    &\quad \quad+ \E_{\beta, x \sim \cD_j}[(\hat{y}_{W_2, X_2, Y_2}(x) - \E_{W_2, X_2, Y_2}[\hat{y}_{W_2, X_2, Y_2}(x)])^2]\\
    &= \E_{\beta, x \sim \cD_j} \mathbb{V}_{W_1, X_1, Y_1}(\hat{y}_{W_1, X_1, Y_1}(x)) + \E_{\beta, x \sim \cD_j} \mathbb{V}_{W_2, X_2, Y_2}(\hat{y}_{W_2, X_2, Y_2}(x))  = 2 V_j.
\end{align*}
Plugging in the variance $V_j$ given in Theorem \ref{thm:biasandvariance}, we obtain the formula for $\stsd_{\textnormal{I}}^j (\phi, \psi, \gamma)$.

\subsubsection{Decomposition of $\stsd_{\textnormal{SS}}^j (\phi, \psi, \gamma)$}
Writing $F_i = \sigma(W_i X / \sqrt{d})$, $f_i = \sigma(W_i x / \sqrt{d})$, $K_i = \frac{1}{N} F_i^\top F_i + \gamma I_n$ for $i \in \{1, 2\}$, we can write shared-sample disagreement as
\begin{align}\label{eqn:inddecompose}
    \stsd_{\textnormal{SS}}^j (\phi, \psi, \gamma) &= \frac{1}{N^2} \E [(Y^\top K_1^{-1} F_1^\top f_1 - Y^\top K_2^{-1} F_2^\top f_2 )^2] \nonumber \\
    &= \frac{2}{N^2} \E[f_1^\top F_1 K_1^{-1} Y Y^\top K_1^{-1} F_1^\top f_1] - \frac{2}{N^2} \E[f_2^\top F_2 K_2^{-1} Y Y^\top K_1^{-1} F_1^\top f_1 ] \nonumber\\
    &= D_1 - D_2.
\end{align}
The term $D_1$ was computed in (A268), (A279), (A462), (A546) of \cite{tripuraneni2021covariate} as
\begin{align}\label{eqn:d1}
    D_1 = 2V_j + \frac{2 \rho_j \kappa^2}{\rho_{\rm s} \phi} \mathcal{I}_{3, 2}^j.
\end{align}
Plugging in $Y = X^\top \beta /\sqrt{d} + \boldsymbol{\varepsilon}$, where $\boldsymbol{\varepsilon} = (\ep_1, \dots, \ep_n)^\top \in \R^n$, the term $D_2$ becomes
\begin{align*}
    D_2 = &\frac{2}{d N^2} \E_{W_i, X}\operatorname{tr}[ K_2^{-1} X^\top \E_\beta [\beta \beta^\top] X  K_1^{-1} F_1^\top \E_{x \sim \cD_j, \theta}[f_1 f_2^\top] F_2 ] \\
    &\quad \quad+ \frac{4}{\sqrt{d}N^2} \E_{W_i, X}[ K_2^{-1} X^\top \E_{\beta, \boldsymbol{\varepsilon}} [\beta \boldsymbol{\varepsilon}^\top]  K_1^{-1} F_1^\top \E_{x \sim \cD_j, \theta}[f_1f_2^\top] F_2]\\
    &\quad \quad+ \frac{2}{N^2} \E_{W_i, X} \operatorname{tr} [K_2^{-1} \E_{\boldsymbol{\varepsilon}} [\boldsymbol{\varepsilon} \boldsymbol{\varepsilon}^\top]  K_1^{-1} F_1^\top \E_{x \sim \cD_j, \theta}[f_1 f_2^\top] F_2 ]\\
    = &\frac{2}{d N^2} \E_{W_i, X}\operatorname{tr}[K_2^{-1} X^\top X  K_1^{-1} F_1^\top \E_{x \sim \cD_j, \theta}[f_1 f_2^\top] F_2]\\
    &\quad \quad + \frac{2 \sigma_\ep^2}{N^2} \E_{W_i, X} \operatorname{tr} [K_2^{-1}  K_1^{-1} F_1^\top \E_{x \sim \cD_j, \theta}[f_1 f_2^\top] F_2].
\end{align*}
From the Gaussian equivalence \eqref{eqn:ge}, we have
\begin{align*}
    \E_{x \sim \cD_j, \theta}[f_1 f_2^\top] = \frac{\rho_j}{d} W_1 \Sigma_j W_2^\top.
\end{align*}
Therefore,
\begin{align}\label{eqn:d2}
    D_2 &= \frac{2\rho_j}{d^2 N^2} \E_{W_i, X} \operatorname{tr}[W_1 \Sigma_j W_2^\top F_2 K_2^{-1} X^\top X K_1^{-1} F_1^\top] + \frac{2 \sigma_\ep^2 \rho_j}{d N^2} \E_{W_i, X} \operatorname{tr}[K_1^{-1} F_1^\top W_1 \Sigma_j W_2^\top F_2 K_2^{-1}] \nonumber \\
    &= D_{21} + D_{22}.
\end{align}
We can write $X = \Sigma_{\rm s}^\frac{1}{2} Z$ for $Z \in \R^{d \times n}$ with i.i.d. standard Gaussian entries.
Thus,
\begin{align*}
    D_{21} &= \frac{2\rho_j}{d^2 N^2} \E_{W_i, Z} \operatorname{tr}[W_1 \Sigma_j W_2^\top F_2 K_2^{-1} Z^\top \Sigma_{\rm s} Z K_1^{-1} F_1^\top],\\
    D_{22} &= \frac{2\sigma_\ep^2 \rho_j}{d N^2} \E_{W_i, Z}\operatorname{tr}[ K_1^{-1} F_1^\top W_1 \Sigma_j W_2^\top F_2 K_2^{-1}].
\end{align*}
Now, we use the linear pencil method \cite{helton2018applications} to build a block matrix such that (1) each block is either deterministic or a constant multiple of $Z, W_i, \Theta_i$ and (2) $D_{21}$ or $D_{22}$ appears as trace of a block of its inverse.
Then, we compute the operator-valued Cauchy transform of the block matrix and extract $D_{21}$ and $D_{22}$ from the result.

\subsubsection{Preliminary computations}\label{sec:auxcomp}
We present some preliminary computations that will be used in later sections.
We will also use the linear pencil $Q^0$ as a building block when constructing other linear pencils.
Most of the computations here are adopted from Section A.9.6.1 of \cite{tripuraneni2021covariate}.
For clarity and to be self-contained, we provide our own version of the same result updated in some minor ways.

Using 
$W, Z$ and other notations from Section \ref{sec:prelim} and
$\Theta$  from \eqref{eqn:ge},
let
\begin{align*}
    Q^0 = \begin{bmatrix} I_n & \frac{\sqrt{\rho_{\rm s} \omega_{\rm s}}\Theta^\top}{\gamma \sqrt{N}} & \frac{\sqrt{\rho_{\rm s}}Z^\top}{\gamma\sqrt{d}} & \cdot & \cdot& \cdot\\ -\frac{\Theta \sqrt{\rho_{\rm s} \omega_{\rm s}}}{\sqrt{N}} & I_{N} & \cdot & \cdot & -\frac{\sqrt{\rho_{\rm s}}W}{\sqrt{N}} & \cdot \\ \cdot & \cdot & I_{d} & -\Sigma_{\rm s}^\frac{1}{2} & \cdot & \cdot\\ \cdot & -\frac{W^\top}{\sqrt{N}} & \cdot & I_{d} & \cdot & \cdot \\ \cdot & \cdot& \cdot & \cdot & I_{d} & -\Sigma_{\rm s}^\frac{1}{2}\\ -\frac{Z}{\sqrt{d}} & \cdot & \cdot & \cdot & \cdot & I_{d} \end{bmatrix}.
\end{align*}
Recall from Example \ref{ex:normtrace} that we denote the normalized trace of a matrix $A$ by $\overline{\operatorname{tr}}(A)$.
Define the block-wise normalized expected trace of $(Q^0)^{-1}$ by $G^0 = (\operatorname{id} \otimes \mathbb{E} \overline{\operatorname{tr}}) ((Q^0)^{-1})$.
From block matrix inversion, we see
\begin{align}\label{eqn:g0blockinversion}
     &G_{1,1}^0 = \gamma\, \mathbb{E}\overline{\operatorname{tr}}(K^{-1}), \quad G_{3,6}^0 = \frac{\gamma \sqrt{\rho_{\rm s}}\,\E \overline{\operatorname{tr}} [\Sigma_{\rm s} W^\top \hat{K}^{-1} W]}{N}, \quad G_{5, 4}^0 = -\frac{\sqrt{\rho_{\rm s}}\, \E \overline{\operatorname{tr}} [\Sigma_{\rm s} Z K^{-1} Z^\top]}{d},
\end{align}
in which $\hat{K} = \frac{1}{N}FF^\top  + \gamma I_{N}$. We augment the matrix $Q^0$ to form the symmetric matrix $\bar{Q}^0$ as 
\begin{align*}
    \bar{Q}^0 = \begin{bmatrix} \cdot & (Q^0)^\top \\ Q^0 & \cdot \end{bmatrix}.
\end{align*}
This matrix can be written as 
\begin{align*}
    \bar{Q}^0 &= \bar{Z}^0 - \bar{Q}_{W, Z, \Theta}^0 - \bar{Q}_\Sigma^0 \\
    &=\begin{bmatrix} \cdot & I_{n + 4d + N} \\ I_{n + 4d + N} & \cdot \end{bmatrix} - \begin{bmatrix} \cdot & (Q_{W, Z, \Theta}^0)^\top \\ Q_{W, Z, \Theta}^0 & \cdot \end{bmatrix} -\begin{bmatrix} 0 & (Q_\Sigma^0)^\top \\ Q_\Sigma^0 & \cdot \end{bmatrix},
\end{align*}
with
\begin{align*}
    Q_{W,Z,\Theta}^0= \begin{bmatrix} \cdot & -\frac{\sqrt{\rho_{\rm s} \omega_{\rm s}}\Theta^\top}{\gamma \sqrt{N}} & -\frac{\sqrt{\rho_{\rm s}}Z^\top}{\gamma\sqrt{d}} & \cdot & \cdot& \cdot\\ \frac{\Theta \sqrt{\rho_{\rm s} \omega_{\rm s}}}{\sqrt{N}} & \cdot & \cdot & \cdot & \frac{\sqrt{\rho_{\rm s}}W}{\sqrt{N}} & \cdot \\ \cdot & \cdot & \cdot & \cdot & \cdot & \cdot\\ \cdot & \frac{W^\top}{\sqrt{N}} & \cdot & \cdot & \cdot & \cdot \\ \cdot & \cdot& \cdot & \cdot & \cdot & \cdot\\ \frac{Z}{\sqrt{d}} & \cdot & \cdot & \cdot & \cdot & \cdot \end{bmatrix}\quad \text{and}\quad
    Q_{\Sigma}^0 = \begin{bmatrix} \cdot & \cdot & \cdot & \cdot & \cdot& \cdot\\ \cdot & \cdot & \cdot & \cdot & \cdot & \cdot \\ \cdot & \cdot & \cdot & \Sigma_{\rm s}^\frac{1}{2} & \cdot & \cdot\\ \cdot & \cdot & \cdot & \cdot & \cdot & \cdot \\ \cdot & \cdot& \cdot & \cdot & \cdot & \Sigma_{\rm s}^\frac{1}{2}\\ \cdot & \cdot & \cdot & \cdot & \cdot & \cdot \end{bmatrix}.
\end{align*}
Defining $\bar G^0$ as below, we have
\begin{align*}
    \bar{G}^0 &= \begin{bmatrix} \cdot & G^0 \\ (G^0)^\top & \cdot \end{bmatrix} = \begin{bmatrix} \cdot & (\operatorname{id} \otimes \mathbb{E} \overline{\operatorname{tr}}) ((Q^0)^{-1}) \\ (\operatorname{id} \otimes \mathbb{E} \overline{\operatorname{tr}}) (((Q^0)^\top)^{-1}) & \cdot \end{bmatrix}\\
    &= (\operatorname{id} \otimes \mathbb{E} \overline{\operatorname{tr}}) \begin{bmatrix} \cdot & (Q^0)^{-1} \\ ((Q^0)^{\top})^{-1} & \cdot \end{bmatrix} = (\operatorname{id} \otimes \mathbb{E} \overline{\operatorname{tr}}) ((\bar{Q}^0)^{-1}).
\end{align*}
Thus, $\bar{G}^0$ can be viewed as the operator-valued Cauchy transform of $\bar{Q}_{W, Z, \Theta}^0 + \bar{Q}_\Sigma^0$ (in the space we consider in Remark \ref{rmk:rectangular}),
\begin{equation*}
    \bar{G}^0 = (\operatorname{id} \otimes \mathbb{E} \overline{\operatorname{tr}}) (\bar{Z}^0 - \bar{Q}_{W, Z, \Theta}^0 - \bar{Q}_\Sigma^0)^{-1} = \mathcal{G}_{\bar{Q}_{W, Z, \Theta}^0 + \bar{Q}_\Sigma^0}(\bar{Z}^0).
\end{equation*}
Here, we implicitly used the canonical inclusion defined in \eqref{eqn:inclusion} to write
\begin{align*}
\bar Z^0 = \begin{bmatrix} \cdot & I_6 \\ I_6 & \cdot \end{bmatrix}.
\end{align*}
Since $\bar Q_\Sigma^0$ is deterministic, the matrices $\bar Q_{W, Z, \Theta}^0$ and $\bar Q_{\Sigma}^0$ are asymptotically freely independent according to Definition \ref{def:operatorfreeness}.
Hence by the subordination formula \eqref{eqn:subordination}, 
\begin{align}
    \bar{G}^0 &= \mathcal{G}_{\bar{Q}_\Sigma^0}(\bar{Z}^0 - \mathcal{R}_{\bar{Q}_{W, Z, \Theta}^0}(\bar{G}^0)) = (\operatorname{id} \otimes \mathbb{E} \overline{\operatorname{tr}})(\bar{Z}^0 - \mathcal{R}_{\bar{Q}_{W, Z, \Theta}^0}(\bar{G}^0) - \bar{Q}_\Sigma^0)^{-1}\label{eq:subordination_Q}.
\end{align}
Since $\bar{Q}_{W, Z, \Theta}^0$ consists of i.i.d. Gaussian blocks, we use \eqref{eqn:gaussianrtransform}
to find the $R$-transform $\mathcal{R}_{\bar{Q}_{W, Z, \Theta}^0}(\bar{G}^0)$ of the form
 \begin{align*}
     \mathcal{R}_{\bar{Q}_{W, Z, \Theta}^0}(\bar{G}^0) = \begin{bmatrix} \cdot & (R^0)^\top \\ R^0 & \cdot \end{bmatrix}.
 \end{align*}
For example, to find $R_{1, 1}^0$, we look for a block in the first row of $\bar{Q}_{W, Z, \Theta}^0$ and a block in the first column of $\bar{Q}_{W, Z, \Theta}^0$ such that they are transpose to each other up to a constant factor.
There are two such pairs, ((1, 2)-block, (2, 1)-block) and ((1, 3)-block, (6, 1)-block).
Therefore, the equation \eqref{eqn:gaussianrtransform} gives
\begin{align*}
    R_{1, 1}^0 = -\frac{\rho_{\rm s} \omega_{\rm s}}{\gamma}G_{2, 2}^0 - \frac{\sqrt{\rho_{\rm s}}}{\gamma}G_{3,6}^0.
\end{align*}
 
Repeating the same procedure, the non-zero blocks of $R^0$ are
\begin{align*}
    &R_{1,1}^0 = -\frac{\rho_{\rm s} \omega_{\rm s}}{\gamma}G_{2, 2}^0 - \frac{\sqrt{\rho_{\rm s}}}{\gamma}G_{3,6}^0,\quad R_{2,2}^0 = -\frac{\rho_{\rm s} \omega_{\rm s}\psi}{\gamma\phi} G_{1, 1}^0 + \sqrt{\rho_{\rm s}}\psi G_{5,4}^0,\\
    &R_{4,5}^0 = \sqrt{\rho_{\rm s}} G_{2,2}^0, \quad R_{6,3}^0 = -\frac{\sqrt{\rho_{\rm s}}G_{1,1}^0}{\gamma\phi}.
\end{align*}
Plugging this into equation \eqref{eq:subordination_Q}, we obtain self-consistent equations for $G^1$.
For example,
\begin{align*}
    G_{3, 6}^0 &= \E \overline{\operatorname{tr}} [(I_{n + 4d + N} - R^0 - Q^0_\Sigma)]_{3, 6} = \E \overline{\operatorname{tr}} \left[ \gamma \sqrt{\rho_{\rm s}} \phi G_{2, 2}^0  \Sigma_{\rm s} (\gamma \phi I_d + \rho_{\rm s} G_{1, 1}^0 G_{2, 2}^0 \Sigma_{\rm s})^{-1}\right]\\
    &= \E_\mu\left[\frac{\lambda^{\rm s}\gamma\sqrt{\rho_{\rm s}}\phi G_{2,2}^0}{\gamma\phi + \lambda^{\rm s} \rho_{\rm s} G_{1,1}^0G_{2,2}^0}\right].
\end{align*}

Similarly,
\begin{align*}
    G_{1,1}^0 &= \frac{\gamma}{\gamma + \rho_{\rm s} \omega_{\rm s} G_{2,2}^0   + \sqrt{\rho_{\rm s}}G_{3,6}^0}, \quad G_{2,2}^0 = \frac{\gamma\phi}{\gamma\phi+  \rho_{\rm s} \omega_{\rm s} \psi G_{1,1}^0-\gamma\sqrt{\rho_{\rm s}}\psi\phi G_{5,4}^0}\\    G_{3,6}^0&=\E_\mu\left[\frac{\lambda^{\rm s}\gamma\sqrt{\rho_{\rm s}}\phi G_{2,2}^0}{\gamma\phi + \lambda^{\rm s} \rho_{\rm s} G_{1,1}^0G_{2,2}^0}\right], \quad G_{5,4}^0 = -\E_\mu\left[\frac{\lambda^{\rm s}
    \sqrt{\rho_{\rm s}}G_{1,1}^0}{\gamma\phi + \lambda^{\rm s} \rho_{\rm s} G_{1,1}^0G_{2,2}^0}\right].
\end{align*}

Now, by eliminating $G_{3, 6}^0, G_{5,4}^0$ and expressing in terms of $\kappa, \tau,$ and $\bar{\tau}$ defined in \eqref{eqn:defkappa} and \eqref{eqn:deftau}, we can show that $\E\overline{\operatorname{tr}}(K^{-1}) = \frac{G_{1,1}^0}{\gamma} = \tau$, and $\E\overline{\operatorname{tr}}(\hat{K}^{-1}) = \frac{G_{2,2}^0}{\gamma} = \bar{\tau}$. Thus, using equation \eqref{eqn:deffunctionals} we have
\begin{align}\label{eqn:g0values}
    G_{1,1}^0 = \gamma \tau, \quad G_{2,2}^0 = \gamma\bar{\tau},\quad G_{3,6}^0 = \gamma\sqrt{\rho_{\rm s}}\bar{\tau}\cI_{1,1}^{\rm s}, \quad G_{5,4}^0 = -\frac{\sqrt{\rho_{\rm s}}\tau \mathcal{I}_{1, 1}^{\rm s}}{\phi}.
\end{align}

\subsubsection{Computation of $D_{21}$}
Define $Q^{1}$ by

\begingroup
\setlength\arraycolsep{0pt}
\begin{align*}
  \scalemath{0.95}{
  Q^{1} = \begin{bmatrix}
  I_n & \frac{\sqrt{\rho_{\rm s} \omega_{\rm s}}\Theta_2^\top}{\gamma \sqrt{N}} & \frac{\sqrt{\rho_{\rm s}}Z^\top}{\gamma \sqrt{d}} & \cdot &\cdot & \cdot& \cdot &\cdot & \cdot& \cdot &\cdot & \cdot &\cdot & \cdot\\
  -\frac{\sqrt{\rho_{\rm s} \omega_{\rm s}}\Theta_2}{\sqrt{N}} & I_{N} & \cdot & \cdot & -\frac{\sqrt{\rho_{\rm s}}W_2}{\sqrt{N}} & \cdot &\cdot & \cdot& \cdot &\cdot & \cdot& \cdot &\cdot & \cdot\\
  \cdot & \cdot & I_{d} & -\Sigma_{\rm s}^\frac{1}{2} & \cdot & \cdot &\cdot & \cdot& \cdot &\cdot & \cdot& \cdot &\cdot & \cdot\\
  \cdot & -\frac{W_2^\top}{\sqrt{N}} & \cdot & I_{d} & \cdot & \cdot &\cdot & \cdot& \cdot &\cdot & \cdot& \frac{\Sigma_{\rm s}^\frac{1}{2}}{\sqrt{\rho_{\rm s}}} &\cdot & \cdot\\
  \cdot & \cdot & \cdot & \cdot & I_{d} & -\Sigma_{\rm s}^\frac{1}{2} &\cdot & \cdot& \cdot &\cdot & \cdot& \cdot &\cdot & \cdot\\
  -\frac{Z}{\sqrt{d}} & \cdot & \cdot & \cdot & \cdot & I_{d} &\cdot & \cdot& \cdot &\cdot & \cdot& \cdot &\cdot & \cdot\\
  \cdot & \cdot& \cdot &\cdot & \cdot& \cdot & I_n & \frac{\sqrt{\rho_{\rm s} \omega_{\rm s}}\Theta_1^\top}{\gamma \sqrt{N}}& \frac{\sqrt{\rho_{\rm s}} Z^\top}{\gamma \sqrt{d}} &\cdot &\cdot &\cdot &\cdot & \cdot\\ 
  \cdot & \cdot& \cdot &\cdot & \cdot& \cdot & -\frac{\sqrt{\rho_{\rm s} \omega_{\rm s}}\Theta_1}{\sqrt{N}} & I_{N} & \cdot & \cdot & -\frac{\sqrt{\rho_{\rm s}}W_1}{\sqrt{N}} & \cdot &\cdot & \cdot\\ 
  \cdot & \cdot& \cdot &\cdot & \cdot& \cdot & \cdot & \cdot & I_{d} & -\Sigma_{\rm s}^\frac{1}{2} & \cdot & \cdot &\cdot & \cdot\\ 
  \cdot & \cdot& \cdot &\cdot & \cdot& \cdot & \cdot & -\frac{W_1^\top}{\sqrt{N}} & \cdot & I_{d} & \cdot & \cdot &\cdot & \cdot\\ 
  \cdot & \cdot& \cdot & \cdot & \cdot& \cdot & \cdot & \cdot & \cdot & \cdot & I_{d} & -\Sigma_{\rm s}^\frac{1}{2} & \frac{\Sigma_j}{\sqrt{\rho_{\rm s}}} & \cdot\\
  \cdot & \cdot& \cdot &\cdot & \cdot& \cdot &-\frac{Z}{\sqrt{d}} & \cdot & \cdot & \cdot & \cdot & I_{d} &\cdot & \cdot\\
  \cdot & \cdot & \cdot & \cdot & \cdot & \cdot & \cdot & \cdot & \cdot & \cdot & \cdot & \cdot & I_d & -\frac{W_2^\top}{\sqrt{N}}\\
  \cdot & \cdot & \cdot & \cdot & \cdot & \cdot & \cdot & \cdot & \cdot & \cdot & \cdot & \cdot & \cdot & I_N
  \end{bmatrix}}.
\end{align*}
\endgroup

Define the block-wise normalized expected trace of $(Q^{1})^{-1}$ by $G^{1} = (\operatorname{id} \otimes \mathbb{E} \overline{\operatorname{tr}}) ((Q^{1})^{-1})$.
Then, by block matrix inversion we have
\begin{align*}
    G_{2, 14}^{1} = \frac{\psi}{d^2 N^2} \mathbb{E} \operatorname{tr} [ W_1 \Sigma_j W_2^\top F_{2} K_{2}^{-1} Z^\top \Sigma_{\rm s} Z K_{1}^{-1} F_{1}^\top] = \frac{\psi}{2\rho_j} D_{21}.
\end{align*}

We augment $Q^{1}$ to the symmetric matrix $\bar{Q}^{1}$ as
\begin{align*}
    \bar{Q}^{1} = \begin{bmatrix} \cdot & (Q^{1})^\top \\ Q^{1} & \cdot \end{bmatrix}
\end{align*}
and write
\begin{align*}
    \bar{Q}^{1} &= \bar{Z}^{1} - \bar{Q}_{W, Z, \Theta}^{1} - \bar{Q}_\Sigma^{1} \\
    &=\begin{bmatrix} \cdot & I_{2n + 9d + 3N} \\ I_{2n + 9d + 3N} & \cdot \end{bmatrix} - \begin{bmatrix} \cdot & (Q_{W, Z, \Theta}^{1})^\top \\ Q_{W, Z, \Theta}^{1} & \cdot \end{bmatrix} - \begin{bmatrix} \cdot & (Q_\Sigma^{1})^\top \\ Q_\Sigma^{1} & \cdot \end{bmatrix},
\end{align*}
where
\begingroup
\setlength\arraycolsep{2pt}
\begin{align*}
  \scalemath{1.0}{
  Q^{1}_{W, Z, \Theta} = \begin{bmatrix}
 \cdot & -\frac{\sqrt{\rho_{\rm s} \omega_{\rm s}}\Theta_2^\top}{\gamma \sqrt{N}} & -\frac{\sqrt{\rho_{\rm s}}Z^\top}{\gamma \sqrt{d}} & \cdot &\cdot & \cdot& \cdot &\cdot & \cdot& \cdot &\cdot & \cdot &\cdot & \cdot\\
  \frac{\sqrt{\rho_{\rm s} \omega_{\rm s}}\Theta_2}{\sqrt{N}} & \cdot & \cdot & \cdot & \frac{\sqrt{\rho_{\rm s}}W_2}{\sqrt{N}} & \cdot &\cdot & \cdot& \cdot &\cdot & \cdot& \cdot &\cdot & \cdot\\
  \cdot & \cdot & \cdot & \cdot & \cdot & \cdot &\cdot & \cdot& \cdot &\cdot & \cdot& \cdot &\cdot & \cdot\\
  \cdot & \frac{W_2^\top}{\sqrt{N}} & \cdot & \cdot & \cdot & \cdot &\cdot & \cdot& \cdot &\cdot & \cdot& \cdot &\cdot & \cdot\\
  \cdot & \cdot & \cdot & \cdot & \cdot & \cdot &\cdot & \cdot& \cdot &\cdot & \cdot& \cdot &\cdot & \cdot\\
  \frac{Z}{\sqrt{d}} & \cdot & \cdot & \cdot & \cdot & \cdot &\cdot & \cdot& \cdot &\cdot & \cdot& \cdot &\cdot & \cdot\\
  \cdot & \cdot& \cdot &\cdot & \cdot& \cdot & \cdot & -\frac{\sqrt{\rho_{\rm s} \omega_{\rm s}}\Theta_1^\top}{\gamma \sqrt{N}}& -\frac{\sqrt{\rho_{\rm s}} Z^\top}{\gamma \sqrt{d}} &\cdot &\cdot &\cdot &\cdot & \cdot\\ 
  \cdot & \cdot& \cdot &\cdot & \cdot& \cdot & \frac{\sqrt{\rho_{\rm s} \omega_{\rm s}}\Theta_1}{\sqrt{N}} & \cdot & \cdot & \cdot & \frac{\sqrt{\rho_{\rm s}}W_1}{\sqrt{N}} & \cdot &\cdot & \cdot\\ 
  \cdot & \cdot& \cdot &\cdot & \cdot& \cdot & \cdot & \cdot & \cdot & \cdot & \cdot & \cdot &\cdot & \cdot\\ 
  \cdot & \cdot& \cdot &\cdot & \cdot& \cdot & \cdot & \frac{W_1^\top}{\sqrt{N}} & \cdot & \cdot & \cdot & \cdot &\cdot & \cdot\\ 
  \cdot & \cdot& \cdot & \cdot & \cdot& \cdot & \cdot & \cdot & \cdot & \cdot & \cdot & \cdot & \cdot & \cdot\\
  \cdot & \cdot& \cdot &\cdot & \cdot& \cdot &\frac{Z}{\sqrt{d}} & \cdot & \cdot & \cdot & \cdot & \cdot &\cdot & \cdot\\
  \cdot & \cdot & \cdot & \cdot & \cdot & \cdot & \cdot & \cdot & \cdot & \cdot & \cdot & \cdot & \cdot & \frac{W_2^\top}{\sqrt{N}}\\
  \cdot & \cdot & \cdot & \cdot & \cdot & \cdot & \cdot & \cdot & \cdot & \cdot & \cdot & \cdot & \cdot & \cdot
  \end{bmatrix}}
\end{align*}
\endgroup
and
\begin{align*}
  Q_{\Sigma}^{1} = \begin{bmatrix} \cdot & \cdot & \cdot & \cdot & \cdot & \cdot & \cdot & \cdot & \cdot & \cdot & \cdot & \cdot & \cdot & \cdot \\
  \cdot & \cdot & \cdot & \cdot & \cdot & \cdot & \cdot & \cdot & \cdot & \cdot & \cdot & \cdot & \cdot & \cdot \\
  \cdot & \cdot & \cdot & \Sigma_{\rm s}^\frac{1}{2} & \cdot & \cdot & \cdot & \cdot & \cdot & \cdot & \cdot & \cdot & \cdot & \cdot \\
  \cdot & \cdot & \cdot & \cdot & \cdot & \cdot & \cdot & \cdot & \cdot & \cdot & \cdot & -\frac{\Sigma_{\rm s}^\frac{1}{2}}{\sqrt{\rho_{\rm s}}} & \cdot & \cdot \\
  \cdot & \cdot & \cdot & \cdot & \cdot & \Sigma_{\rm s}^\frac{1}{2} & \cdot & \cdot & \cdot & \cdot & \cdot & \cdot & \cdot & \cdot \\
  \cdot & \cdot & \cdot & \cdot & \cdot & \cdot & \cdot & \cdot & \cdot & \cdot & \cdot & \cdot & \cdot & \cdot \\
  \cdot & \cdot & \cdot & \cdot & \cdot & \cdot & \cdot & \cdot & \cdot & \cdot & \cdot & \cdot & \cdot & \cdot\\
  \cdot & \cdot & \cdot & \cdot & \cdot & \cdot & \cdot & \cdot & \cdot & \cdot & \cdot & \cdot & \cdot & \cdot\\
  \cdot & \cdot & \cdot & \cdot & \cdot & \cdot & \cdot & \cdot & \cdot & \Sigma_{\rm s}^\frac{1}{2} & \cdot & \cdot & \cdot & \cdot \\
  \cdot & \cdot & \cdot & \cdot & \cdot & \cdot & \cdot & \cdot & \cdot & \cdot & \cdot & \cdot & \cdot & \cdot \\
  \cdot & \cdot & \cdot & \cdot & \cdot & \cdot & \cdot & \cdot & \cdot & \cdot & \cdot & \Sigma_{\rm s}^\frac{1}{2} & -\frac{\Sigma_j}{\sqrt{\rho_{\rm s}}} & \cdot \\
  \cdot & \cdot & \cdot & \cdot & \cdot & \cdot & \cdot & \cdot & \cdot & \cdot & \cdot & \cdot & \cdot & \cdot \\
  \cdot & \cdot & \cdot & \cdot & \cdot & \cdot & \cdot & \cdot & \cdot & \cdot & \cdot & \cdot & \cdot & \cdot \\
  \cdot & \cdot & \cdot & \cdot & \cdot & \cdot & \cdot & \cdot & \cdot & \cdot & \cdot & \cdot & \cdot & \cdot\end{bmatrix}.
\end{align*}
Then defining $\bar{G}^{1}$ below,
\begin{align*}
    \bar{G}^{1} &= \begin{bmatrix} \cdot & G^{1} \\ (G^{1})^\top & \cdot \end{bmatrix} = \begin{bmatrix} \cdot & (\operatorname{id} \otimes \mathbb{E} \overline{\operatorname{tr}}) ((Q^{1})^{-1}) \\ (\operatorname{id} \otimes \mathbb{E} \overline{\operatorname{tr}}) (((Q^{1})^\top)^{-1}) & \cdot \end{bmatrix}\\
    &= (\operatorname{id} \otimes \mathbb{E} \overline{\operatorname{tr}}) \begin{bmatrix} \cdot & (Q^{1})^{-1} \\ ((Q^{1})^\top)^{-1} & \cdot \end{bmatrix} = (\operatorname{id} \otimes \mathbb{E} \overline{\operatorname{tr}}) ((\bar{Q}^{1})^{-1})
\end{align*}
can be viewed as the operator-valued Cauchy transform of $\bar{Q}_{W, Z, \Theta}^{1} + \bar{Q}_\Sigma^{1}$ (in the space we consider in Remark \ref{rmk:rectangular}), i.e.,
\begin{align*}
    \bar{G}^{1} = (\operatorname{id} \otimes \mathbb{E} \overline{\operatorname{tr}})(\bar Z^{1} - \bar Q_{W, Z, \Theta}^{1} - \bar Q_\Sigma^{1})^{-1} = \mathcal{G}_{\bar{Q}_{W, Z, \Theta}^{1} + \bar{Q}_\Sigma^{1}}(\bar{Z}^{1}).
\end{align*}
Further by the subordination formula \eqref{eqn:subordination},
\begin{align}\label{eqn:d21subordination}
    \bar{G}^{1} &= \mathcal{G}_{\bar{Q}_\Sigma^{1}}(\bar{Z}^{1} - \mathcal{R}_{\bar{Q}_{W, Z, \Theta}^{1}}(\bar{G}^{1})) = (\operatorname{id} \otimes \mathbb{E} \overline{\operatorname{tr}})(\bar{Z}^{1} - \mathcal{R}_{\bar{Q}_{W, Z, \Theta}^{1}}(\bar{G}^{1}) - \bar{Q}_\Sigma^{1})^{-1}.
\end{align}
Since $\bar{Q}_{W, Z, \Theta}^{1}$ consists of i.i.d. Gaussian blocks, by \eqref{eqn:gaussianrtransform}, its limiting $R$-transform has a form
\begin{align*}
    \mathcal{R}_{\bar{Q}_{W, Z, \Theta}^{1}}(\bar{G}^{1}) = \begin{bmatrix} \cdot & (R^{1})^\top \\ R^{1} & \cdot \end{bmatrix},
\end{align*}
where the non-zero blocks of $R^{1}$ are
\begin{align*}
    &R_{1, 1}^{1} = -\frac{\rho_{\rm s} \omega_{\rm s}}{\gamma} G_{2,2}^{1} - \frac{\sqrt{\rho_{\rm s}}}{\gamma}G_{3, 6}^{1}, \quad R_{1, 7}^{1} = -\frac{\sqrt{\rho_{\rm s}}}{\gamma} G_{3, 12}^{1}, \quad R_{2, 2}^{1} = -\frac{\psi \rho_{\rm s} \omega_{\rm s}}{\gamma \phi}G_{1, 1}^{1} + \sqrt{\rho_{\rm s}} \psi G_{5, 4}^{1},\\
    &R_{2, 14}^{1} = \sqrt{\rho_{\rm s}}\psi G_{5, 13}^{1}, \quad R_{4, 5}^{1} = \sqrt{\rho_{\rm s}} G_{2, 2}^{1}, \quad
    R_{6, 3}^{1} = -\frac{\sqrt{\rho_{\rm s}}}{\gamma \phi} G_{1, 1}^{1}, \quad R_{6, 9}^{1} = -\frac{\sqrt{\rho_{\rm s}}}{\gamma \phi} G_{1, 7}^{1}, \\
    &R_{7, 1}^{1} = -\frac{\sqrt{\rho_{\rm s}}}{\gamma} G_{9, 6}^{1} = 0, \quad R_{7, 7}^{1} = -\frac{\rho_{\rm s} \omega_{\rm s}}{\gamma} G_{8, 8}^{1} - \frac{\sqrt{\rho_{\rm s}}}{\gamma} G_{9, 12}^{1}, \quad R_{8, 8}^{1} = -\frac{\psi\rho_{\rm s} \omega_{\rm s}}{\gamma \phi} G_{7, 7}^{1} + \sqrt{\rho_{\rm s}}\psi G_{11, 10}^{1},\\
    &R_{10, 11}^{1} = \sqrt{\rho_{\rm s}} G_{8, 8}^{1}, \quad R_{12, 3}^{1} = -\frac{\sqrt{\rho_{\rm s}}}{\gamma \phi} G_{7, 1}^{1} = 0, \quad R_{12, 9}^{1} = -\frac{\sqrt{\rho_{\rm s}}}{\gamma \phi}G_{7, 7}^{1}, \quad R_{13, 5}^{1} = \sqrt{\rho_{\rm s}} G_{14, 2}^{1} = 0.
\end{align*}
We used the fact that $G_{9, 6}^1 = G_{7, 1}^1 = G_{14, 2}^1 = 0$, which we obtain from block matrix inversion of $Q^1$.

Computing the block-matrix inverse of $Q^{1}$ and from equations \eqref{eqn:g0blockinversion}, \eqref{eqn:g0values}, we see
\begin{align*}
    G_{1, 1}^{1} &= G_{7, 7}^{1} = \gamma \E \overline{\operatorname{tr}}(K^{-1}) = G_{1, 1}^0 = \gamma \tau, \quad G_{2, 2}^{1} = G_{8, 8}^{1} = \gamma \E \overline{\operatorname{tr}}(\hat{K}^{-1}) = G_{2, 2}^0 = \gamma \bar \tau,\\
    G_{3, 6}^{1} &= G_{9, 12}^{1} = \frac{\gamma \sqrt{\rho_{\rm s}}\,\E \overline{\operatorname{tr}} [\Sigma_{\rm s} W^\top \hat{K}^{-1} W]}{N} = G_{3, 6}^0 =  \gamma \sqrt{\rho_{\rm s}}  \bar\tau \mathcal{I}_{1, 1}^{\rm s},\\
    G_{5, 4}^{1} &= G_{11, 10}^{1} = -\frac{\sqrt{\rho_{\rm s}}\, \E \overline{\operatorname{tr}} [\Sigma_{\rm s} Z K^{-1} Z^\top]}{d}= G_{5, 4}^0 =  -\frac{\sqrt{\rho_{\rm s}}\tau \mathcal{I}_{1, 1}^{\rm s}}{\phi}.
\end{align*}
Plugging these into \eqref{eqn:d21subordination}, we obtain self-consistent equations.
For example,
\begin{align*}
G_{2, 14}^{1} &= \mathbb{E} \overline{\operatorname{tr}} [(I_{2n + 9d + 3N} - R^{1} - Q_\Sigma^{1})^{-1}]_{2, 14}\\
&= -\frac{\gamma \sqrt{\rho_{\rm s}} \psi \phi  G_{5, 13}^{1}}{\gamma \phi  (-1 + \sqrt{\rho_{\rm s}} \psi  G_{5, 4}^{1}) - \psi \rho_{\rm s} \omega_{\rm s} G_{1, 1}^{1}} = \gamma \sqrt{\rho_{\rm s}} \bar \tau \psi G_{5, 13}^{1}.
\end{align*}
Similarly,
\begin{align*}
    G_{5, 13}^{1} 
    &= \mathbb{E}_\mu \left[ \frac{ \lambda^{\rm s} \lambda^j  \gamma \sqrt{\rho_{\rm s}} \phi G_{1, 7}^{1} G_{2,2}^{1} +  (\lambda^{\rm s})^2 \lambda^j \sqrt{\rho_{\rm s}} (G_{1, 1}^{1})^2  G_{2, 2}^{1}}{(\gamma \phi  + \lambda^{\rm s} \rho_{\rm s} G_{1, 1}^{1} G_{2, 2}^{1} )^2} \right] = \sqrt{\rho_{\rm s}} \bar \tau  \mathcal{I}_{2, 2}^j G_{1, 7}^{1} + \frac{\gamma \sqrt{\rho_{\rm s}} \tau^2 \bar \tau}{\phi} \mathcal{I}_{3, 2}^j,\\
    G_{1, 7}^{1} &= -\frac{\gamma \sqrt{\rho_{\rm s}} G_{3, 12}^{1}}{(\gamma + \sqrt{\rho_{\rm s}} G_{3, 6}^{1} + \rho_{\rm s} \omega_{\rm s} G_{2, 2}^{1})^2} = -\gamma \sqrt{\rho_{\rm s}} \tau^2 G_{3, 12}^{1},\\
    G_{3, 12}^{1} &= -\mathbb{E}_\mu \left[ \frac{\lambda^{\rm s} \gamma^2 \phi^2 + (\lambda^{\rm s})^2 \gamma \rho_{\rm s}^2 \phi  G_{1, 7}^{1} (G_{2, 2}^{1})^2}{\sqrt{\rho_{\rm s}}(\gamma \phi  + \lambda^{\rm s} \rho_{\rm s}  G_{1, 1}^{1} G_{2, 2}^{1} )^2} \right] = -\frac{\phi}{\sqrt{\rho_{\rm s}}} \mathcal{I}_{1, 2}^{\rm s} - \gamma \rho_{\rm s}^\frac{3}{2} \bar \tau^2 \mathcal{I}_{2, 2}^{\rm s} G_{1, 7}^{1}.
\end{align*}
Eliminating $G_{3, 12}^{1}$ and using $\kappa = \gamma \rho_{\rm s} \tau \bar \tau$,
\begin{align*}
    &G_{1, 7}^{1} =  \gamma  \tau^2 \phi \mathcal{I}_{1, 2}^{\rm s} + \kappa^2 \mathcal{I}_{2, 2}^{\rm s} G_{1, 7}^{1}  \Rightarrow G_{1, 7}^{1} = \frac{\gamma \tau^2 \phi \mathcal{I}_{1, 2}^{\rm s}}{1 - \kappa^2 \mathcal{I}_{2, 2}^{\rm s}}.
\end{align*}
Therefore,
\begin{align*}
    G_{2, 14}^{1} &= \gamma \sqrt{\rho_{\rm s}} \bar\tau \psi G_{5, 13}^{1} = \gamma \rho_{\rm s} \bar\tau^2 \psi \mathcal{I}_{2, 2}^j G_{1, 7}^{1} + \frac{\gamma^2 \rho_{\rm s} \tau^2 \bar\tau^2 \psi}{\phi} \mathcal{I}_{3,2}^j\\
    &= \frac{\gamma^2 \rho_{\rm s} \tau^2 \bar\tau^2 \psi \phi \mathcal{I}_{1,2}^{\rm s} \mathcal{I}_{2, 2}^j}{1 - \kappa^2 \mathcal{I}_{2, 2}^{\rm s}} + \frac{\gamma^2 \rho_{\rm s} \tau^2 \bar\tau^2 \psi}{\phi} \mathcal{I}_{3,2}^j = \kappa^2 \left( \frac{\psi \phi \mathcal{I}_{1, 2}^{\rm s} \mathcal{I}_{2, 2}^j}{\rho_{\rm s}(1 - \kappa^2 \mathcal{I}_{2, 2}^{\rm s})} + \frac{\psi \mathcal{I}_{3, 2}^j}{\rho_{\rm s} \phi} \right).
\end{align*}
Finally,
\begin{align}\label{eqn:d21}
    D_{21} = \frac{2 \rho_j}{\psi} G_{2, 14}^{1} = \frac{2 \rho_j \kappa^2}{\rho_{\rm s}} \left( \frac{ \phi \mathcal{I}_{1, 2}^{\rm s} \mathcal{I}_{2, 2}^j}{1 - \kappa^2 \mathcal{I}_{2, 2}^{\rm s}} + \frac{ \mathcal{I}_{3, 2}^j}{\phi} \right).
\end{align}

\subsubsection{Computation of $D_{22}$}
Let
\begingroup
\setlength\arraycolsep{0pt}
\begin{align*}
  \scalemath{1.0}{
  Q^{2} = \begin{bmatrix}
  I_n & \frac{\sqrt{\rho_{\rm s} \omega_{\rm s}}\Theta_2^\top}{\gamma \sqrt{N}} & \frac{\sqrt{\rho_{\rm s}}Z^\top}{\gamma \sqrt{d}} & \cdot &\cdot & \cdot& \cdot &\cdot & \cdot& \cdot &\cdot & \cdot\\
  -\frac{\sqrt{\rho_{\rm s} \omega_{\rm s}}\Theta_2}{\sqrt{N}} & I_{N} & \cdot & \cdot & -\frac{\sqrt{\rho_{\rm s}}W_2}{\sqrt{N}} & \cdot &\cdot & \cdot& \cdot &\cdot & \cdot& \cdot\\
  \cdot & \cdot & I_{d} & -\Sigma_{\rm s}^\frac{1}{2} & \cdot & \cdot &\cdot & \cdot& \cdot &\cdot & \cdot& \cdot \\
  \cdot & -\frac{W_2^\top}{\sqrt{N}} & \cdot & I_{d} & \cdot & \cdot &\cdot & \cdot& \cdot &\cdot & \cdot& \cdot \\
  \cdot & \cdot & \cdot & \cdot & I_{d} & -\Sigma_{\rm s}^\frac{1}{2} &\cdot & \cdot& \cdot &\cdot & \cdot& \cdot \\
  -\frac{Z}{\sqrt{d}} & \cdot & \cdot & \cdot & \cdot & I_{d} &\cdot & \cdot& \cdot &\cdot & \cdot& \cdot\\
  \cdot & \cdot& \cdot &\cdot & \cdot& \cdot & I_n & \frac{\sqrt{\rho_{\rm s} \omega_{\rm s}}\Theta_1^\top}{\gamma \sqrt{N}}& \frac{\sqrt{\rho_{\rm s}} Z^\top}{\gamma \sqrt{d}} &\cdot &\cdot &\cdot\\ 
  \cdot & \cdot& \cdot &\cdot & \cdot& \cdot & -\frac{\sqrt{\rho_{\rm s} \omega_{\rm s}}\Theta_1}{\sqrt{N}} & I_{N} & \cdot & \cdot & -\frac{\sqrt{\rho_{\rm s}}W_1}{\sqrt{N}} & \cdot\\ 
  \cdot & \cdot& \cdot &\cdot & \cdot& \cdot & \cdot & \cdot & I_{d} & -\Sigma_{\rm s}^\frac{1}{2} & \cdot & \cdot\\ 
  \cdot & \cdot& \cdot &\cdot & \cdot& \cdot & \cdot & -\frac{W_1^\top}{\sqrt{N}} & \cdot & I_{d} & \cdot & \cdot\\ 
  \cdot & \cdot& \cdot &\Sigma_{j} & \cdot& \cdot & \cdot & \cdot & \cdot & \cdot & I_{d} & -\Sigma_{\rm s}^\frac{1}{2}\\
  \cdot & \cdot& \cdot &\cdot & \cdot& \cdot &-\frac{Z}{\sqrt{d}} & \cdot & \cdot & \cdot & \cdot & I_{d}
  \end{bmatrix}}
\end{align*}
\endgroup
and $G^{2} = (\operatorname{id} \otimes \mathbb{E} \overline{\operatorname{tr}}) ((Q^{2})^{-1})$.
Then,
\begin{align*}
    G_{7, 1}^{2} = \frac{\gamma \sqrt{\rho_{\rm s}} \phi}{d N^2} \E \operatorname{tr}[K_1^{-1}F_1^\top W_1 \Sigma_{j} W_2^\top F_2 K_2^{-1}] = \frac{\gamma \sqrt{\rho_{\rm s}} \phi}{2\sigma_\ep^2 \rho_j} D_{22}.
\end{align*}
We augment $Q^{2}$ to the symmetric matrix $\bar{Q}^{2}$ as
\begin{align*}
    \bar{Q}^{2} = \begin{bmatrix} \cdot & (Q^{2})^\top \\ Q^{2} & \cdot \end{bmatrix}
\end{align*}
and write
\begin{align*}
    \bar{Q}^{2} &= \bar{Z}^{2} - \bar{Q}_{W, Z, \Theta}^{2} - \bar{Q}_\Sigma^{2} \\
    &=\begin{bmatrix} \cdot & I_{2n + 8d + 2N} \\ I_{2n + 8d + 2N} & \cdot \end{bmatrix} - \begin{bmatrix} \cdot & (Q_{W, Z, \Theta}^{2})^\top \\ Q_{W, Z, \Theta}^{2} & \cdot \end{bmatrix} - \begin{bmatrix} \cdot & (Q_\Sigma^{2})^\top \\ Q_\Sigma^{2} & \cdot \end{bmatrix},
\end{align*}
where
\begingroup
\setlength\arraycolsep{2pt}
\begin{align*} Q_{W, Z, \Theta}^{2} = 
  \begin{bmatrix}
  \cdot & -\frac{\sqrt{\rho_{\rm s} \omega_{\rm s}}\Theta_2^\top}{\gamma \sqrt{N}} & -\frac{\sqrt{\rho_{\rm s}}Z^\top}{\gamma \sqrt{d}} & \cdot &\cdot & \cdot& \cdot &\cdot & \cdot& \cdot &\cdot & \cdot\\
  \frac{\sqrt{\rho_{\rm s} \omega_{\rm s}}\Theta_2}{\sqrt{N}} & \cdot & \cdot & \cdot & \frac{\sqrt{\rho_{\rm s}}W_2}{\sqrt{N}} & \cdot &\cdot & \cdot& \cdot &\cdot & \cdot& \cdot\\
  \cdot & \cdot & \cdot & \cdot & \cdot & \cdot &\cdot & \cdot& \cdot &\cdot & \cdot& \cdot \\
  \cdot & \frac{W_2^\top}{\sqrt{N}} & \cdot & \cdot & \cdot & \cdot &\cdot & \cdot& \cdot &\cdot & \cdot& \cdot \\
  \cdot & \cdot & \cdot & \cdot & \cdot & \cdot &\cdot & \cdot& \cdot &\cdot & \cdot& \cdot \\
  \frac{Z}{\sqrt{d}} & \cdot & \cdot & \cdot & \cdot & \cdot &\cdot & \cdot& \cdot &\cdot & \cdot& \cdot\\
  \cdot & \cdot& \cdot &\cdot & \cdot& \cdot & \cdot & -\frac{\sqrt{\rho_{\rm s} \omega_{\rm s}}\Theta_1^\top}{\gamma \sqrt{N}}& -\frac{\sqrt{\rho_{\rm s}} Z^\top}{\gamma \sqrt{d}} &\cdot &\cdot &\cdot\\ 
  \cdot & \cdot& \cdot &\cdot & \cdot& \cdot & \frac{\sqrt{\rho_{\rm s} \omega_{\rm s}}\Theta_1}{\sqrt{N}} & \cdot & \cdot & \cdot & \frac{\sqrt{\rho_{\rm s}}W_1}{\sqrt{N}} & \cdot\\ 
  \cdot & \cdot& \cdot &\cdot & \cdot& \cdot & \cdot & \cdot & \cdot & \cdot & \cdot & \cdot\\ 
  \cdot & \cdot& \cdot &\cdot & \cdot& \cdot & \cdot & \frac{W_1^\top}{\sqrt{N}} & \cdot & \cdot & \cdot & \cdot\\ 
  \cdot & \cdot& \cdot & \cdot & \cdot& \cdot & \cdot & \cdot & \cdot & \cdot & \cdot & \cdot\\
  \cdot & \cdot& \cdot &\cdot & \cdot& \cdot &\frac{Z}{\sqrt{d}} & \cdot & \cdot & \cdot & \cdot & \cdot
  \end{bmatrix},
\end{align*}
\endgroup
and
\begin{align*}
    Q_{\Sigma}^{2} = \begin{bmatrix}
    \cdot & \cdot & \cdot & \cdot &\cdot & \cdot& \cdot &\cdot & \cdot& \cdot &\cdot & \cdot\\
    \cdot & \cdot & \cdot & \cdot & \cdot & \cdot &\cdot & \cdot& \cdot &\cdot & \cdot& \cdot\\
    \cdot & \cdot & \cdot & \Sigma_{\rm s}^\frac{1}{2} & \cdot & \cdot &\cdot & \cdot& \cdot &\cdot & \cdot& \cdot \\
    \cdot & \cdot & \cdot & \cdot & \cdot & \cdot &\cdot & \cdot& \cdot &\cdot & \cdot& \cdot \\
    \cdot & \cdot & \cdot & \cdot & \cdot & \Sigma_{\rm s}^\frac{1}{2} &\cdot & \cdot& \cdot &\cdot & \cdot& \cdot \\
    \cdot & \cdot & \cdot & \cdot & \cdot & \cdot &\cdot & \cdot& \cdot &\cdot & \cdot& \cdot\\
    \cdot & \cdot& \cdot &\cdot & \cdot& \cdot & \cdot & \cdot& \cdot &\cdot &\cdot &\cdot\\ 
    \cdot & \cdot& \cdot &\cdot & \cdot& \cdot & \cdot & \cdot & \cdot & \cdot & \cdot & \cdot\\ 
    \cdot & \cdot& \cdot &\cdot & \cdot& \cdot & \cdot & \cdot & \cdot & \Sigma_{\rm s}^\frac{1}{2} & \cdot & \cdot\\ 
    \cdot & \cdot& \cdot &\cdot & \cdot& \cdot & \cdot & \cdot & \cdot & \cdot & \cdot & \cdot\\ 
    \cdot & \cdot& \cdot & -\Sigma_{j} & \cdot& \cdot & \cdot & \cdot & \cdot & \cdot & \cdot & \Sigma_{\rm s}^\frac{1}{2}\\
    \cdot & \cdot& \cdot &\cdot & \cdot& \cdot &\cdot & \cdot & \cdot & \cdot & \cdot & \cdot
    \end{bmatrix}.
\end{align*}
Defining $\bar{G}^2$ below,
\begin{align*}
    \bar{G}^{2} &= \begin{bmatrix} \cdot & G^{2} \\ (G^{2})^\top & \cdot \end{bmatrix} = \begin{bmatrix} \cdot & (\operatorname{id} \otimes \mathbb{E} \overline{\operatorname{tr}}) ((Q^{2})^{-1}) \\ (\operatorname{id} \otimes \mathbb{E} \overline{\operatorname{tr}}) (((Q^{2})^\top)^{-1}) & \cdot\end{bmatrix}\\
    &= (\operatorname{id} \otimes \mathbb{E} \overline{\operatorname{tr}}) \begin{bmatrix} \cdot & (Q^{2})^{-1} \\ ((Q^{2})^\top)^{-1} & \cdot\end{bmatrix} = (\operatorname{id} \otimes \mathbb{E} \overline{\operatorname{tr}}) ((\bar{Q}^{2})^{-1}).
\end{align*}
It can be viewed as the operator-valued Cauchy transform of $\bar{Q}_{W, Z, \Theta}^{2} + \bar{Q}_\Sigma^{2}$ (in the space we consider in Remark \ref{rmk:rectangular}), i.e.,
\begin{align*}
    \bar{G}^{2} = (\operatorname{id} \otimes \mathbb{E} \overline{\operatorname{tr}})(\bar Z^{2} - \bar Q_{W, Z, \Theta}^{2} - \bar Q_\Sigma^{2})^{-1} = \mathcal{G}_{\bar{Q}_{W, Z, \Theta}^{2} + \bar{Q}_\Sigma^{2}}(\bar{Z}^{2}).
\end{align*}
Further by the subordination formula \eqref{eqn:subordination},
\begin{align}\label{eqn:d22subordination}
    \bar{G}^{2} &= \mathcal{G}_{\bar{Q}_\Sigma^{2}}(\bar{Z}^{2} - \mathcal{R}_{\bar{Q}_{W, Z, \Theta}^{2}}(\bar{G}^{2})) = (\operatorname{id} \otimes \mathbb{E} \overline{\operatorname{tr}})(\bar{Z}^{2} - \mathcal{R}_{\bar{Q}_{W, Z, \Theta}^{2}}(\bar{G}^{2}) - \bar{Q}_\Sigma^{2})^{-1}.
\end{align}
Since $\bar{Q}_{W, Z, \Theta}^{2}$ consists of i.i.d. Gaussian blocks, by \eqref{eqn:gaussianrtransform}, its limiting $R$-transform has a form
\begin{align*}
    \mathcal{R}_{\bar{Q}_{W, Z, \Theta}^{2}}(\bar{G}^{2}) = \begin{bmatrix} \cdot & (R^{2})^\top \\ R^{2} & \cdot \end{bmatrix},
\end{align*}
where the non-zero blocks of $R^{2}$ are
\begin{align*}
    &R_{1, 1}^{2} = -\frac{\rho_{\rm s} \omega_{\rm s}}{\gamma} G_{2, 2}^{2} - \frac{\sqrt{\rho_{\rm s}}}{\gamma} G_{3, 6}^{2}, \quad R_{1, 7}^{2} = -\frac{\sqrt{\rho_{\rm s}}}{\gamma} G_{3, 12}^{2} = 0, \quad R_{2, 2}^{2} = -\frac{\psi\rho_{\rm s} \omega_{\rm s}}{\gamma \phi} G_{1, 1}^{2} + \sqrt{\rho_{\rm s}} \psi G_{5, 4}^{2},\\
    &R_{4, 5}^{2} = \sqrt{\rho_{\rm s}} G_{2, 2}^{2}, \quad R_{6, 3}^{2} = -\frac{\sqrt{\rho_{\rm s}}}{\gamma \phi} G_{1, 1}^{2}, \quad R_{6, 9}^{2} = -\frac{\sqrt{\rho_{\rm s}}}{\gamma \phi}G_{1, 7}^{2} = 0, \quad R_{7, 1}^{2} = -\frac{\sqrt{\rho_{\rm s}}}{\gamma} G_{9, 6}^{2},\\
    &R_{7, 7}^{2} = -\frac{\rho_{\rm s} \omega_{\rm s}}{\gamma} G_{8, 8}^{2} - \frac{\sqrt{\rho_{\rm s}}}{\gamma} G_{9, 12}^{2}, \quad R_{8, 8}^{2} = -\frac{\psi\rho_{\rm s} \omega_{\rm s}}{\gamma \phi} G_{7, 7}^{2} + \sqrt{\rho_{\rm s}} \psi G_{11, 10}^{2}, \quad  R_{10, 11}^{2} = \sqrt{\rho_{\rm s}} G_{8, 8}^{2},\\
    &R_{12, 3}^{2} = -\frac{\sqrt{\rho_{\rm s}}}{\gamma \phi} G_{7, 1}^{2}, \quad R_{12, 9}^{2} = -\frac{\sqrt{\rho_{\rm s}}}{\gamma \phi}G_{7, 7}^{2}.
\end{align*}
We used the fact that $G_{3, 12}^2 = G_{1, 7}^2 = 0$, which we obtain from block matrix inversion of $Q^2$.
From block matrix inversion of $Q^{2}$ and equations \eqref{eqn:g0blockinversion}, \eqref{eqn:g0values}, we have
\begin{align*}
G_{1, 1}^{2} &= G_{7, 7}^{2} = \gamma \E \overline{\operatorname{tr}}(K^{-1}) = G_{1, 1}^0 = \gamma \tau, \quad G_{2, 2}^{2} = G_{8, 8}^{2} = \gamma \E \overline{\operatorname{tr}}(\hat{K}^{-1}) = G_{2, 2}^0 = \gamma \bar \tau,\\
G_{3, 6}^{2} &= G_{9, 12}^{2} = \frac{\gamma \sqrt{\rho_{\rm s}}\,\E \overline{\operatorname{tr}} [\Sigma_{\rm s} W^\top \hat{K}^{-1} W]}{N} = G_{3, 6}^0 = \gamma \sqrt{\rho_{\rm s}} \bar \tau \mathcal{I}_{1, 1}^{\rm s},\\
G_{5, 4}^{2} &= G_{11, 10}^{2} = -\frac{\sqrt{\rho_{\rm s}}\, \E \overline{\operatorname{tr}} [\Sigma_{\rm s} Z K^{-1} Z^\top]}{d}  = G_{5, 4}^0 = -\frac{\sqrt{\rho_{\rm s}} \tau \mathcal{I}_{1, 1}^{\rm s}}{\phi}.
\end{align*}
Plugging these into \eqref{eqn:d22subordination}, we have the following self-consistent equations
\begin{align*}
    G_{7, 1}^{2} &= - \frac{\gamma \sqrt{\rho_{\rm s}} G_{9 ,6}^{2}}{(\gamma + \sqrt{\rho_{\rm s}} G_{3, 6}^{2} + \rho_{\rm s} \omega_{\rm s} G_{2,2}^{2})^2} = -\gamma \sqrt{\rho_{\rm s}} \tau^2 G_{9 ,6}^{2},\\
    G_{9, 6}^{2} &= - \mathbb{E}_\mu \left[ \frac{(\lambda^{\rm s})^2 \gamma \rho_{\rm s}^\frac{3}{2} \phi (G_{2, 2}^{2})^2 G_{7, 1}^{2} + \lambda^{\rm s} \lambda^j \gamma^2 \rho_{\rm s} \phi^2 (G_{2, 2}^{2})^2}{(\gamma \phi + \lambda^{\rm s} \rho_{\rm s} G_{1,1}^{2} G_{2,2}^{2})^2} \right] = -\gamma \rho_{\rm s}^\frac{3}{2} \bar\tau^2 \mathcal{I}_{2, 2}^{\rm s} G_{7, 1}^{2} - \gamma^2 \rho_{\rm s} \bar\tau^2 \phi \mathcal{I}_{2, 2}^j.
\end{align*}
Solving for $G_{7, 1}^{2}$,
\begin{align*}
    G_{7, 1}^{2} = \frac{\kappa^2 \gamma \phi \mathcal{I}_{2, 2}^j}{\sqrt{\rho_{\rm s}}(1 - \kappa^2 \mathcal{I}_{2, 2}^{\rm s})}.
\end{align*}
Therefore,
\begin{align}\label{eqn:d22}
    D_{22} = \frac{2 \sigma_\ep^2 \rho_j}{\gamma \sqrt{\rho_{\rm s}} \phi} G_{7, 1}^{2} = \frac{2 \rho_j \kappa^2 \sigma_\ep^2 \mathcal{I}_{2, 2}^j}{\rho_{\rm s}(1 - \kappa^2\mathcal{I}_{2, 2}^{\rm s})}.
\end{align}

\subsubsection{Computation of $\stsd_{\textnormal{SS}}^j (\phi, \psi, \gamma)$}
Combining equations \eqref{eqn:inddecompose}, \eqref{eqn:d1}, \eqref{eqn:d2}, \eqref{eqn:d21}, \eqref{eqn:d22}, we get
\begin{align*}
    \stsd_\textnormal{SS}^j (\phi, \psi, \gamma) = \stsd_\textnormal{I}^j (\phi, \psi, \gamma) -  \frac{2\rho_j\kappa^2(\sigma_\ep^2 + \phi \cI^{\rm s}_{1, 2}) \cI^j_{2, 2}}{\rho_{\rm s}(1 - \kappa^2 \cI_{2, 2}^{\rm s})}.
\end{align*}

\subsubsection{Decomposition of $\stsd_{\textnormal{SW}}^j (\phi, \psi, \gamma)$}
Writing $F_i = \sigma(W X_i / \sqrt{d})$, $f = \sigma(W x / \sqrt{d})$, $K_i = \frac{1}{N} F_i^\top F_i + \gamma I_n$ for $i \in \{1, 2\}$, we can write SW disagreement as
\begin{align}\label{eqn:newdecompose}
\stsd_{\textnormal{SW}}^j (\phi, \psi, \gamma) &= \frac{1}{N^2} \E [(Y_1^\top K_1^{-1} F_1^\top f - Y_2^\top K_2^{-1} F_2^\top f)^2] \nonumber \\
    &= \frac{2}{N^2} \E[f^\top F_1 K_1^{-1} Y_1 Y_1^\top K_1^{-1} F_1^\top f] - \frac{2}{N^2} \E[f^\top F_2 K_2^{-1} Y_2 Y_1^\top K_1^{-1} F_1^\top f ] \nonumber\\
    &= D_1 - D_3.
\end{align}
The term $D_1$ is given in \eqref{eqn:d1}.
Plugging in $Y_i = X_i^\top \beta /\sqrt{d} + \boldsymbol{\varepsilon}_i$, where $\boldsymbol{\varepsilon}_i = (\ep_{i1}, \dots, \ep_{in})^\top \in \R^n$, the term $D_3$ becomes
\begin{align*}
    D_3 = &\frac{2}{d N^2} \E_{W, X_i}\operatorname{tr}[F_2 K_2^{-1} X_2^\top \E_\beta [\beta \beta^\top] X_1  K_1^{-1} F_1^\top \E_{x \sim \cD_j, \theta}[f f^\top] ] \\
    &+ \frac{4}{\sqrt{d}N^2} \E_{W, X_i}[ F_2 K_2^{-1} X_2^\top \E_{\beta, \boldsymbol{\varepsilon}_1} [\beta \boldsymbol{\varepsilon}_1^\top]  K_1^{-1} F_1^\top \E_{x \sim \cD_j, \theta}[ff^\top] ]\\
    &+ \frac{2}{N^2} \E_{W, X_i} \operatorname{tr} [F_2 K_2^{-1} \E_{\boldsymbol{\varepsilon}_i} [\boldsymbol{\varepsilon}_2 \boldsymbol{\varepsilon}_1^\top]  K_1^{-1} F_1^\top \E_{x \sim \cD_j, \theta}[f f^\top] ]\\
    = &\frac{2}{d N^2} \E_{W, X_i}\operatorname{tr}[ F_2 K_2^{-1} X_2^\top X_1  K_1^{-1} F_1^\top \E_{x \sim \cD_j, \theta}[f f^\top] ].
\end{align*}
From the Gaussian equivalence \eqref{eqn:ge}, we have
\begin{align*}
    \E_{x \sim \cD_j, \theta}[f f^\top] = \frac{\rho_j}{d} W \Sigma_j W^\top + \rho_j \omega_j I_N.
\end{align*}
Therefore,
\begin{align}\label{eqn:d3}
    D_3 &=  \frac{2\rho_j}{d^2 N^2} \E_{W, X_i}\operatorname{tr} [W \Sigma_j W^\top F_2 K_2^{-1} X_2^\top X_1 K_1^{-1} F_1^\top] + \frac{2\rho_j \omega_j}{d N^2} \E_{W, X_i}\operatorname{tr}\left[ F_2 K_2^{-1} X_2^\top X_1  K_1^{-1} F_1^\top  \right] \nonumber \\
    &= D_{31} + D_{32}.
\end{align}
We can write $X_i = \Sigma_{\rm s}^\frac{1}{2} Z_i$ for $Z_i \in \R^{d \times n}$ with i.i.d. standard Gaussian entries.
Thus,
\begin{align*}
    D_{31} &= \frac{2\rho_j}{d^2 N^2} \E_{W, Z_i}\operatorname{tr} [W \Sigma_j W^\top F_2 K_2^{-1} Z_2^\top \Sigma_{\rm s} Z_1 K_1^{-1} F_1^\top],\\
    D_{32} &= \frac{2\rho_j \omega_j}{d N^2} \E_{W, Z_i}\operatorname{tr}\left[ F_2 K_2^{-1} Z_2^\top \Sigma_{\rm s} Z_1  K_1^{-1} F_1^\top  \right].
\end{align*}
Now, we use the linear pencil method to compute $D_{31}$ and $D_{32}$.

\subsubsection{Computation of $D_{31}$}
Let
\begingroup
\setlength\arraycolsep{0pt}
\begin{align*}
    Q^{3} = \scalemath{0.95}{\begin{bmatrix}
  I_n & \frac{\sqrt{\rho_{\rm s} \omega_{\rm s}}\Theta_2^\top}{\gamma \sqrt{N}} & \frac{\sqrt{\rho_{\rm s}}Z_2^\top}{\gamma \sqrt{d}} & \cdot &\cdot & \cdot& \cdot &\cdot & \cdot& \cdot &\cdot & \cdot &\cdot & \cdot\\
  -\frac{\sqrt{\rho_{\rm s} \omega_{\rm s}}\Theta_2}{\sqrt{N}} & I_{N} & \cdot & \cdot & -\frac{\sqrt{\rho_{\rm s}}W}{\sqrt{N}} & \cdot &\cdot & \cdot& \cdot &\cdot & \cdot& \cdot &\cdot & \cdot\\
  \cdot & \cdot & I_{d} & -\Sigma_{\rm s}^\frac{1}{2} & \cdot & \cdot &\cdot & \cdot& \cdot &\cdot & \cdot& \cdot &\cdot & \cdot\\
  \cdot & -\frac{W^\top}{\sqrt{N}} & \cdot & I_{d} & \cdot & \cdot &\cdot & \cdot& \cdot &\cdot & \cdot& \frac{\Sigma_{\rm s}^\frac{1}{2}}{\sqrt{\rho_{\rm s}}} &\cdot & \cdot\\
  \cdot & \cdot & \cdot & \cdot & I_{d} & -\Sigma_{\rm s}^\frac{1}{2} &\cdot & \cdot& \cdot &\cdot & \cdot& \cdot &\cdot & \cdot\\
  -\frac{Z_2}{\sqrt{d}} & \cdot & \cdot & \cdot & \cdot & I_{d} &\cdot & \cdot& \cdot &\cdot & \cdot& \cdot &\cdot & \cdot\\
  \cdot & \cdot& \cdot &\cdot & \cdot& \cdot & I_n & \frac{\sqrt{\rho_{\rm s} \omega_{\rm s}}\Theta_1^\top}{\gamma \sqrt{N}}& \frac{\sqrt{\rho_{\rm s}} Z_1^\top}{\gamma \sqrt{d}} &\cdot &\cdot &\cdot &\cdot & \cdot\\ 
  \cdot & \cdot& \cdot &\cdot & \cdot& \cdot & -\frac{\sqrt{\rho_{\rm s} \omega_{\rm s}}\Theta_1}{\sqrt{N}} & I_{N} & \cdot & \cdot & -\frac{\sqrt{\rho_{\rm s}}W}{\sqrt{N}} & \cdot &\cdot & \cdot\\ 
  \cdot & \cdot& \cdot &\cdot & \cdot& \cdot & \cdot & \cdot & I_{d} & -\Sigma_{\rm s}^\frac{1}{2} & \cdot & \cdot &\cdot & \cdot\\ 
  \cdot & \cdot& \cdot &\cdot & \cdot& \cdot & \cdot & -\frac{W^\top}{\sqrt{N}} & \cdot & I_{d} & \cdot & \cdot &\cdot & \cdot\\ 
  \cdot & \cdot& \cdot & \cdot & \cdot& \cdot & \cdot & \cdot & \cdot & \cdot & I_{d} & -\Sigma_{\rm s}^\frac{1}{2} & \frac{\Sigma_j}{\sqrt{\rho_{\rm s}}} & \cdot\\
  \cdot & \cdot& \cdot &\cdot & \cdot& \cdot &-\frac{Z_1}{\sqrt{d}} & \cdot & \cdot & \cdot & \cdot & I_{d} &\cdot & \cdot\\
  \cdot & \cdot & \cdot & \cdot & \cdot & \cdot & \cdot & \cdot & \cdot & \cdot & \cdot & \cdot & I_d & -\frac{W^\top}{\sqrt{N}}\\
  \cdot & \cdot & \cdot & \cdot & \cdot & \cdot & \cdot & \cdot & \cdot & \cdot & \cdot & \cdot & \cdot & I_N
  \end{bmatrix}}
\end{align*}
\endgroup
and $G^{3} = (\operatorname{id} \otimes \mathbb{E} \overline{\operatorname{tr}}) ((Q^{3})^{-1})$.
Then,
\begin{align*}
    G_{2, 14}^{3} = \frac{\psi}{d^2 N^2} \E_{W, Z_i}\operatorname{tr} [W \Sigma_j W^\top F_2 K_2^{-1} Z_2^\top \Sigma_{\rm s} Z_1 K_1^{-1} F_1^\top]  = \frac{\psi}{2\rho_j} D_{31}.
\end{align*}
We augment $Q^{3}$ to the symmetric matrix $\bar{Q}^{3}$ as
\begin{align*}
    \bar{Q}^{3} = \begin{bmatrix} 0 & (Q^{3})^\top \\ Q^{3} & 0 \end{bmatrix}
\end{align*}
and write
\begin{align*}
    \bar{Q}^{3} &= \bar{Z}^{3} - \bar{Q}_{W, Z, \Theta}^{3} - \bar{Q}_\Sigma^{3} \\
    &=\begin{bmatrix} 0 & I_{2n + 9d + 3N} \\ I_{2n + 9d + 3N} & 0 \end{bmatrix} - \begin{bmatrix} 0 & (Q_{W, Z, \Theta}^{3})^\top \\ Q_{W, Z, \Theta}^{3} & 0 \end{bmatrix} - \begin{bmatrix} 0 & (Q_\Sigma^{3})^\top \\ Q_\Sigma^{3} & 0 \end{bmatrix},
\end{align*}
where
\begingroup
\setlength\arraycolsep{2pt}
\begin{align*} Q_{W, Z, \Theta}^{3} = \begin{bmatrix}
 \cdot & -\frac{\sqrt{\rho_{\rm s} \omega_{\rm s}}\Theta_2^\top}{\gamma \sqrt{N}} & -\frac{\sqrt{\rho_{\rm s}}Z_2^\top}{\gamma \sqrt{d}} & \cdot &\cdot & \cdot& \cdot &\cdot & \cdot& \cdot &\cdot & \cdot &\cdot & \cdot\\
  \frac{\sqrt{\rho_{\rm s} \omega_{\rm s}}\Theta_2}{\sqrt{N}} & \cdot & \cdot & \cdot & \frac{\sqrt{\rho_{\rm s}}W}{\sqrt{N}} & \cdot &\cdot & \cdot& \cdot &\cdot & \cdot& \cdot &\cdot & \cdot\\
  \cdot & \cdot & \cdot & \cdot & \cdot & \cdot &\cdot & \cdot& \cdot &\cdot & \cdot& \cdot &\cdot & \cdot\\
  \cdot & \frac{W^\top}{\sqrt{N}} & \cdot & \cdot & \cdot & \cdot &\cdot & \cdot& \cdot &\cdot & \cdot& \cdot &\cdot & \cdot\\
  \cdot & \cdot & \cdot & \cdot & \cdot & \cdot &\cdot & \cdot& \cdot &\cdot & \cdot& \cdot &\cdot & \cdot\\
  \frac{Z_2}{\sqrt{d}} & \cdot & \cdot & \cdot & \cdot & \cdot &\cdot & \cdot& \cdot &\cdot & \cdot& \cdot &\cdot & \cdot\\
  \cdot & \cdot& \cdot &\cdot & \cdot& \cdot & \cdot & -\frac{\sqrt{\rho_{\rm s} \omega_{\rm s}}\Theta_1^\top}{\gamma \sqrt{N}}& -\frac{\sqrt{\rho_{\rm s}} Z_1^\top}{\gamma \sqrt{d}} &\cdot &\cdot &\cdot &\cdot & \cdot\\ 
  \cdot & \cdot& \cdot &\cdot & \cdot& \cdot & \frac{\sqrt{\rho_{\rm s} \omega_{\rm s}}\Theta_1}{\sqrt{N}} & \cdot & \cdot & \cdot & \frac{\sqrt{\rho_{\rm s}}W}{\sqrt{N}} & \cdot &\cdot & \cdot\\ 
  \cdot & \cdot& \cdot &\cdot & \cdot& \cdot & \cdot & \cdot & \cdot & \cdot & \cdot & \cdot &\cdot & \cdot\\ 
  \cdot & \cdot& \cdot &\cdot & \cdot& \cdot & \cdot & \frac{W^\top}{\sqrt{N}} & \cdot & \cdot & \cdot & \cdot &\cdot & \cdot\\ 
  \cdot & \cdot& \cdot & \cdot & \cdot& \cdot & \cdot & \cdot & \cdot & \cdot & \cdot & \cdot & \cdot & \cdot\\
  \cdot & \cdot& \cdot &\cdot & \cdot& \cdot &\frac{Z_1}{\sqrt{d}} & \cdot & \cdot & \cdot & \cdot & \cdot &\cdot & \cdot\\
  \cdot & \cdot & \cdot & \cdot & \cdot & \cdot & \cdot & \cdot & \cdot & \cdot & \cdot & \cdot & \cdot & \frac{W^\top}{\sqrt{N}}\\
  \cdot & \cdot & \cdot & \cdot & \cdot & \cdot & \cdot & \cdot & \cdot & \cdot & \cdot & \cdot & \cdot & \cdot
  \end{bmatrix}
\end{align*}
\endgroup
and
\begin{align*}
    Q_{\Sigma}^{3} = \begin{bmatrix} \cdot & \cdot & \cdot & \cdot & \cdot & \cdot & \cdot & \cdot & \cdot & \cdot & \cdot & \cdot & \cdot & \cdot \\
  \cdot & \cdot & \cdot & \cdot & \cdot & \cdot & \cdot & \cdot & \cdot & \cdot & \cdot & \cdot & \cdot & \cdot \\
  \cdot & \cdot & \cdot & \Sigma_{\rm s}^\frac{1}{2} & \cdot & \cdot & \cdot & \cdot & \cdot & \cdot & \cdot & \cdot & \cdot & \cdot \\
  \cdot & \cdot & \cdot & \cdot & \cdot & \cdot & \cdot & \cdot & \cdot & \cdot & \cdot & -\frac{\Sigma_{\rm s}^\frac{1}{2}}{\sqrt{\rho_{\rm s}}} & \cdot & \cdot \\
  \cdot & \cdot & \cdot & \cdot & \cdot & \Sigma_{\rm s}^\frac{1}{2} & \cdot & \cdot & \cdot & \cdot & \cdot & \cdot & \cdot & \cdot \\
  \cdot & \cdot & \cdot & \cdot & \cdot & \cdot & \cdot & \cdot & \cdot & \cdot & \cdot & \cdot & \cdot & \cdot \\
  \cdot & \cdot & \cdot & \cdot & \cdot & \cdot & \cdot & \cdot & \cdot & \cdot & \cdot & \cdot & \cdot & \cdot\\
  \cdot & \cdot & \cdot & \cdot & \cdot & \cdot & \cdot & \cdot & \cdot & \cdot & \cdot & \cdot & \cdot & \cdot\\
  \cdot & \cdot & \cdot & \cdot & \cdot & \cdot & \cdot & \cdot & \cdot & \Sigma_{\rm s}^\frac{1}{2} & \cdot & \cdot & \cdot & \cdot \\
  \cdot & \cdot & \cdot & \cdot & \cdot & \cdot & \cdot & \cdot & \cdot & \cdot & \cdot & \cdot & \cdot & \cdot \\
  \cdot & \cdot & \cdot & \cdot & \cdot & \cdot & \cdot & \cdot & \cdot & \cdot & \cdot & \Sigma_{\rm s}^\frac{1}{2} & -\frac{\Sigma_j}{\sqrt{\rho_{\rm s}}} & \cdot \\
  \cdot & \cdot & \cdot & \cdot & \cdot & \cdot & \cdot & \cdot & \cdot & \cdot & \cdot & \cdot & \cdot & \cdot \\
  \cdot & \cdot & \cdot & \cdot & \cdot & \cdot & \cdot & \cdot & \cdot & \cdot & \cdot & \cdot & \cdot & \cdot \\
  \cdot & \cdot & \cdot & \cdot & \cdot & \cdot & \cdot & \cdot & \cdot & \cdot & \cdot & \cdot & \cdot & \cdot\end{bmatrix}.
\end{align*}
Defining $\bar{G}^3$ below,
\begin{align*}
    \bar{G}^{3} &= \begin{bmatrix} 0 & G^{3} \\ (G^{3})^\top & 0 \end{bmatrix} = \begin{bmatrix} 0 & (\operatorname{id} \otimes \mathbb{E} \overline{\operatorname{tr}}) ((Q^{3})^{-1}) \\ (\operatorname{id} \otimes \mathbb{E} \overline{\operatorname{tr}}) (((Q^{3})^\top)^{-1}) & 0\end{bmatrix}\\
    &= (\operatorname{id} \otimes \mathbb{E} \overline{\operatorname{tr}}) \begin{bmatrix} 0 & (Q^{3})^{-1} \\ ((Q^{3})^\top)^{-1} & 0\end{bmatrix} = (\operatorname{id} \otimes \mathbb{E} \overline{\operatorname{tr}}) ((\bar{Q}^{3})^{-1}).
\end{align*}
It can be viewed as the operator-valued Cauchy transform of $\bar{Q}_{W, Z, \Theta}^{3} + \bar{Q}_\Sigma^{3}$ (in the space we consider in Remark \ref{rmk:rectangular}), i.e.,
\begin{align*}
    \bar{G}^{3} = (\operatorname{id} \otimes \mathbb{E} \overline{\operatorname{tr}})(\bar Z^{3} - \bar Q_{W, Z, \Theta}^{3} - \bar Q_\Sigma^{3})^{-1} = \mathcal{G}_{\bar{Q}_{W, Z, \Theta}^{3} + \bar{Q}_\Sigma^{3}}(\bar{Z}^{3}).
\end{align*}
Further by the subordination formula \eqref{eqn:subordination},
\begin{align}\label{eqn:d3subordination}
    \bar{G}^{3} &= \mathcal{G}_{\bar{Q}_\Sigma^{3}}(\bar{Z}^{} - \mathcal{R}_{\bar{Q}_{W, Z, \Theta}^{3}}(\bar{G}^{3})) = (\operatorname{id} \otimes \mathbb{E} \overline{\operatorname{tr}})(\bar{Z}^{3} - \mathcal{R}_{\bar{Q}_{W, Z, \Theta}^{3}}(\bar{G}^{3}) - \bar{Q}_\Sigma^{3})^{-1}.
\end{align}
Since $\bar{Q}_{W, Z, \Theta}^{3}$ consists of i.i.d. Gaussian blocks, by \eqref{eqn:gaussianrtransform}, its limiting $R$-transform has the form
\begin{align*}
    \mathcal{R}_{\bar{Q}_{W, Z, \Theta}^{3}}(\bar{G}^{3}) = \begin{bmatrix} 0 & (R^{3})^\top \\ R^{3} & 0 \end{bmatrix},
\end{align*}
where the non-zero blocks of $R^{3}$ are
\begin{align*}
    &R_{1,1}^3 = -\frac{\rho_{\rm s} \omega_{\rm s}}{\gamma} G_{2, 2}^3 -\frac{\sqrt{\rho_{\rm s}}}{\gamma} G_{3, 6}^3, \quad R_{2, 2}^{3} = -\frac{\psi\rho_{\rm s} \omega_{\rm s}}{\gamma \phi} G_{1, 1}^{3} + \sqrt{\rho_{\rm s}} \psi G_{5, 4}^{3}, \quad R_{2, 8}^3 = \sqrt{\rho_{\rm s}} \psi G_{5, 10}^3,\\
    &R_{2, 14}^3 = \sqrt{\rho_{\rm s}} \psi G_{5, 13}^3, \quad R_{4, 5}^3 = \sqrt{\rho_{\rm s}} G_{2, 2}^3, \quad R_{4, 11}^3 = \sqrt{\rho_{\rm s}} G_{2, 8}^3, \quad R_{6, 3}^3 = -\frac{\sqrt{\rho_{\rm s}}}{\gamma \phi} G_{1, 1}^3,\\
    &R_{7, 7}^3 = -\frac{\rho_{\rm s} \omega_{\rm s}}{\gamma} G_{8, 8}^3 -\frac{\sqrt{\rho_{\rm s}}}{\gamma} G_{9, 12}^3,\quad R_{8, 2}^3 = \sqrt{\rho_{\rm s}} \psi G_{11, 4}^3 = 0, \quad R_{8, 8}^{3} = -\frac{\psi\rho_{\rm s} \omega_{\rm s}}{\gamma \phi} G_{7, 7}^{3} + \sqrt{\rho_{\rm s}} \psi G_{11, 10}^{3}, \\
    &R_{8, 14}^3 = \sqrt{\rho_{\rm s}} \psi G_{11, 13}^3, \quad R_{10, 5}^3 = \sqrt{\rho_{\rm s}} G_{8, 2}^3 = 0,\quad R_{10, 11}^3 = \sqrt{\rho_{\rm s}} G_{8, 8}^3, \\
    &R_{12, 9}^3 = -\frac{\sqrt{\rho_{\rm s}}}{\gamma \phi} G_{7, 7}^3, \quad R_{13, 5}^3 = \sqrt{\rho_{\rm s}} G_{14, 2}^3 = 0, \quad R_{13, 11}^3 = \sqrt{\rho_{\rm s}} G_{14, 8}^3 = 0.
\end{align*}
We used the fact that $G_{11, 4}^3 = G_{8, 2}^3 = G_{14, 2}^3 = G_{14, 8}^3 = 0$, which we obtain from block matrix inversion of $Q^3$.

Further from block matrix inversion of $Q^{3}$ and equations \eqref{eqn:g0blockinversion}, \eqref{eqn:g0values}, we have
\begin{align*}
    &G_{1, 1}^3 = G_{7, 7}^3 = \gamma \E \overline{\operatorname{tr}}(K^{-1}) = G_{1, 1}^0 = \gamma \tau, \quad G_{2, 2}^3 = G_{8, 8}^3 = \gamma \E \overline{\operatorname{tr}}(\hat{K}^{-1}) = G_{2, 2}^0 = \gamma \bar \tau,\\
    &G_{3, 6}^3 = G_{9, 12}^3 = \frac{\gamma \sqrt{\rho_{\rm s}}\,\E \overline{\operatorname{tr}} [\Sigma_{\rm s} W^\top \hat{K}^{-1} W]}{N} = G_{3, 6}^0 =  \gamma \sqrt{\rho_{\rm s}}  \bar\tau \mathcal{I}_{1, 1}^{\rm s},\\
    &G_{5, 4}^3 = G_{11, 10}^3 = -\frac{\sqrt{\rho_{\rm s}}\, \E \overline{\operatorname{tr}} [\Sigma_{\rm s} Z K^{-1} Z^\top]}{d}= G_{5, 4}^0 =  -\frac{\sqrt{\rho_{\rm s}}\tau \mathcal{I}_{1, 1}^{\rm s}}{\phi}.
\end{align*}
Plugging these into \eqref{eqn:d3subordination}, we have the following self-consistent equations
\begin{align*}
&G_{2, 14}^3 = \gamma^2 \rho_{\rm s} \bar\tau^2 \psi^2 G_{5, 10}^3 G_{11, 13}^3 + \gamma \sqrt{\rho_{\rm s}} \bar\tau \psi G_{5, 13}^3, \quad G_{5, 10}^3 = -\frac{\sqrt{\rho_{\rm s}} \tau^2}{\phi}\cI_{2, 2}^{\rm s} + \frac{\rho_{\rm s}^\frac{3}{2} \tau^2}{\phi} \cI_{2, 2}^{\rm s} G_{2, 8}^3,\\
&G_{2, 8}^3 = \gamma^2 \sqrt{\rho_{\rm s}} \bar\tau^2 \psi G_{5, 10}^3, \quad G_{5, 13}^3 = \sqrt{\rho_{\rm s}} \tau \cI_{2, 2}^j G_{2, 8}^3 + \frac{\gamma \sqrt{\rho_{\rm s}} \tau^2 \bar\tau}{\phi} \cI_{3, 2}^j, \quad G_{11, 13}^3 = -\frac{\cI_{1, 1}^j}{\sqrt{\rho_{\rm s}}}.
\end{align*}
Solving for $G_{5, 10}^3$ gives
\begin{align*}
G_{5, 10}^3 = -\frac{\sqrt{\rho_{\rm s}} \tau^2 \cI_{2, 2}^{\rm s}}{\phi - \psi \kappa^2 \cI_{2, 2}^{\rm s}}.
\end{align*}
Plugging in $G_{5, 10}^3, G_{11, 13}^3, G_{5, 13}^3$ to find $G_{2, 14}^3$, we get
\begin{align}\label{eqn:d31}
    D_{31} = \frac{2 \rho_j}{\psi} G_{2, 14}^3 = \frac{2\rho_j \psi \phi \kappa^2 \cI_{2, 2}^{\rm s} \cI_{1, 2}^j}{\rho_{\rm s}(\phi - \psi \kappa^2 \cI_{2, 2}^{\rm s})} + \frac{2\rho_j \kappa^2}{\rho_{\rm s} \phi} \cI_{3, 2}^j.
\end{align}

\subsubsection{Computation of $D_{32}$}
Let
\begingroup
\setlength\arraycolsep{0pt}
\begin{align*}
    Q^{4} = \begin{bmatrix}
  I_n & \frac{\sqrt{\rho_{\rm s} \omega_{\rm s}}\Theta_2^\top}{\gamma \sqrt{N}} & \frac{\sqrt{\rho_{\rm s}}Z_2^\top}{\gamma \sqrt{d}} & \cdot &\cdot & \cdot& \cdot &\cdot & \cdot& \cdot &\cdot & \cdot\\
  -\frac{\sqrt{\rho_{\rm s} \omega_{\rm s}}\Theta_2}{\sqrt{N}} & I_{N} & \cdot & \cdot & -\frac{\sqrt{\rho_{\rm s}}W}{\sqrt{N}} & \cdot &\cdot & \cdot& \cdot &\cdot & \cdot& \cdot\\
  \cdot & \cdot & I_{d} & -\Sigma_{\rm s}^\frac{1}{2} & \cdot & \cdot &\cdot & \cdot& \cdot &\cdot & \cdot& \cdot \\
  \cdot & -\frac{W^\top}{\sqrt{N}} & \cdot & I_{d} & \cdot & \cdot &\cdot & \cdot& \cdot &\cdot & I_d & \cdot \\
  \cdot & \cdot & \cdot & \cdot & I_{d} & -\Sigma_{\rm s}^\frac{1}{2} &\cdot & \cdot& \cdot &\cdot & \cdot& \cdot \\
  -\frac{Z_2}{\sqrt{d}} & \cdot & \cdot & \cdot & \cdot & I_{d} &\cdot & \cdot& \cdot &\cdot & \cdot& \cdot\\
  \cdot & \cdot& \cdot &\cdot & \cdot& \cdot & I_n & \frac{\sqrt{\rho_{\rm s} \omega_{\rm s}}\Theta_1^\top}{\gamma \sqrt{N}}& \frac{\sqrt{\rho_{\rm s}} Z_1^\top}{\gamma \sqrt{d}} &\cdot &\cdot &\cdot\\ 
  \cdot & \cdot& \cdot &\cdot & \cdot& \cdot & -\frac{\sqrt{\rho_{\rm s} \omega_{\rm s}}\Theta_1}{\sqrt{N}} & I_{N} & \cdot & \cdot & -\frac{\sqrt{\rho_{\rm s}}W}{\sqrt{N}} & \cdot\\ 
  \cdot & \cdot& \cdot &\cdot & \cdot& \cdot & \cdot & \cdot & I_{d} & -\Sigma_{\rm s}^\frac{1}{2} & \cdot & \cdot\\ 
  \cdot & \cdot& \cdot &\cdot & \cdot& \cdot & \cdot & -\frac{W^\top}{\sqrt{N}} & \cdot & I_{d} & \cdot & \cdot\\ 
  \cdot & \cdot& \cdot &\cdot & \cdot& \cdot & \cdot & \cdot & \cdot & \cdot & I_{d} & -\Sigma_{\rm s}^\frac{1}{2}\\
  \cdot & \cdot& \cdot &\cdot & \cdot& \cdot &-\frac{Z_1}{\sqrt{d}} & \cdot & \cdot & \cdot & \cdot & I_{d}
  \end{bmatrix}
\end{align*}
\endgroup
and $G^{4} = (\operatorname{id} \otimes \mathbb{E} \overline{\operatorname{tr}}) ((Q^{4})^{-1})$.
Then,
\begin{align*}
    G_{2, 8}^{4} = -\frac{\sqrt{\rho_{\rm s}}}{dN^2} \E_{W, Z_i}\operatorname{tr}\left[ F_2 K_2^{-1} Z_2^\top \Sigma_{\rm s} Z_1  K_1^{-1} F_1^\top  \right] = -\frac{\sqrt{\rho_{\rm s}}}{2\rho_j \omega_j} D_{32}.
\end{align*}
We augment $Q^{4}$ to the symmetric matrix $\bar{Q}^{4}$ as
\begin{align*}
    \bar{Q}^{4} = \begin{bmatrix} 0 & (Q^{4})^\top \\ Q^{4} & 0 \end{bmatrix}
\end{align*}
and write
\begin{align*}
    \bar{Q}^{4} &= \bar{Z}^{4} - \bar{Q}_{W, Z, \Theta}^{4} - \bar{Q}_\Sigma^{4} \\
    &=\begin{bmatrix} 0 & I_{2n + 8d + 2N} \\ I_{2n + 8d + 2N} & 0 \end{bmatrix} - \begin{bmatrix} 0 & (Q_{W, Z, \Theta}^{4})^\top \\ Q_{W, Z, \Theta}^{4} & 0 \end{bmatrix} - \begin{bmatrix} 0 & (Q_\Sigma^{4})^\top \\ Q_\Sigma^{4} & 0 \end{bmatrix},
\end{align*}
where
\begingroup
\setlength\arraycolsep{2pt}
\begin{align*} Q_{W, Z, \Theta}^{4} = \begin{bmatrix}
  \cdot & -\frac{\sqrt{\rho_{\rm s} \omega_{\rm s}}\Theta_2^\top}{\gamma \sqrt{N}} & -\frac{\sqrt{\rho_{\rm s}}Z_2^\top}{\gamma \sqrt{d}} & \cdot &\cdot & \cdot& \cdot &\cdot & \cdot& \cdot &\cdot & \cdot\\
  \frac{\sqrt{\rho_{\rm s} \omega_{\rm s}}\Theta_2}{\sqrt{N}} & \cdot & \cdot & \cdot & \frac{\sqrt{\rho_{\rm s}}W}{\sqrt{N}} & \cdot &\cdot & \cdot& \cdot &\cdot & \cdot& \cdot\\
  \cdot & \cdot & \cdot & \cdot & \cdot & \cdot &\cdot & \cdot& \cdot &\cdot & \cdot& \cdot \\
  \cdot & \frac{W^\top}{\sqrt{N}} & \cdot & \cdot & \cdot & \cdot &\cdot & \cdot& \cdot &\cdot & \cdot & \cdot \\
  \cdot & \cdot & \cdot & \cdot & \cdot & \cdot &\cdot & \cdot& \cdot &\cdot & \cdot& \cdot \\
  \frac{Z_2}{\sqrt{d}} & \cdot & \cdot & \cdot & \cdot & \cdot &\cdot & \cdot& \cdot &\cdot & \cdot& \cdot\\
  \cdot & \cdot& \cdot &\cdot & \cdot& \cdot & \cdot & -\frac{\sqrt{\rho_{\rm s} \omega_{\rm s}}\Theta_1^\top}{\gamma \sqrt{N}}& -\frac{\sqrt{\rho_{\rm s}} Z_1^\top}{\gamma \sqrt{d}} &\cdot &\cdot &\cdot\\ 
  \cdot & \cdot& \cdot &\cdot & \cdot& \cdot & \frac{\sqrt{\rho_{\rm s} \omega_{\rm s}}\Theta_1}{\sqrt{N}} & \cdot & \cdot & \cdot & \frac{\sqrt{\rho_{\rm s}}W}{\sqrt{N}} & \cdot\\ 
  \cdot & \cdot& \cdot &\cdot & \cdot& \cdot & \cdot & \cdot & \cdot & \cdot & \cdot & \cdot\\ 
  \cdot & \cdot& \cdot &\cdot & \cdot& \cdot & \cdot & \frac{W^\top}{\sqrt{N}} & \cdot & \cdot & \cdot & \cdot\\ 
  \cdot & \cdot& \cdot &\cdot & \cdot& \cdot & \cdot & \cdot & \cdot & \cdot & \cdot & \cdot\\
  \cdot & \cdot& \cdot &\cdot & \cdot& \cdot &\frac{Z_1}{\sqrt{d}} & \cdot & \cdot & \cdot & \cdot & \cdot
  \end{bmatrix}
\end{align*}
\endgroup
and
\begin{align*}
    Q_{\Sigma}^{4} = \begin{bmatrix}
  \cdot & \cdot & \cdot & \cdot &\cdot & \cdot& \cdot &\cdot & \cdot& \cdot &\cdot & \cdot\\
  \cdot & \cdot & \cdot & \cdot & \cdot & \cdot &\cdot & \cdot& \cdot &\cdot & \cdot& \cdot\\
  \cdot & \cdot & \cdot & \Sigma_{\rm s}^\frac{1}{2} & \cdot & \cdot &\cdot & \cdot& \cdot &\cdot & \cdot& \cdot \\
  \cdot & \cdot & \cdot & \cdot & \cdot & \cdot &\cdot & \cdot& \cdot &\cdot & I_d & \cdot \\
  \cdot & \cdot & \cdot & \cdot & \cdot & \Sigma_{\rm s}^\frac{1}{2} &\cdot & \cdot& \cdot &\cdot & \cdot& \cdot \\
  \cdot & \cdot & \cdot & \cdot & \cdot & \cdot &\cdot & \cdot& \cdot &\cdot & \cdot& \cdot\\
  \cdot & \cdot& \cdot &\cdot & \cdot& \cdot & \cdot & \cdot & \cdot &\cdot &\cdot &\cdot\\ 
  \cdot & \cdot& \cdot &\cdot & \cdot& \cdot & \cdot & \cdot & \cdot & \cdot & \cdot & \cdot\\ 
  \cdot & \cdot& \cdot &\cdot & \cdot& \cdot & \cdot & \cdot & \cdot & \Sigma_{\rm s}^\frac{1}{2} & \cdot & \cdot\\ 
  \cdot & \cdot& \cdot &\cdot & \cdot& \cdot & \cdot & \cdot & \cdot & \cdot & \cdot & \cdot\\ 
  \cdot & \cdot& \cdot &\cdot & \cdot& \cdot & \cdot & \cdot & \cdot & \cdot &\cdot & \Sigma_{\rm s}^\frac{1}{2}\\
  \cdot & \cdot& \cdot &\cdot & \cdot& \cdot &\cdot & \cdot & \cdot & \cdot & \cdot & \cdot
  \end{bmatrix}.
\end{align*}
Defining $\bar{G}^4$ below,
\begin{align*}
    \bar{G}^{4} &= \begin{bmatrix} 0 & G^{4} \\ (G^{4})^\top & 0 \end{bmatrix} = \begin{bmatrix} 0 & (\operatorname{id} \otimes \mathbb{E} \overline{\operatorname{tr}}) ((Q^{4})^{-1}) \\ (\operatorname{id} \otimes \mathbb{E} \overline{\operatorname{tr}}) (((Q^{4})^\top)^{-1}) & 0\end{bmatrix}\\
    &= (\operatorname{id} \otimes \mathbb{E} \overline{\operatorname{tr}}) \begin{bmatrix} 0 & (Q^{4})^{-1} \\ ((Q^{4})^\top)^{-1} & 0\end{bmatrix} = (\operatorname{id} \otimes \mathbb{E} \overline{\operatorname{tr}}) ((\bar{Q}^{4})^{-1}).
\end{align*}
It can be viewed as the operator-valued Cauchy transform of $\bar{Q}_{W, Z, \Theta}^{4} + \bar{Q}_\Sigma^{4}$ (in the space we consider in Remark \ref{rmk:rectangular}), i.e.,
\begin{align*}
    \bar{G}^{4} = (\operatorname{id} \otimes \mathbb{E} \overline{\operatorname{tr}})(\bar Z^{4} - \bar Q_{W, Z, \Theta}^{4} - \bar Q_\Sigma^{4})^{-1} = \mathcal{G}_{\bar{Q}_{W, Z, \Theta}^{4} + \bar{Q}_\Sigma^{4}}(\bar{Z}^{4}).
\end{align*}
Further by the subordination formula \eqref{eqn:subordination},
\begin{align}\label{eqn:d32subordination}
    \bar{G}^{4} &= \mathcal{G}_{\bar{Q}_\Sigma^{4}}(\bar{Z}^{} - \mathcal{R}_{\bar{Q}_{W, Z, \Theta}^{4}}(\bar{G}^{4})) = (\operatorname{id} \otimes \mathbb{E} \overline{\operatorname{tr}})(\bar{Z}^{4} - \mathcal{R}_{\bar{Q}_{W, Z, \Theta}^{4}}(\bar{G}^{4}) - \bar{Q}_\Sigma^{4})^{-1}.
\end{align}
Since $\bar{Q}_{W, Z, \Theta}^{4}$ consists of i.i.d. Gaussian blocks, by \eqref{eqn:gaussianrtransform}, its limiting $R$-transform has a form
\begin{align*}
    \mathcal{R}_{\bar{Q}_{W, Z, \Theta}^{4}}(\bar{G}^{4}) = \begin{bmatrix} 0 & (R^{4})^\top \\ R^{4} & 0 \end{bmatrix},
\end{align*}
where the non-zero blocks of $R^{4}$ are
\begin{align*}
&R_{1, 1}^4 = -\frac{\rho_{\rm s} \omega_{\rm s}}{\gamma} G_{2, 2}^4 - \frac{\sqrt{\rho_{\rm s}}}{\gamma} G_{3, 6}^4, \quad R_{2, 2}^4 = -\frac{\rho_{\rm s} \omega_{\rm s} \psi}{\gamma \phi} G_{1, 1}^4 + \sqrt{\rho_{\rm s}} \psi G_{5, 4}^4, \quad R_{2, 8}^4 = \sqrt{\rho_{\rm s}} \psi G_{5, 10}^4,\\
&R_{4, 5}^4 = \sqrt{\rho_{\rm s}} G_{2, 2}^4, \quad R_{4, 11}^4 = \sqrt{\rho_{\rm s}} G_{2, 8}^4, \quad R_{6, 3}^4 = -\frac{\sqrt{\rho_{\rm s}}}{\gamma \phi} G_{1, 1}^4, \quad R_{7, 7}^4 = -\frac{\rho_{\rm s} \omega_{\rm s}}{\gamma} G_{8, 8}^4 - \frac{\sqrt{\rho_{\rm s}}}{\gamma} G_{9, 12}^4,\\
&R_{8, 2}^4 = \sqrt{\rho_{\rm s}} \psi G_{11, 4}^4 = 0, \quad R_{8, 8}^4 = -\frac{\rho_{\rm s} \omega_{\rm s} \psi}{\gamma \phi} G_{7, 7}^4 + \sqrt{\rho_{\rm s}} \psi G_{11, 10}^4, \quad R_{10, 5}^4 = \sqrt{\rho_{\rm s}} G_{8, 2}^4 = 0,\\
&R_{10, 11}^4 = \sqrt{\rho_{\rm s}} G_{8, 8}^4, \quad R_{12, 9}^4 = -\frac{\sqrt{\rho_{\rm s}}}{\gamma \phi} G_{7, 7}^4.
\end{align*}
We used the fact that $G_{11, 4}^4 = G_{8, 2}^4 = 0$, which we obtain from block matrix inversion of $Q^4$.

Further from block matrix inversion of $Q^{4}$ and equations \eqref{eqn:g0blockinversion}, \eqref{eqn:g0values}, we have
\begin{align*}
    &G_{1, 1}^4 = G_{7, 7}^4 = \gamma \E \overline{\operatorname{tr}}(K^{-1}) = G_{1, 1}^0 = \gamma \tau, \quad G_{2, 2}^4 = G_{8, 8}^4 = \gamma \E \overline{\operatorname{tr}}(\hat{K}^{-1}) = G_{2, 2}^0 = \gamma \bar \tau,\\
    &G_{3, 6}^4 = G_{9, 12}^4 = \frac{\gamma \sqrt{\rho_{\rm s}}\,\E \overline{\operatorname{tr}} [\Sigma_{\rm s} W^\top \hat{K}^{-1} W]}{N} = G_{3, 6}^0 =  \gamma \sqrt{\rho_{\rm s}}  \bar\tau \mathcal{I}_{1, 1}^{\rm s},\\
    &G_{5, 4}^4 = G_{11, 10}^4 = -\frac{\sqrt{\rho_{\rm s}}\, \E \overline{\operatorname{tr}} [\Sigma_{\rm s} Z K^{-1} Z^\top]}{d}= G_{5, 4}^0 =  -\frac{\sqrt{\rho_{\rm s}}\tau \mathcal{I}_{1, 1}^{\rm s}}{\phi}.
\end{align*}
Plugging these into \eqref{eqn:d32subordination}, we have the following self-consistent equations
\begin{align*}
G_{2, 8}^4 = \frac{\sqrt{\rho_{\rm s}} \psi \phi^2 G_{5, 10}^4}{(\phi + \rho_{\rm s} \tau \psi (\omega_{\rm s} + \cI_{1, 1}^{\rm s}))^2}, \quad G_{5, 10}^4 = -\frac{\rho_{\rm s} \tau^2}{\phi} \cI_{2, 2}^{\rm s} + \frac{\rho_{\rm s}^\frac{3}{2} \tau^2}{\phi} \cI_{2, 2}^{\rm s} G_{2, 8}^4.
\end{align*}
Solving for $G_{2, 8}^4$ and plugging in to $D_{32}$, we get
\begin{align}\label{eqn:d32}
    D_{32} =  -\frac{2\rho_j \omega_j}{\sqrt{\rho_{\rm s}}} G_{2, 8}^4 = \frac{2\rho_j \omega_j \psi \kappa^2 \cI_{2, 2}^{\rm s}}{\rho_{\rm s}(\phi - \psi \kappa^2 \cI_{2, 2}^{\rm s})}.
\end{align}

\subsubsection{Computation of $\stsd_{\textnormal{SW}}^j (\phi, \psi, \gamma)$}
Combining equations \eqref{eqn:newdecompose}, \eqref{eqn:d1}, \eqref{eqn:d3}, \eqref{eqn:d31}, \eqref{eqn:d32}, we get
\begin{align*}
    \stsd_\textnormal{SW}^j (\phi, \psi, \gamma) = \stsd_\textnormal{I}^j (\phi, \psi, \gamma) -  \frac{2\rho_j \psi \kappa^2(\omega_j + \phi \cI^{j}_{1, 2}) \cI^{\rm s}_{2, 2}}{\rho_{\rm s}(\phi - \psi \kappa^2 \cI_{2, 2}^{\rm s})}.
\end{align*}

\subsection{Proof of Corollary \ref{cor:ridgeless}}\label{sec:cor32proof}
Since $\kappa \leq 1 / \omega_{\rm s}$ for any $\gamma > 0$ by \eqref{eqn:defkappa}, we know $\lim_{\gamma \to 0} \gamma \kappa = 0$.
Thus from \eqref{eqn:deftau}, we have
\begin{align*}
\lim_{\gamma \to 0} \gamma \tau = \frac{|\psi - \phi| + \psi - \phi}{2 \psi}, \quad \lim_{\gamma \to 0} \gamma \bar\tau = 1 - \frac{\psi}{\phi} + \frac{\psi}{\phi} \lim_{\gamma \to 0} \gamma \tau = \frac{|\psi - \phi| + \phi - \psi}{2 \phi}.
\end{align*}

By Condition \ref{cond:jointspectral} and the dominated convergence theorem, the functionals $\cI_{a, b}^{\rm s}, \cI_{a, b}^{\rm t}$ and their derivatives with respect to $\kappa$ are continuous in $\kappa$.
Applying the implicit function theorem to the self-consistent equation \eqref{eqn:defkappa}, viewing it as a function of $\kappa$ and $\gamma$, we find that $\kappa$ is differentiable with respect to $\gamma$ and thus continuous.
Therefore, the limit of $\kappa$, $\cI_{a, b}^{\rm s}$, $\cI_{a, b}^{\rm t}$ when $\gamma \to 0$ is well defined.
Plugging these limits into Theorem \ref{thm:asymp}, we reach

\begin{align*}
        \lim_{\gamma \to 0} \stsd_\textnormal{I}^j (\phi, \psi, \gamma) &= \frac{2 \rho_j \psi \kappa}{\rho_{\rm s}|\phi - \psi|} (\sigma_\ep^2 + \mathcal{I}_{1, 1}^{\rm s})(\omega_{j} + \mathcal{I}_{1, 1}^{j}) +  \begin{cases} \frac{2\rho_j\kappa(\sigma_\ep^2 + \phi \cI^{\rm s}_{1, 2}) \cI^j_{2, 2}}{\rho_{\rm s}(\omega_{\rm s} + \phi \cI^{\rm s}_{1, 2})} & \phi > \psi, \\ \frac{2\rho_j \kappa(\omega_j + \phi \cI_{1, 2}^j) \cI_{2, 2}^{\rm s}}{\rho_{\rm s}(\omega_{\rm s} + \phi \cI_{1, 2}^{\rm s})}  & \phi < \psi, \end{cases}
\end{align*}
\begin{align}\label{eqn:disshareproof}
        \lim_{\gamma \to 0} \stsd&_\textnormal{SS}^j (\phi, \psi, \gamma) = \lim_{\gamma \to 0} \stsd_\textnormal{I}^j (\phi, \psi, \gamma) - \frac{2\rho_j\kappa^2(\sigma_\ep^2 + \phi \cI^{\rm s}_{1, 2}) \cI^j_{2, 2}}{\rho_{\rm s}(1 - \kappa^2 \cI_{2, 2}^{\rm s})},
\end{align}
and
\begin{align}\label{eqn:disSWproof}
        \lim_{\gamma \to 0} \stsd&_\textnormal{SW}^j (\phi, \psi, \gamma) = \lim_{\gamma \to 0} \stsd_\textnormal{I}^j (\phi, \psi, \gamma) - \frac{2\rho_j \psi \kappa^2(\omega_j + \phi \cI^j_{1, 2}) \cI^{\rm s}_{2, 2}}{\rho_{\rm s}(\phi - \psi \kappa^2 \cI_{2, 2}^{\rm s})}.
\end{align}
    
From equation \eqref{eqn:deffunctionals}, we have $\cI_{1, 1}^{\rm s} = \phi \cI_{1, 2}^{\rm s} + \kappa \cI_{2, 2}^{\rm s}$.
Also by \eqref{eqn:defkappa} and \eqref{eqn:deftau}, $\omega_{\rm s} = \frac{1 - \gamma \tau}{\kappa} - \cI_{1, 1}^{\rm s}$.
Therefore,
\begin{align}\label{eqn:omegaidentity}
\omega_{\rm s} + \phi \cI_{1, 2}^{\rm s} = \frac{1 - \gamma \tau}{\kappa} - \cI_{1, 1}^{\rm s} + \phi \cI_{1, 2}^{\rm s} = \frac{1 - \gamma \tau}{\kappa} - \kappa \cI_{2, 2}^{\rm s}.
\end{align}
In the ridgeless limit $\gamma \to 0$, the equation \eqref{eqn:omegaidentity} gives
\begin{align}\label{eqn:identityproof}
    \lim_{\gamma \to 0} \frac{1}{\omega_{\rm s} + \phi \cI_{1, 2}^{\rm s}} = \begin{cases} \lim_{\gamma \to 0} \frac{\kappa}{1 - \kappa^2 \cI_{2, 2}^{\rm s}} & \phi > \psi, \\ \lim_{\gamma \to 0} \frac{\psi \kappa}{\phi - \psi \kappa^2 \cI_{2, 2}^{\rm s}} & \phi < \psi. \end{cases} 
\end{align}
Putting \eqref{eqn:disshareproof}, \eqref{eqn:disSWproof}, \eqref{eqn:identityproof} together, we conclude
\begin{align*}
        \lim_{\gamma \to 0} \stsd_\textnormal{SS}^j (\phi, \psi, \gamma) 
        = \frac{2 \rho_j \psi \kappa}{\rho_{\rm s}|\phi - \psi|} &(\sigma_\ep^2 + \mathcal{I}_{1, 1}^{\rm s})(\omega_{j} + \mathcal{I}_{1, 1}^{j})\\
        &+\begin{cases} 0 & \phi > \psi, \\ \frac{2\rho_j \kappa}{\rho_{\rm s}} \left(\frac{(\omega_j + \phi \cI_{1, 2}^j) \cI_{2, 2}^{\rm s}}{\omega_{\rm s} + \phi \cI_{1, 2}^{\rm s}} - \frac{\kappa(\sigma_\ep^2 + \phi \cI^{\rm s}_{1, 2}) \cI^j_{2, 2}}{1 - \kappa^2 \cI_{2, 2}^{\rm s}} \right)& \phi < \psi, \end{cases}\\
        \lim_{\gamma \to 0} \stsd_\textnormal{SW}^j (\phi, \psi, \gamma) = \frac{2 \rho_j \psi \kappa}{\rho_{\rm s}|\phi - \psi|} &(\sigma_\ep^2 + \mathcal{I}_{1, 1}^{\rm s})(\omega_{j} + \mathcal{I}_{1, 1}^{j})\\
        &+  \begin{cases} \frac{2\rho_j \kappa}{\rho_{\rm s}} \left( \frac{(\sigma_\ep^2 + \phi \cI_{1, 2}^{\rm s}) \cI_{2, 2}^j}{\omega_{\rm s} + \phi \cI_{1, 2}^{\rm s}} - \frac{\psi \kappa (\omega_j + \phi \cI_{1, 2}^j) \cI_{2, 2}^{\rm s}}{\phi - \psi \kappa^2 \cI_{2, 2}^{\rm s}} \right) & \phi > \psi, \\ 0 & \phi < \psi. \end{cases}
    \end{align*}

\subsection{Proof of Theorem \ref{thm:linear relation}}
By Corollary \ref{cor:ridgeless}, disagreement in the ridgeless and overparameterized regime is given by
\begin{align*}
    &\lim_{\gamma \to 0} \stsd_\textnormal{I}^j (\phi, \psi, \gamma) = \frac{2 \rho_j \psi \kappa}{\rho_{\rm s}|\phi - \psi|} (\sigma_\ep^2 + \mathcal{I}_{1, 1}^{\rm s})(\omega_{j} + \mathcal{I}_{1, 1}^{j}) + \frac{2\rho_j\kappa(\sigma_\ep^2 + \phi \cI^{\rm s}_{1, 2}) \cI^j_{2, 2}}{\rho_{\rm s}(\omega_{\rm s} + \phi \cI^{\rm s}_{1, 2})},\\
    &\lim_{\gamma \to 0} \stsd_\textnormal{SS}^j (\phi, \psi, \gamma) = \frac{2 \rho_j \psi \kappa}{\rho_{\rm s}|\phi - \psi|} (\sigma_\ep^2 + \mathcal{I}_{1, 1}^{\rm s})(\omega_{j} + \mathcal{I}_{1, 1}^{j}).
\end{align*}
The self-consistent equation \eqref{eqn:defkapparidgeless} in the overpametrized regime $\phi > \psi$ is
\begin{align*}
\kappa = \frac{1}{\omega_{\rm s} + \cI_{1, 1}^{\rm s}(\kappa)},
\end{align*}
which is independent of $\psi$.
Consequently, the unique positive solution $\kappa$ is also independent of $\psi$.
This proves that the slope $a$ and the intercept $b_{\textnormal{I}}$ defined in Theorem \ref{thm:linear relation} are independent of $\psi$ as well.
One can checking \eqref{eqn:disontheline} via \eqref{eqn:identityproof} and simple algebra.

\subsection{Proof of Theorem \ref{thm:approxlinear1}}
Let $a(\gamma), b_\textnormal{I}(\gamma), b_\textnormal{SS}(\gamma)$ be defined by \eqref{eqn:slopeintercept}, but with $\kappa$ in the self-consistent equation \eqref{eqn:defkappa} with general $\gamma$, instead of the self-consistent equation \eqref{eqn:defkapparidgeless} in the ridgeless limit.
With this notation, we have $a = a(0), b_\textnormal{I} = b_\textnormal{I}(0), b_\textnormal{SS} = b_\textnormal{SS}(0)$.
By Theorem \ref{thm:asymp} and the triangle inequality, deviation from the line is bounded by
\begin{align}\label{eqn:totalbound}
|\stsd_{i}^{\rm t}(\phi, &\psi, \gamma) - a \stsd_{i}^{\rm s}(\phi, \psi, \gamma) - b_i| \nonumber \\
&\leq |\stsd_{i}^{\rm t}(\phi, \psi, \gamma) - a(\gamma) \stsd_{i}^{\rm s}(\phi, \psi, \gamma) - b_i(\gamma)|\\ &\hspace{3cm}+ |a(\gamma) - a(0)| |\stsd_{i}^{\rm s}(\phi, \psi, \gamma) | + |b_i(\gamma) - b_i(0)| \nonumber \\
&\leq A_1 + A_2 + \stsd_{i}^{\rm s}(\phi, \psi, \gamma) |a(\gamma) - a(0)| + |b_i(\gamma) - b_i(0)|, \quad i \in \{\textnormal{I}, \textnormal{SS}\},
\end{align}
where
\begin{align*}
A_1 &= \frac{2 \psi  \gamma \tau \kappa \cI_{2, 2}^{\rm s} |\rho_{\rm t}(\omega_{\rm t} + \phi \cI_{1, 2}^{\rm t}) - a \rho_{\rm s}(\omega_{\rm s} + \phi \cI_{1, 2}^{\rm s})|}{\phi \gamma + \rho_{\rm s}(\psi \gamma \tau + \phi \gamma \bar \tau)(\omega_{\rm s} + \phi \cI_{1, 2}^{\rm s})},\\
A_2 &= 2 (\sigma_\ep^2 + \phi \cI_{1, 2}^{\rm s}) |\rho_{\rm t} \cI_{2, 2}^{\rm t} - a \rho_{\rm s} \cI_{2, 2}^{\rm s}| \left\vert \frac{\kappa \phi \gamma \bar \tau}{\phi \gamma + \rho_{\rm s}(\psi \gamma \tau + \phi \gamma \bar \tau)(\omega_{\rm s} + \phi \cI_{1, 2}^{\rm s})} - \frac{\kappa^2}{\rho_{\rm s}(1 - \kappa
^2 \cI_{2, 2}^{\rm s})}\right\vert.
\end{align*}
In what follows, we bound each of these terms.
We will use $O(\cdot)$ notation to hide constants depending on $\phi, \mu, \sigma_\ep^2, \sigma$.
For example, we can write $\cI_{a, b}^{j} = O(1)$ for $j \in \{\rm s, \rm t\}$ since we assume in Condition \ref{cond:jointspectral} that $\mu$ is compactly supported.
\subsubsection{Bounding $A_1$}
We know $a \leq \rho_{\rm t}(\omega_{\rm t} + \cI_{1, 1}^{\rm t}) / \rho_{\rm s} \omega_{\rm s}$ by \eqref{eqn:slopeintercept}.
Thus,
\begin{align}\label{eqn:A1-1}
\cI_{2, 2}^{\rm s} |\rho_{\rm t}(\omega_{\rm t} + \phi \cI_{1, 2}^{\rm t}) - a \rho_{\rm s}(\omega_{\rm s} + \phi \cI_{1, 2}^{\rm s})| = O(1).
\end{align}
By \eqref{eqn:deftau} and since $\sqrt{x^2 + y^2} \leq |x| + |y|$ for any $x, y \in \mathbb{R}$,
\begin{align}\label{eqn:A1-2}
2 \psi \gamma \tau = \sqrt{(\psi - \phi)^2 + 4\kappa \psi \phi \gamma / \rho_{\rm s}} + \psi - \phi \leq \sqrt{\frac{4 \kappa \psi \phi \gamma}{\rho_{\rm s}}} = O(\sqrt{\psi \gamma}).
\end{align}
Again by \eqref{eqn:deftau}, $\psi \gamma \tau + \phi \gamma \bar \tau = \sqrt{(\psi - \phi)^2 + 4 \kappa \psi \phi \gamma / \rho_{\rm s}}$.
Therefore,
\begin{align}\label{eqn:A1-3}
\frac{\kappa}{\phi\gamma + \rho_{\rm s} (\psi \gamma \tau + \phi \gamma \bar \tau)(\omega_{\rm s} + \phi \cI_{1, 2}^{\rm s})} \leq \frac{\kappa}{\rho_{\rm s} (\psi \gamma \tau + \phi \gamma \bar \tau)(\omega_{\rm s} + \phi \cI_{1, 2}^{\rm s})} = O\left( \frac{1}{1 - \psi / \phi + \sqrt{\psi \gamma} }\right).
\end{align}
Here, we used $\kappa \leq \frac{1}{\omega_{\rm s}} = O(1)$ by \eqref{eqn:defkappa}.
Combining \eqref{eqn:A1-1}, \eqref{eqn:A1-2}, \eqref{eqn:A1-3}, we reach
\begin{align}\label{eqn:A1}
A_1 = O\left( \frac{\sqrt{\psi \gamma}}{1 - \psi / \phi + \sqrt{\psi \gamma}}\right).
\end{align}

\subsubsection{Bounding $A_2$}
Similar to \eqref{eqn:A1-1}, we have
\begin{align}\label{eqn:A2-1}
    2(\sigma_\ep^2 + \phi \cI_{1, 2}^{\rm s}) |\rho_{\rm t} \cI_{2, 2}^{\rm t} - a \rho_{\rm s} \cI_{2, 2}^{\rm s}| = O(1).
\end{align}
By \eqref{eqn:omegaidentity},
\begin{align}\label{eqn:A2identity}
    \frac{\kappa^2}{\rho_{\rm s}(1 - \kappa^2 \cI_{2, 2}^{\rm s})} = \frac{\kappa^2}{\rho_{\rm s}[\gamma \tau +  \kappa(\omega_{\rm s} + \phi \cI_{1, 2}^{\rm s})]}.
\end{align}
From \eqref{eqn:A2identity} and $\kappa = \gamma \rho_{\rm s} \tau \bar \tau$,
\begin{align*}
&\left\vert \frac{\kappa \phi \gamma \bar \tau}{\phi \gamma + \rho_{\rm s}(\psi \gamma \tau + \phi \gamma \bar \tau)(\omega_{\rm s} + \phi \cI_{1, 2}^{\rm s})} - \frac{\kappa^2}{\rho_{\rm s}(1 - \kappa
^2 \cI_{2, 2}^{\rm s})}\right\vert\\
&\hspace{3cm} = \frac{\kappa^2 (\omega_{\rm s} + \phi \cI_{1, 2}^{\rm s}) \psi \gamma \tau}{[\phi \gamma + \rho_{\rm s}(\psi \gamma \tau + \phi \gamma \bar \tau)(\omega_{\rm s} + \phi \cI_{1, 2}^{\rm s})][\gamma \tau +  \kappa(\omega_{\rm s} + \phi \cI_{1, 2}^{\rm s})]}.
\end{align*}
From \eqref{eqn:A1-2}, \eqref{eqn:A1-3}, and $\kappa(\omega_{\rm s} + \phi \cI_{1, 2}^{\rm s}) / [\gamma \tau + \kappa(\omega_{\rm s} + \phi \cI_{1, 2}^{\rm s})] \leq 1$, we get
\begin{align}\label{eqn:A2-2}
    \left\vert \frac{\kappa \phi \gamma \bar \tau}{\phi \gamma + \rho_{\rm s}(\psi \gamma \tau + \phi \gamma \bar \tau)(\omega_{\rm s} + \phi \cI_{1, 2}^{\rm s})} - \frac{\kappa^2}{\rho_{\rm s}(1 - \kappa
^2 \cI_{2, 2}^{\rm s})}\right\vert = O\left( \frac{\sqrt{\psi \gamma}}{1 - \psi/\phi + \sqrt{\psi \gamma}} \right).
\end{align}
Putting \eqref{eqn:A2-1} and \eqref{eqn:A2-2} together,
\begin{align}\label{eqn:A2}
    A_2 = O\left( \frac{\sqrt{\psi \gamma}}{1 - \psi/\phi + \sqrt{\psi \gamma}} \right).
\end{align}

\subsubsection{Bounding $\stsd_{\textnormal{I}}^{\rm s}(\phi, \psi, \gamma)$ and $\stsd_{\textnormal{SS}}^{\rm s}(\phi, \psi, \gamma)$}
By Theorem \ref{thm:asymp} and the equations \eqref{eqn:deftau}, \eqref{eqn:A1-2}, \eqref{eqn:A1-3}, we have
\begin{align}\label{eqn:Ibound}
    \stsd_{\textnormal{I}}^{\rm s}(\phi, \psi, \gamma) = O\left( \frac{1 + \sqrt{\psi \gamma}}{1 - \psi / \phi + \sqrt{\psi \gamma}} \right).
\end{align}
By Theorem \ref{thm:asymp} and the equations \eqref{eqn:A1-2}, \eqref{eqn:A1-3}, \eqref{eqn:A2-2}, we have
\begin{align}\label{eqn:SSbound}
    \stsd_{\textnormal{SS}}^{\rm s}(\phi, \psi, \gamma) = O\left( \frac{\psi + \sqrt{\psi \gamma}}{1 - \psi / \phi + \sqrt{\psi \gamma}} \right).
\end{align}

\subsubsection{Bounding $|a(\gamma) - a(0)|$}
From the argument in Section \ref{sec:cor32proof}, we know $a(\gamma)$ is differentiable with respect to $\gamma$.
By the chain rule and \eqref{eqn:deffunctionals},
\begin{align}\label{eqn:aprime}
 \frac{\partial a}{\partial \gamma} = \frac{\partial \kappa}{\partial \gamma} \times \frac{-\cI_{2, 2}^{\rm t} (\omega_{\rm s} + \cI_{1, 1}^{\rm s}) + \cI_{2, 2}^{\rm s}(\omega_{\rm t} + \cI_{1, 1}^{\rm t})}{(\omega_{\rm s} + \cI_{1, 1}^{\rm s})^2}.
\end{align}
By implicit differentiation of \eqref{eqn:defkappa}, we have
\begin{align}\label{eqn:partialkappa}
\frac{\partial \kappa}{\partial \gamma} = -\frac{\kappa}{\phi \gamma + \rho_{\rm s} (\psi \gamma \tau + \phi \gamma \bar\tau) (\omega_{\rm s} + \phi \cI_{1, 2}^{\rm s})}.
\end{align}
We have $\vert(-\cI_{2, 2}^{\rm t} (\omega_{\rm s} + \cI_{1, 1}^{\rm s}) + \cI_{2, 2}^{\rm s}(\omega_{\rm t} + \cI_{1, 1}^{\rm t})) / (\omega_{\rm s} + \cI_{1, 1}^{\rm s})^2 \vert = O(1)$ and
\begin{align*}
    &\left\vert \frac{\partial \kappa}{\partial \gamma} \right\vert = O\left( \frac{1}{\sqrt{(\psi - \phi)^2 + \psi \phi \gamma}} \right)
\end{align*}
since $\psi \gamma \tau + \phi \gamma \bar \tau = \sqrt{(\psi - \phi)^2 + 4 \kappa \psi \phi \gamma / \rho_{\rm s}}$.
Therefore,
\begin{align}\label{eqn:adiffbound}
|a(\gamma) - a(0)| &= \left\vert \int_0^\gamma \frac{\partial a}{\partial \gamma}(u)  du \right\vert \leq \int_0^\gamma \left\vert \frac{\partial a}{\partial \gamma}(u) \right\vert du \nonumber\\
&= O \left( \int_0^\gamma \frac{1}{\sqrt{(\psi - \phi)^2 + \psi \phi u}}  du \right) = O \left( \frac{\gamma}{ 1- \psi / \phi + \sqrt{\psi \gamma}} \right).
\end{align}

\subsubsection{Bounding $|b_\textnormal{I}(\gamma) - b_\textnormal{I}(0)|$}
From the argument in Section \ref{sec:cor32proof}, we know $b_\textnormal{I}(\gamma)$ is differentiable with respect to $\gamma$.
In \eqref{eqn:slopeintercept}, the terms $\frac{\kappa^2}{1 - \kappa^2 \cI_{2, 2}^{\rm s}}$, $\sigma_\ep^2 + \phi \cI_{1, 2}^{\rm s}$, $\rho_{\rm t} - a \rho_{\rm s} \cI_{2, 2}^{\rm s}$ and their derivatives with respect to $\kappa$ are $O(1)$.
Thus,
\begin{align*}
\left\vert \frac{\partial b_\textnormal{I}}{\partial \gamma} \right\vert = O\left( \left\vert \frac{\partial \kappa}{\partial \gamma} \right\vert\right) =  O\left( \frac{1}{\sqrt{(\psi - \phi)^2 + \psi \phi \gamma}} \right).
\end{align*}
Therefore,
\begin{align}\label{eqn:bIdiffbound}
|b_\textnormal{I}(\gamma) - b_\textnormal{I}(0)| &= \left\vert \int_0^\gamma \frac{\partial b_\textnormal{I}}{\partial \gamma}(u)  du \right\vert \leq \int_0^\gamma \left\vert \frac{\partial b_\textnormal{I}}{\partial \gamma}(u) \right\vert du \nonumber\\
&= O \left( \int_0^\gamma \frac{1}{\sqrt{(\psi - \phi)^2 + \psi \phi u}}  du \right) = O \left( \frac{\gamma}{ 1- \psi / \phi + \sqrt{\psi \gamma}} \right).
\end{align}

Theorem \ref{thm:approxlinear1} is proved by combining equations \eqref{eqn:totalbound}, \eqref{eqn:A1}, \eqref{eqn:A2}, \eqref{eqn:Ibound}, \eqref{eqn:SSbound}, \eqref{eqn:adiffbound}, \eqref{eqn:bIdiffbound}.

\subsection{Proof of Corollary \ref{cor:approxlinear2}}
By $E_j = B_j + V_j = B_j + \frac{1}{2} \stsd_{\textnormal{I}}^{j}(\phi, \psi, \gamma)$ and \eqref{eqn:brisk}, we have
\begin{align*}
    |E_{\rm t} -a E_{\rm s} - b_{\textnormal{risk}}| \leq \frac{1}{2} |\stsd_{\textnormal{I}}^{\rm t}(\phi, \psi, \gamma) - a \stsd_{\textnormal{I}}^{\rm s}(\phi, \psi, \gamma) - b_{\textnormal{I}}| + \left\vert B_{\rm t} - a B_{\rm s} - \lim_{\gamma \to 0} (B_{\rm t} - a B_{\rm s}) \right\vert.
\end{align*}
Since the derivatives of $\cI_{1, 1}^j, \cI_{1, 2}^j$ with respect to $\gamma$ is $O(1)$.
We have
\begin{align*}
\left\vert B_{\rm t} - a B_{\rm s} - \lim_{\gamma \to 0} (B_{\rm t} - a B_{\rm s}) \right\vert = O(\gamma)
\end{align*}
by the mean value theorem.
The conclusion follows from Theorem \ref{thm:approxlinear1}.

\section{Recap of \cite{tripuraneni2021covariate}}\label{sec:recap}
In this section, we restate some relevant results of \cite{tripuraneni2021covariate}, in the special cases $\Sigma^* = \Sigma_{\rm s}$ or $\Sigma^* = \Sigma_{\rm t}$.
See \cite{tripuraneni2021covariate} for the original theorems.
For a test distribution $x \sim \normal( 0, \Sigma^*)$, define the risk by
\begin{align*}
    E_{\Sigma^*} = \E_{x, \beta, X, Y, W}[(\beta^\top x - \hat{y}_{W, X, Y} (x))^2].
\end{align*}
We have the following bias-variance decomposition
\begin{align*}
    E_{\Sigma^*} &= \E_{x, \beta}[(\beta^\top x - \E_{W, X, Y} [\hat{y}_{W, X, Y}(x)])^2] + \E_{x, \beta}[\mathbb{V}_{W, X, Y}(\hat{y}_{W, X, Y}(x))]\\
    &= B_{\Sigma^*} + V_{\Sigma^*}.
\end{align*}
We consider the high-dimensional limit $n, d, N \to \infty$ with $d / n \to \phi$ and $d / N \to \psi$ of the above quantities when $\Sigma^* = \Sigma_{\rm s}$ or $\Sigma^* = \Sigma_{\rm t}$ converge as in our main text:
\begin{align*}
    E_j &= \lim_{n,d,N \to \infty} E_{\Sigma_{j}}, \quad B_j = \lim_{n,d,N \to \infty} B_{\Sigma_{j}}, \quad V_j = \lim_{n,d,N \to \infty} V_{\Sigma_{j}}, \quad j \in \{\rm s, \rm t\}.
\end{align*}

\begin{theorem}[Theorem 5.1 of \cite{tripuraneni2021covariate}]\label{thm:biasandvariance}
For  $j \in \{\rm s, \rm t\}$, the asymptotic bias and variance are given by
\begin{align*}
    B_{j} &= \left(1 - \sqrt{\frac{\rho_j}{\rho_{\rm s}}} \right)^2 m_j + 2 \left(1 - \sqrt{\frac{\rho_j}{\rho_{\rm s}}} \right) \sqrt{\frac{\rho_j}{\rho_{\rm s}}} \mathcal{I}_{1, 1}^j + \frac{\rho_j \phi}{\rho_{\rm s}} \mathcal{I}_{1, 2}^j,\\
    V_{j} &= - \frac{\rho_j \psi}{\phi} \frac{\partial \kappa}{\partial \gamma} \left[ \mathcal{I}_{1, 1}^{\rm s}(\omega_{\rm s} + \phi \mathcal{I}_{1, 2}^{\rm s})(\omega_j + \mathcal{I}_{1, 1}^j) + \frac{\phi^2}{\psi} \gamma \bar\tau \mathcal{I}_{1, 2}^{\rm s} \mathcal{I}_{2, 2}^j \right.\\
    &\quad \left. + \gamma \tau \mathcal{I}_{2, 2}^{\rm s} (\omega_j + \phi \mathcal{I}_{1, 2}^j) + \sigma_\ep^2 \left( (\omega_{\rm s} + \phi \mathcal{I}_{1, 2}^{\rm s})(\omega_j + \mathcal{I}_{1, 1}^j) + \frac{\phi}{\psi} \gamma \bar\tau \mathcal{I}_{2, 2}^j \right)  \right],
\end{align*}
where $\kappa, \tau, \bar\tau$ are defined in \eqref{eqn:defkappa} and \eqref{eqn:deftau}.
\end{theorem}

In the ridgeless limit $\gamma \to 0$, the variance $V_j$ is further simplified as follows.
\begin{corollary}[Corollary 5.1 of \cite{tripuraneni2021covariate}]
For  $j \in \{\rm s, \rm t\}$, the asymptotic variance in the ridgeless limit is
\begin{align*}
    \lim_{\gamma \to 0} V_{j} &= \frac{\rho_j \psi \kappa}{\rho_{\rm s}|\phi - \psi|}  (\sigma_\ep^2 + \mathcal{I}_{1, 1}^{\rm s})(\omega_j + \mathcal{I}_{1, 1}^j) + \begin{cases} \frac{\rho_j \kappa}{\rho_{\rm s}} \left(1 - \frac{\kappa(\omega_{\rm s} - \sigma_\ep^2)}{1 - \kappa^2 \mathcal{I}_{2, 2}^{\rm s}}\right) \mathcal{I}_{2, 2}^j & \phi \geq \psi, \\ \frac{\rho_j \kappa^2 \psi \mathcal{I}_{2, 2}^{\rm s}}{\rho_{\rm s}(\phi - \kappa^2 \psi \mathcal{I}_{2, 2}^{\rm s})}(\omega_j + \phi \mathcal{I}_{1, 2}^j) & \phi < \psi, \end{cases}
\end{align*}
where $\kappa$ is defined in \eqref{eqn:defkapparidgeless}.
\end{corollary}

Another important observation is that there is a linear relation between the asymptotic error under source and target domain.
\begin{proposition}[Proposition 5.6 of \cite{tripuraneni2021covariate}]\label{prop:linearrisk}
We assume $\phi$ is fixed.
In the ridgeless limit $\gamma \to 0$ and the overparameterized regime $\phi \geq \psi$, the error $E_{\rm t}$ is linear in $E_{\rm s}$, as a function of $\psi$.
That is,
\begin{align*}
    \lim_{\gamma \to 0} E_{\rm t} = b_{\textnormal{risk}} + \frac{\rho_{\rm t}(\omega_{\rm t} + \cI_{1, 1}^{\rm t})}{\rho_{\rm s}(\omega_{\rm s} + \cI_{1, 1}^{\rm s})}\lim_{\gamma \to 0} E_{\rm s},
\end{align*}
where the intercept
\begin{align}\label{eqn:brisk}
b_{\textnormal{risk}} = \frac{1}{2} b_{\textnormal{I}} + \lim_{\gamma \to 0} (B_{\rm t} - a B_{\rm s})
\end{align}
and the slope $\rho_{\rm t}(\omega_{\rm t} + \cI_{1, 1}^{\rm t}) / \rho_{\rm s}(\omega_{\rm s} + \cI_{1, 1}^{\rm s})$ are independent of $\psi$.
\end{proposition}

\section{Additional Experiments}
\label{sec:additional_experiments}
\vspace{-0.3cm}
\subsection{Estimation of the slope}\label{sec:slopeestimation}
\vspace{-0.1cm}
Let $\hat \Sigma_{\rm s}, \hat \Sigma_{\rm t}$ be sample covariance of test inputs from the source and target domains, respectively.
Denote the eigenvalues and corresponding eigenvectors of $\hat \Sigma_{\rm s}$ by $\hat \lambda_1^{\rm s}, \dots, \hat \lambda_d^{\rm s}$ and $\hat v_1, \dots, \hat v_d$.
Define $\hat \lambda_i^{\rm t} = \hat v_i^\top \hat \Sigma_{\rm t} \hat v_i$ for $i \in [d]$.
For $j \in \{\rm s, \rm t\}$, we estimate $\cI_{a, b}^j(\kappa)$ by
\begin{align*}
    \hat \cI_{a, b}^j(\kappa) = \frac{\phi}{d} \sum_{i = 1}^d \frac{(\hat \lambda_i^{\rm s})^{a - 1} \hat \lambda^j_i}{(\phi + \kappa \hat \lambda^{\rm s}_i)^b}.
\end{align*}
We estimate the constants defined in \eqref{eqn:defconstants} by replacing $m_j$ with $\hat{m}_j = \overline{\operatorname{tr}}(\hat \Sigma_j)$, $j \in \{\rm s, \rm t\}$.
Now, the self-consistent equation \eqref{eqn:defkapparidgeless} is estimated by
\begin{align*}
    \hat\kappa = \frac{\min(1, \phi / \psi)}{\hat{\omega}_{\rm s} + \hat \cI_{1, 1}^{\rm s}(\hat \kappa)},
\end{align*}
and its unique non-negative solution is denoted by $\hat \kappa$.
The existence and uniqueness of $\hat \kappa$ follows from Lemma A1.2 of \cite{tripuraneni2021covariate}.
We use
\begin{align*}
    \hat a = \frac{\hat{\rho}_t (\hat{\omega}_{\rm t} + \hat \cI_{1, 1}^{\rm t}(\hat \kappa))}{\hat{\rho}_s (\hat{\omega}_{\rm s} + \hat \cI_{1, 1}^{\rm s}(\hat \kappa))}
\end{align*}
as an estimate of the slope $a = \rho_{\rm t}(\omega_{\rm t} + \cI_{1, 1}^{\rm t}) / \rho_{\rm s}(\omega_{\rm s} + \cI_{1, 1}^{\rm s})$.

\vspace{-0.3cm}
\subsection{Deviation from the line}\label{sec:deviation}

\begin{figure}[ht]
\centering
\includegraphics[scale=0.85]{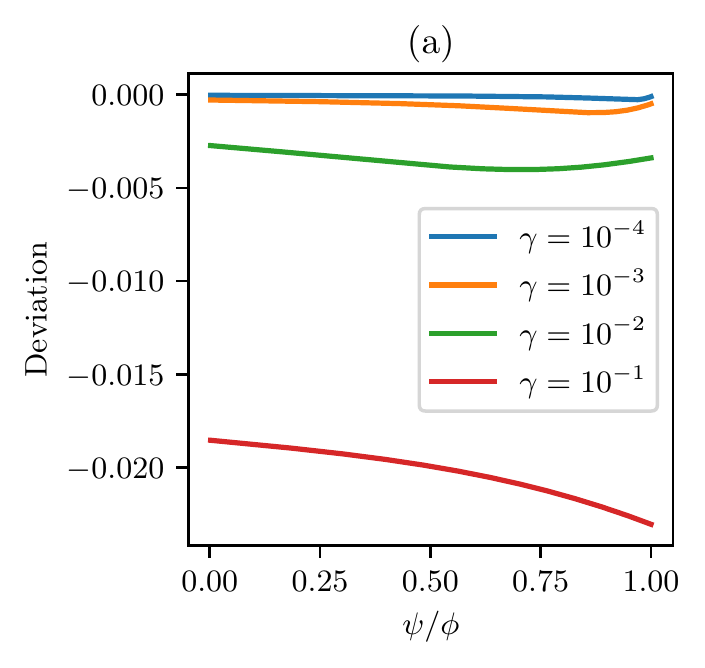}
\includegraphics[scale=0.85]{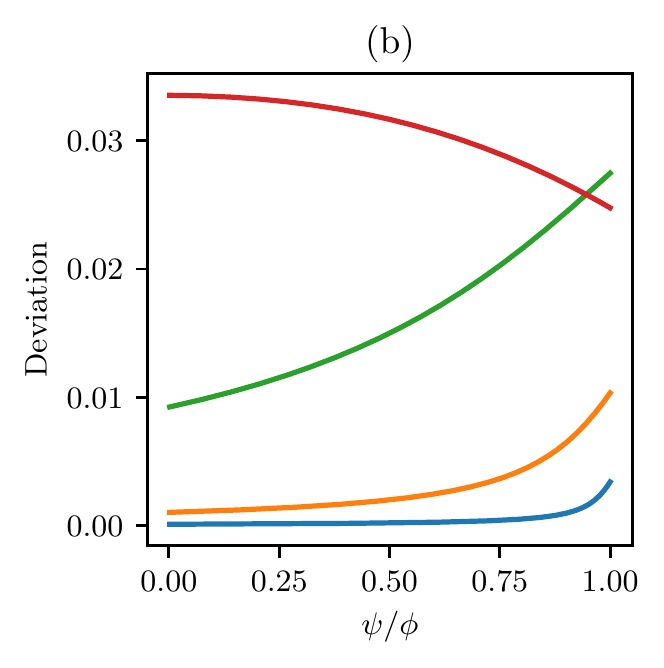}

\caption{\textbf{(a)} Deviation from the line,  $\stsd_{\textnormal{I}}^{\rm t}(\phi, \psi, \gamma) - a \stsd_{\textnormal{I}}^{\rm s}(\phi, \psi, \gamma) - b_I$, as a function of $\psi$ for non-zero $\gamma$. \textbf{(b)} Deviation from the line,  $E_{\rm t} - a E_{\rm s} - b_\textnormal{risk}$, as a function of $\psi$ for non-zero $\gamma$. We use $\phi = 0.5$, $\sigma_\ep^2 = 10^{-4}$, ReLU activation $\sigma$, and $\mu = 0.4 \delta_{(0.1, 1)} + 0.6 \delta_{(1, 0.1)}$}
\label{fig:deviation}
\end{figure}

\vspace{-0.1cm}
Figure~\ref{fig:deviation} displays 
the deviation from the line for I disagreement and risk, when non-zero ridge regularization $\gamma$ is used.
Similar to Figure \ref{fig:linear} \textbf{(b)}, the deviation is smaller for $\gamma$ closer to zero.
However, unlike SS disagreement, 
the deviation is non-zero even in the infinite overparameterization limit $\psi \to 0$.
This is consistent with the upper bound we present in Theorem \ref{thm:approxlinear1} and Corollary \ref{cor:approxlinear2}.

\subsection{Varying Corruption Severity}\label{sec:varyingcorruption}
\begin{figure}
    \centering
    \includegraphics{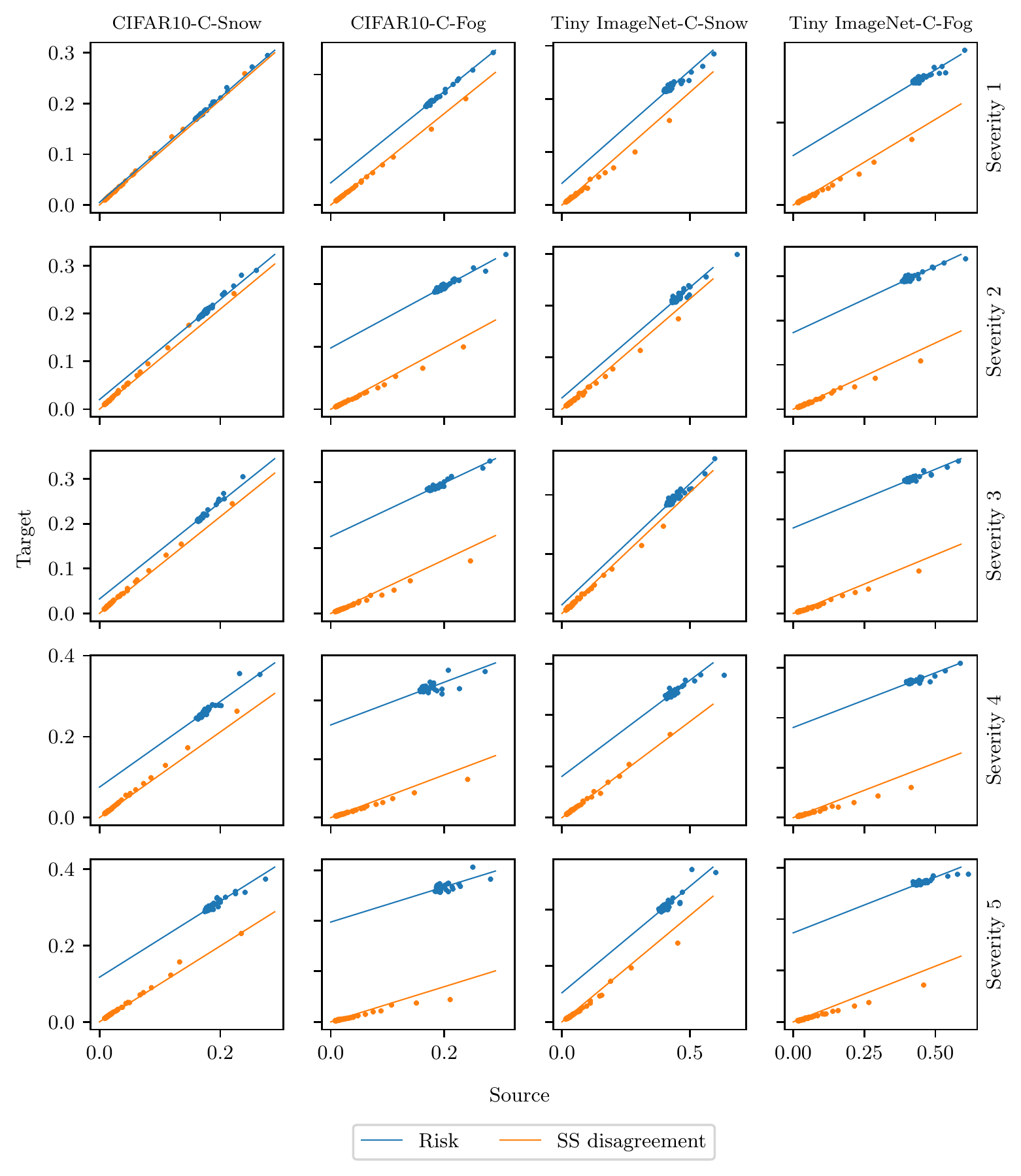}
    \caption{Target vs. source shared-sample disagreeement on CIFAR-10 and Tiny ImageNet with varying corruption severity. Experimental setting is identical to Section \ref{sec:simulation}.}
    \label{fig:large_experiment1}
\end{figure}

CIFAR-10-C and Tiny ImageNet-C has different levels of corruption severity, ranging from one to five.
We only included a few selected results in the main text due to space limitation.
We present the plots for all severity levels in Figure \ref{fig:large_experiment1}.

\subsection{I and SW disagreement}\label{sec:ISWdisagreement}
\begin{figure}
    \centering
    \includegraphics{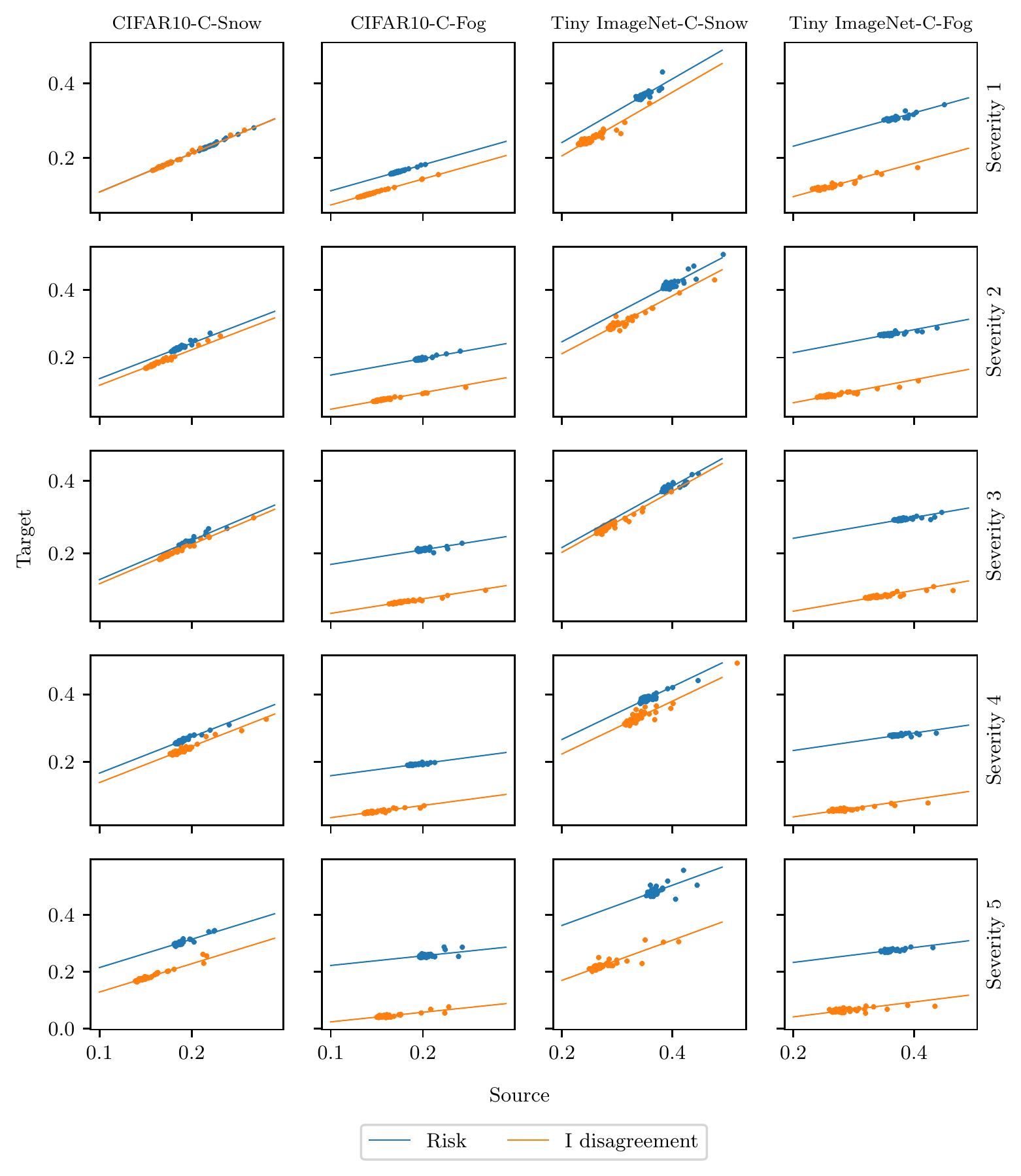}
    \caption{Target vs. source independent disagreeement on CIFAR-10 and Tiny ImageNet with varying corruption severity. Experimental setting is identical to Section \ref{sec:simulation}.}
    \label{fig:Idisagree}
\end{figure}

\begin{figure}
    \centering
    \includegraphics{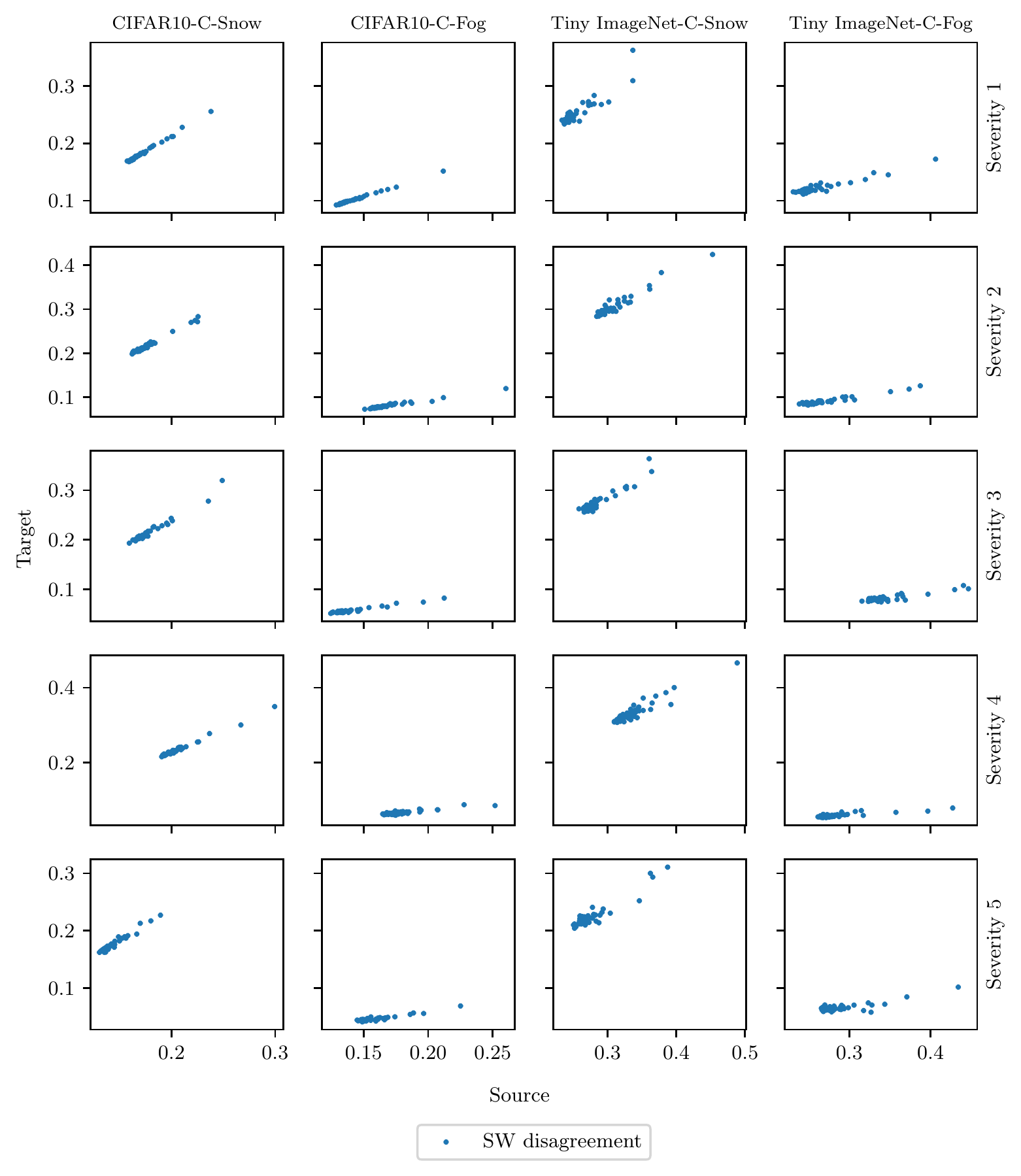}
    \caption{Target vs. source shared-weight disagreeement on CIFAR-10 and Tiny ImageNet with varying corruption severity. Experimental setting is identical to Section \ref{sec:simulation}.}
    \label{fig:SWdisagree}
\end{figure}

In Figure \ref{fig:Idisagree}, Figure \ref{fig:SWdisagree}, Figure \ref{fig:accagreement} \textbf{(a), (b)}, we repeat the experiment in Section \ref{sec:varyingcorruption} for I and SW disagreement.
Since our theory suggests that the disagreement-on-the-line phenomenon does not occur for SW disagreement, we do not plot theoretical predictions for SW disagreement.

\subsection{Accuracy and Agreement}
\begin{figure}[h!]
 \centering
 \includegraphics[scale=0.85]{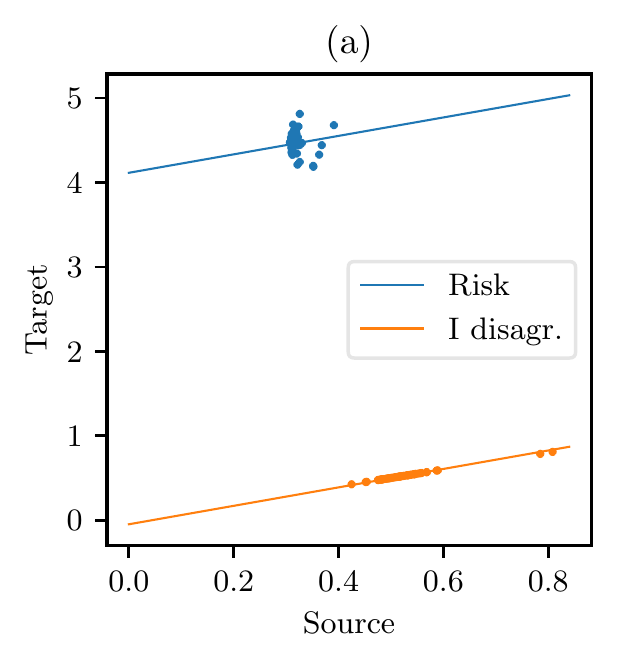}
 \includegraphics[scale=0.85]{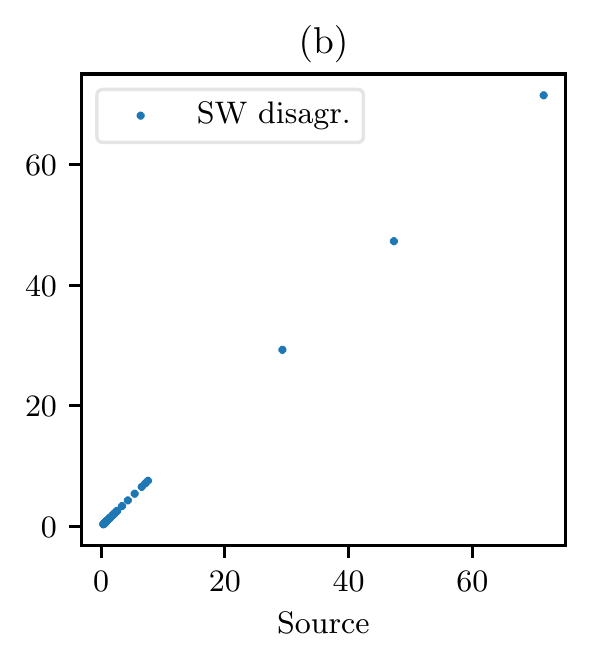}
 \includegraphics[scale=0.85]{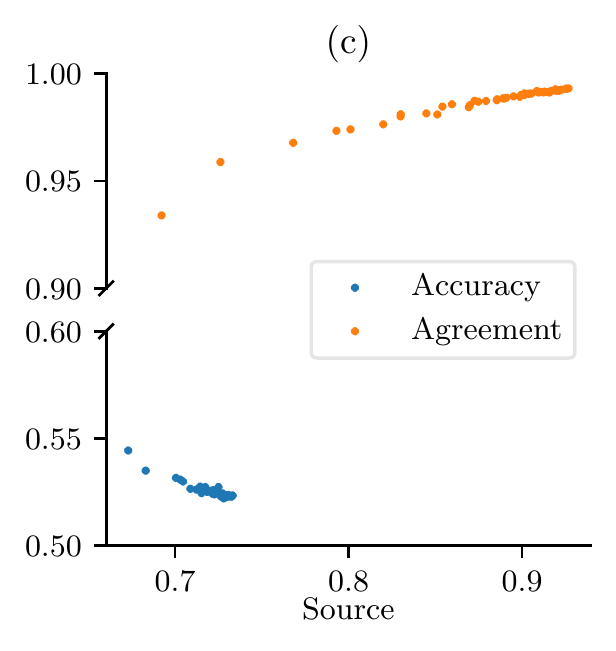}
 \caption{\textbf{(a)} Target vs. source independent disagreement of random features model trained on Camelyon17. \textbf{(b)} Target vs. source shared-weight disagreement of random features model trained on Camelyon17. \textbf{(c)} Target vs. source accuracy and agreement of random features model trained on Camelyon17; Experimental setting is identical to Section \ref{sec:simulation}.}
\label{fig:accagreement}
 \end{figure}

\begin{figure}
    \centering
    \includegraphics{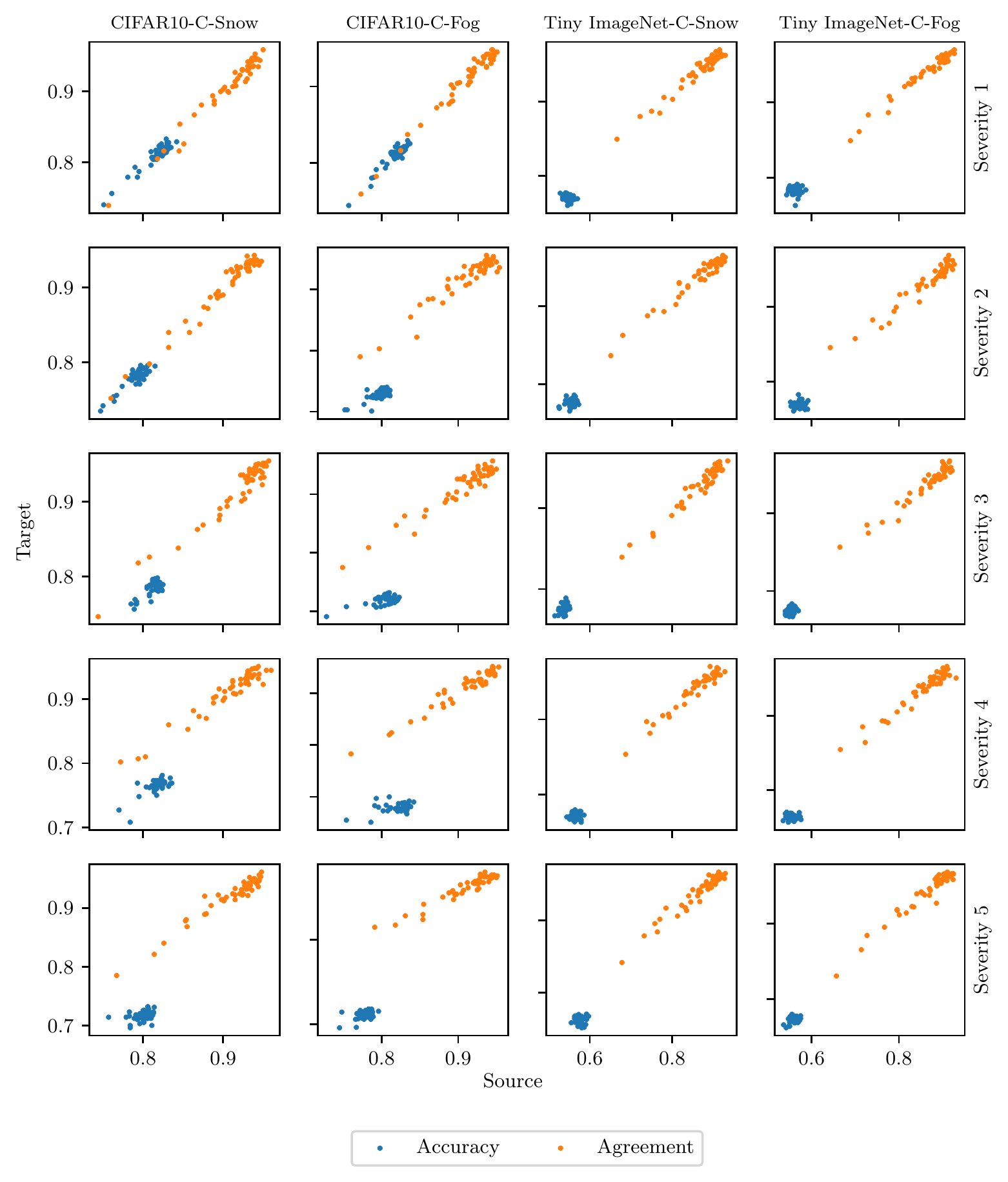}
    \caption{Target vs. source classification accuracy and agreement on CIFAR-10 and Tiny ImageNet with varying corruption severity. Experimental setting is identical to Section \ref{sec:simulation}.}
    \label{fig:large_experiment2}
\end{figure}

In the main text, we consider disagreement and risk defined in terms of mean squared error, but here we present classification accuracy and 0-1 agreement as studied in 
\cite{hacohen2020let,chen2021detecting,jiang2021assessing,nakkiran2020distributional,baekagreement,atanov2022task,pliushch2022deep,kirsch2022note}.
See Figures \ref{fig:large_experiment2} and Figure \ref{fig:accagreement} \textbf{(c)}.

\end{document}